%% file: colm2026_conference.tex
\documentclass{article} 
\usepackage[preprint]{colm2026_conference}

\usepackage{microtype}
\usepackage{hyperref}
\usepackage{url}
\usepackage{booktabs}
\usepackage{graphicx}

\usepackage{lineno}

\usepackage[many,most,skins,theorems]{tcolorbox}
\usepackage{multirow}
\usepackage{amsmath}
\usepackage{algorithm}
\usepackage{algpseudocode}
\usepackage{amssymb}
\usepackage{makecell}
\usepackage{multirow}
\usepackage{subcaption}
\usepackage{cleveref}
\usepackage{wrapfig}

\usepackage{enumitem}
\newenvironment{itemize*}%
 {\leftmargini=20pt\begin{itemize}%
  \setlength{\itemsep}{3pt}%
  \setlength{\parskip}{0pt}%
  }%
 {\end{itemize}} 
\newenvironment{enumerate*}%
 {\begin{enumerate}%
  \setlength{\itemsep}{0pt}%
  \setlength{\parskip}{0pt}}%
 {\end{enumerate}}

\usepackage{xcolor}
\tcbuselibrary{skins,breakable}
\definecolor{rqBlueBg}{HTML}{EAF4FF}

\tcbset{
  aibox/.style={
    width=\linewidth,
    top=7pt,
    bottom=2pt,
    colback=rqBlueBg,
    colframe=black,
    colbacktitle=black,
    enhanced,
    center,
    attach boxed title to top left={yshift=-0.1in,xshift=0.15in},
    boxed title style={boxrule=0pt,colframe=white},
  }
}

\newcounter{takeaway}
\newtcolorbox{takeaway}[1][]{
  aibox,
  title={\refstepcounter{takeaway}Takeaway \thetakeaway},
  #1
}

\definecolor{darkblue}{rgb}{0, 0, 0.5}
\hypersetup{colorlinks=true, citecolor=darkblue, linkcolor=darkblue, urlcolor=darkblue}

\usepackage{pifont}
\newcommand{\cmark}{\textcolor[rgb]{0.0, 0.6, 0.0}{\ding{51}}} 
\newcommand{\xmark}{\textcolor[rgb]{0.7, 0.0, 0.0}{\ding{55}}} 
\newcommand{\gmark}{\textcolor[rgb]{1,0.647,0}{\ding{51}}}

\NewDocumentCommand{\jiayu}
{ mO{} }{\textcolor{blue}{\textsuperscript{\textit{jiayu}}\textsf{\textbf{\small[#1]}}}}

\NewDocumentCommand{\qing}
{ mO{} }{\textcolor{brown}{\textsuperscript{\textit{qing}}\textsf{\textbf{\small[#1]}}}}

\NewDocumentCommand{\cheng}
{ mO{} }{\textcolor{orange}{\textsuperscript{\textit{cheng}}\textsf{\textbf{\small[#1]}}}}

\NewDocumentCommand{\xiusi}
{ mO{} }{\textcolor{cyan}{\textsuperscript{\textit{xiusi}}\textsf{\textbf{\small[#1]}}}}

\NewDocumentCommand{\zhenhailong}
{ mO{} }{\textcolor{purple}{\textsuperscript{\textit{zhenhailong}}\textsf{\textbf{\small[#1]}}}}

\definecolor{mypurple}{RGB}{128,0,128}

\title{AdaPlanBench: Evaluating Adaptive Planning in Large Language Model Agents under World and User Constraints}

\author{
\noindent Jiayu Liu,
Cheng Qian,
Zhenhailong Wang,
Bingxuan Li,
Jiateng Liu, \\
\bfseries\ Qing Zong, 
Heng Wang,
Jeonghwan Kim,
Yumeng Wang, 
Bingxiang He, \\
\bfseries\ Xiusi Chen, 
Yi R. (May) Fung,
Heng Ji \\ \\
University of Illinois Urbana-Champaign \\
}
\newcommand{\BENCH}[1]{\textbf{AdaPlanBench}}
\newcommand{\bench}[1]{AdaPlanBench}

\begin{document}

\ifcolmsubmission
\linenumbers
\fi

\maketitle

\vspace{-0.3in}
\begin{center}
\small

\newcommand{\logoh}{1.35em}

\href{https://github.com/JiayuJeff/AdaPlanBench}{
\raisebox{-0.2\height}{\includegraphics[height=\logoh]{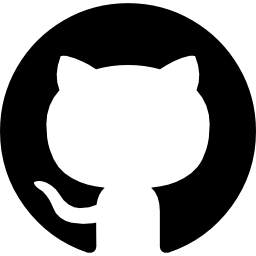}}
\hspace{0.35em}{\textbf{Code}}
}
\quad
\href{https://huggingface.co/datasets/JiayuJeff/AdaPlanBench}{
\raisebox{-0.2\height}{\includegraphics[height=\logoh]{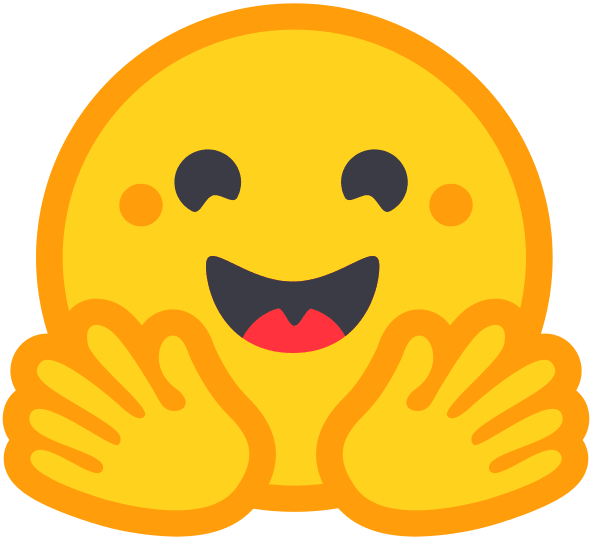}}
\hspace{0.35em}{\textbf{Dataset}}
}

\end{center}

\input{latex/0_abstract}

\input{latex/1_introduction}
\input{latex/2_related_work}
\input{latex/3_benchmark}
\input{latex/4_experiment}

\input{latex/5_analysis}

\input{latex/6_conclusion}

\section{Limitations}

\paragraph{Limited Domain Coverage.}
AdaPlanBench is currently instantiated in the household domain, which provides a natural setting where user preferences and world constraints interact. This choice offers a controlled and realistic testbed for studying adaptive planning, but some domain specific phenomena from settings such as travel, office workflows, or robotics may not yet be covered. We partially mitigate this limitation by focusing on general benchmark properties, including dual constraints, progressive disclosure, and open ended evaluation, which are not unique to household tasks. Future work can instantiate the same framework in additional domains to test how well the findings transfer.

\paragraph{Potential Bias in LLM-based Evaluation.}
Our evaluation relies on LLM judges for constraint checking and rubric scoring, which may introduce model-specific preferences or systematic bias. Although this setup enables scalable evaluation, it is still not equivalent to fully manual assessment. We partially mitigate this risk in two ways: first, we use multiple judges and aggregate their rubric scores to reduce the bias from a certain judge model; second, human annotation shows high consistency with the rubric-based LLM judgments. 

\paragraph{Text-only Evaluation Setting}
AdaPlanBench evaluates adaptive planning in a text-only interaction setting, without visual perception, embodied execution, or direct contact with real environments. This means the benchmark does not capture the full difficulty of real-world agent deployment, especially when perception and action grounding are tightly coupled. At the same time, this is a deliberate trade-off: by removing perception errors and low-level control noise, the benchmark more cleanly isolates adaptive planning abilities under progressively disclosed constraints. Future work can combine our setting with embodied or multimodal environments to study planning under more realistic scenarios.

\paragraph{Simplified Constraint Modeling.}
Our constraint construction adopts object-based world constraints and attribute-based user constraints, which improves clarity and verifiability but cannot fully capture the fine-grained, compositional, and sometimes ambiguous nature of real-world constraints. In particular, real user preferences are often softer, vaguer, and harder to canonicalize than the current benchmark format allows. We partially mitigate this issue by using multi-planner sampling and aggregation to ensure constraint diversity. Nevertheless, the resulting constraint space remains a simplified approximation of real-world planning requirements.

\section*{Ethics statement}

\paragraph{Offensive Content Elimination.} 
Our benchmark focuses on the household domain and a subset of data is manually validated to ensure the dataset is free of offensive material. Consequently, we are confident that it poses no risk of negative societal impact.

\paragraph{Licenses.} 
Our code will be released under the MIT license to allow unrestricted research use. The \bench{} will be distributed under a Creative Commons (CC) license, providing free access for the academic community. Our use of existing models and tools is strictly consistent with their original licenses and intended research purposes. We take full responsibility for any potential rights violations or licensing issues, and all resources comply with their respective terms of use while supporting research purposes.

\paragraph{Models.} 
All open-source models were hosted and executed locally using the vLLM library~\citep{vllm}, while all closed-source models were accessed through their respective official APIs. 
For reproducibility, the experimental settings are detailed in Section~\ref{app:model-setup}.

\paragraph{Data Annotations.}

All data annotation was performed by PhD-level researchers with relevant expertise, ensuring that the process was conducted responsibly and in accordance with ethical standards.

\bibliography{colm2026_conference}
\bibliographystyle{colm2026_conference}

\input{latex/10_Appendix}

\end{document}

%% file: latex/0_abstract.tex
\begin{abstract}

Planning for real-world problems by language models often involves both world and user constraints, which may not be fully specified upfront and are progressively disclosed through interaction. 
However, existing benchmarks still underexplore adaptive planning under such progressively revealed dual constraints.
To address this gap, we introduce \BENCH{}, a dynamic interactive benchmark for evaluating whether Large Language Model (LLM) agents can adaptively plan and re-plan under progressively revealed world and user constraints.
\bench{} is built on 307 household tasks, with a scalable constraint construction pipeline that augments each task with dual constraints.
At runtime, agents interact with the environment in a multi-turn protocol where hidden constraints are revealed only when the agent proposes a plan that violates them, requiring iterative plan revision under accumulating feedback. 
This makes planning challenging, as agents must infer and track constraints from feedback while re-planning effectively.
Experiments on ten leading LLMs show that adaptive planning under dual constraints remains challenging, with the best model reaching only 67.75\% accuracy.
We further observe that performance degrades as more constraints accumulate, with user constraints posing a particularly large challenge and failures often stemming from weaker physical grounding and reduced effectiveness.
These results establish \bench{} as a testbed for dual-constrained interactive planning and highlight the challenge of reliable adaptation to dynamically revealed constraints in LLM agents. 
\end{abstract}

%% file: latex/1_introduction.tex
\section{Introduction}
\label{sec:intro}

Large Language Model (LLM) agents have achieved remarkable success in real-world interactive tasks, including writing complex code~\citep{claudecode,code2math}, operating computers~\citep{UI-TARS,UI-TARS-2}, and supporting scientific discovery~\citep{DR-Tulu,Deep-Research-Agents}.
While these applications differ in domain, they share a common structure: agent capability depends on sustained interaction throughout task execution~\citep{ReAct,park2023generativeagentsinteractivesimulacra}.
This interaction typically unfolds along two closely linked dimensions: \textbf{interaction with users}, through which agents infer goals and preferences, and \textbf{interaction with the external world}, where they gather information and take actions via tools and interfaces~\citep{xi2023risepotentiallargelanguage,Wang_2024}.
Because such interaction is inherently multi-step, it usually requires \textbf{planning}~\citep{huang2024understandingplanningllmagents}: agents must anticipate future outcomes~\citep{current-agent-fail-wm-as-tool} and dynamically adapt their actions as interaction unfolds~\citep{costbench} across both their engagement with users and the external world.
Planning therefore naturally spans both dimensions, reflecting the dual structure of agent interaction. 

However, real-world planning is rarely unconstrained. Because interaction has a dual structure, agents must handle \textbf{dual constraints} from both the user and the world: user constraints such as preferences and priorities, and world constraints such as tool availability and resource limitations.
Existing benchmarks typically consider only one side of this problem, focusing on either user constraints~\citep{UserBench,opentom,URS-dataset} or world constraints~\citep{appworld,tau-2-bench,PlanBench}, leaving their joint handling largely unexplored.
This raises our central research question: \textbf{can LLM agents plan effectively under both user and world constraints?} 
In practice, this problem is further complicated by two key challenges:
(1) \textbf{Progressive constraint disclosure}: constraints are often implicit rather than specified upfront, requiring agents to uncover them incrementally through proactive exploration.
(2) \textbf{Large action and solution spaces}: real-world tasks involve vast spaces of possible actions and solutions, making performance harder to measure. 
These challenges call for a rigorous benchmark that evaluates whether agents can adaptively plan under dual constraints that are progressively revealed in open-ended planning settings.

\begin{table*}[!t]
\centering
\resizebox{\linewidth}{!}{
    \begin{tabular}{lcccccccc}
    \toprule
    \textbf{Benchmark} 
    & \textbf{\makecell{Iterative\\Re-planning}} 
    & \textbf{\makecell{User\\Interaction}}
    & \textbf{\makecell{World\\Interaction}}
    & \textbf{\makecell{Dual\\Constraint}} 
    & \textbf{\makecell{Progressive\\Disclosure}}
    & \textbf{\makecell{Open-Ended\\ Evaluation}} 
    & \textbf{\makecell{Scalable\\Constraints}} \\
    \midrule
    \textit{CostBench}~\citep{costbench} & \gmark & \xmark & \cmark & \gmark & \xmark & \xmark & \gmark
    \\
    \textit{FlowBench}~\citep{FlowBench} & \gmark & \cmark & \cmark & \gmark & \gmark & \xmark & \xmark
    \\ 
    \textit{NaturalPlan}~\citep{NATURAL-PLAN} & \xmark & \xmark & \xmark & \cmark & \xmark & \gmark & \cmark
    \\ 
    \textit{PrefEval}~\citep{PrefEval} & \xmark & \cmark & \xmark & \xmark & \xmark & \gmark & \cmark
    \\ 
    \textit{RealPref}~\citep{RealPref} & \xmark & \xmark & \xmark & \xmark & \cmark & \cmark & \xmark
    \\ 
    \textit{UserBench}~\citep{UserBench} & \xmark & \cmark & \cmark & \xmark & \cmark & \cmark & \cmark
    \\ 
    \textit{PersonaMem-v2}~\citep{PersonaMem-v2} & \xmark & \cmark & \xmark & \xmark & \cmark & \cmark & \cmark
    \\ 
    $\tau$-Bench~\citep{tau-bench} & \xmark & \cmark & \cmark & \gmark & \gmark & \gmark & \xmark
    \\ 
    \textit{$\tau^2$-Bench}~\citep{tau-2-bench} & \xmark & \cmark & \cmark & \xmark & \cmark & \xmark & \cmark
    \\ 
    \textit{TravelPlanner}~\citep{TravelPlanner} & \xmark & \xmark & \cmark & \cmark & \xmark & \cmark & \cmark
    \\ 
    \midrule
    \textit{\BENCH{} \textbf{(Ours)}} & \cmark & \cmark & \cmark & \cmark & \cmark & \cmark & \cmark
    \\
    \bottomrule
    \end{tabular}
}
\vspace{-1.5mm}
\caption{Comparison of \bench{} with prior related benchmarks across seven key properties. For each benchmark, the table reports whether each trait is fully (\cmark), partially (\gmark), or not (\xmark) addressed. 
Detailed explanations are provided in Appendix~\ref{app:traits-exp}.}
\label{tab:comparison-table}
\vspace{-0.2in}
\end{table*}

To this end, we introduce \BENCH{}, a dynamic and interactive benchmark evaluating LLM's adaptive planning ability under world and user constraints. 
We build \bench{} on top of the MacGyver dataset~\citep{MacGyver}, starting from a curated subset of 307 household-domain instances.
Using a scalable automated pipeline, we augment each task with world constraints that capture environmental limitations and user constraints that capture grounded personal preferences.
This setup allows any solution that satisfies these constraints, while preserving a large and effectively unbounded action space.
During evaluation, constraints are withheld at the outset and disclosed progressively when the agent proposes violating actions, forcing the agent to iteratively and adaptively re-plan in response to newly revealed constraints.

We evaluate ten leading open-source and proprietary LLMs on \bench{} and find that even the strongest model achieves only 67.75\% accuracy, while open-weight models typically remain at or below 30\%.
Moreover, planning quality becomes harder to sustain as progressively disclosed constraints accumulate along a trajectory, and deteriorates further as the overall constraint burden increases.
These challenges are not easily alleviated by explicit constraint tracking or rubric-based feedback, and are particularly severe when user constraints account for a large share of the difficulty.
Further analysis suggests that, under accumulated dual constraints, planning failures are often marked by reduced goal effectiveness and weaker physical plausibility.
Overall, \bench{} lays a foundation for future research on adaptive planning agents, motivating the development of models that can dynamically plan, adapt, and revise under real-world dual constraints.

%% file: latex/3_benchmark.tex
\section{AdaPlanBench Construction}

\begin{figure}
    \centering
    \includegraphics[width=\linewidth]{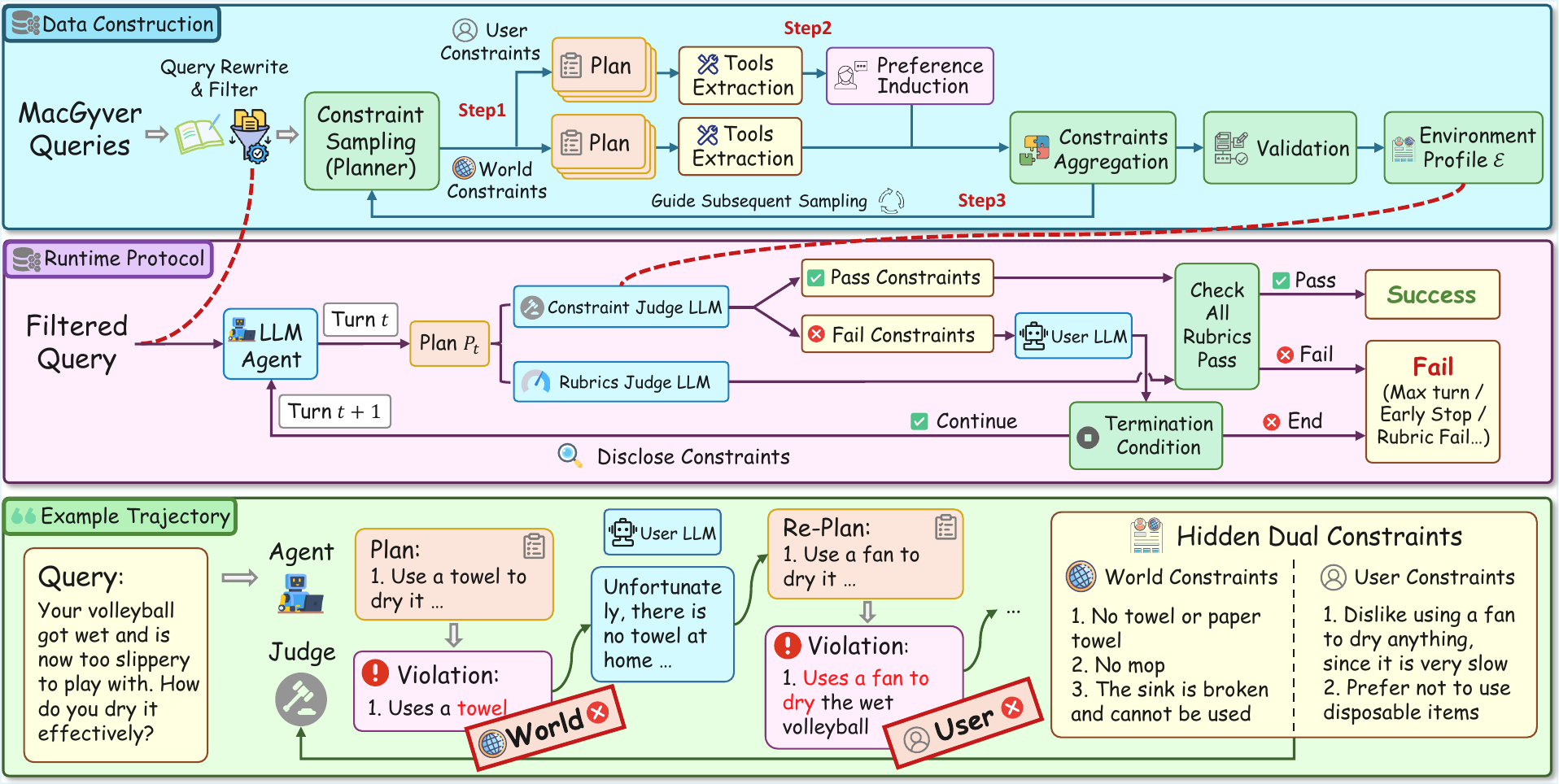}
    \vspace{-0.2in}
    \caption{Overview of AdaPlanBench. \textit{Top}: data construction, where dual constraints are constructed for each query. \textit{Middle}: runtime interaction, where the agent proposes plans, receives feedback on violated constraints, and re-plans iteratively. \textit{Bottom}: an example trajectory showing how hidden constraints are progressively disclosed during interaction.
    }
    \label{fig:overview}
    \vspace{-4mm}
\end{figure}

\BENCH{} is a dynamic, interactive benchmark for evaluating agents' adaptive planning under dual constraints, rooted in household tasks where both world and user constraints naturally arise~\citep{chang2024partnrbenchmarkplanningreasoning}.
As illustrated in Figure~\ref{fig:overview}, AdaPlanBench consists of two tightly coupled parts: an automatic pipeline that constructs world and user constraints for each MacGyver query, and a runtime protocol that progressively reveals violated constraints to elicit adaptive re-planning. 
This setting poses a distinctive challenge for agents, which must infer latent constraints from partial feedback, keep track of previously disclosed violations, and continuously revise their plans under an evolving constraint set.
As shown in \Cref{tab:comparison-table}, \BENCH{} distinguishes itself from prior work by focusing on interactive planning under both user and world constraints.
We discuss related work and the significance of this setting further in Appendix~\ref{app:related-works} and~\ref{app:discussion-significance-generalization}.

\subsection{Data Construction}
\label{sec:data-construction}
To construct each benchmark instance, we first rewrite and filter raw MacGyver queries, and then build a dual-constraint profile for each retained query using a multi-agent framework.
For each filtered query from MacGyver, we construct a dual-constraint profile using a multi-agent framework with specialized components. Formally, each instance is mapped to $(q,\mathcal{E})$, where $q$ is the query and $\mathcal{E}=(\mathcal{B}_{w},\mathcal{B}_{u})$ is the resulting constraint profile, comprising a world constraint set $\mathcal{B}_{w}$ and a user constraint set $\mathcal{B}_{u}$. The framework uses a set of role-specific models \{${\mathcal{M}_{\mathrm{rw}}, \mathcal{M}_{\mathrm{flt}}, \mathcal{M}_{\mathrm{plan}}, \mathcal{M}_{\mathrm{ext}}, \mathcal{M}_{\mathrm{merge}}, \mathcal{M}_{\mathrm{chk}}}$\}, which respectively denote a query rewriter, a binary query filter, a set of planner samplers, a constraint extractor, a merge model, and a constraint checker. Model choice details are provided in Appendix~\ref{app:model_choise}.

\noindent\textbf{Query rewriting and filtering.}
For each raw MacGyver query $q^{\mathrm{raw}}$, we first use the rewriter $\mathcal{M}_{\mathrm{rw}}$ to produce a short, method-agnostic household query {\small $q=\mathcal{M}_{\mathrm{rw}}(q^{\mathrm{raw}})$} by removing explicit resource constraints (e.g., \textit{tools available: ...} or \textit{using only ...}) while preserving the original goal. We then apply a strict binary filter $\mathcal{M}_{\mathrm{flt}}$ to retain only concrete household tasks that require multi-step planning. 
Because this process relaxes only the original resource constraints, we preserve the corresponding MacGyver reference solution for each retained query and use it later for constraint extraction and validation, thus ensuring task solvability. 
Detailed filtering rules are provided in Appendix~\ref{app:data_filtering_rule_exp} and illustrated in \Cref{fig:filtering_rules}.

To construct the dual-constraint profile, we iteratively repeat three steps: (1) sample candidate plans for the task, (2) extract salient tools from the plans and convert them into world or user constraints, and (3) merge the constraints to guide the next round of sampling.

\noindent\textbf{Step 1: Plan sampling.}
We first sample diverse candidate plans to surface the tools and strategies that are likely to matter for the task.
We employ a family of $J$ planner samplers {\small $\{\mathcal{M}^{(j)}_{\mathrm{plan}}\}_{j=1}^{J}$} and run the procedure for at most $R$ rounds, yielding a sequence of progressively enriched constraint profiles $\mathcal{E}_{low}, \mathcal{E}_{mid}, \mathcal{E}_{high}$. At each round $r$, every planner samples plans:
\begin{footnotesize}
\begin{equation}
\pi^{(j)}_{r} = \mathcal{M}^{(j)}_{\mathrm{plan}}\!\left(q,\mathcal{B}^{(j)}_{w,r-1},\mathcal{B}^{(j)}_{u,r-1}\right),
\end{equation}
\end{footnotesize}
where {\small $\mathcal{B}^{(j)}_{w,r-1}$} and {\small $\mathcal{B}^{(j)}_{u,r-1}$} denote the world and user constraint pools accumulated for planner $j$ up to round $r-1$, respectively. Both pools are initialized as empty sets. Intuitively, {\small $\mathcal{B}^{(j)}_{w,r-1}$} records world constraints such as unavailable tools or environmental limitations that subsequent plans must respect, while {\small $\mathcal{B}^{(j)}_{u,r-1}$} records user constraints such as preferences or requirements that later plans should not violate.

\noindent\textbf{Step 2: Constraint extraction.}
We then transform tools and their usage in these plans into grounded world and user constraints.
In our benchmark, world constraints capture tool availability and usability in the environment, whereas user constraints capture whether the tools used in a plan, or the attributes implied by their use, align with user preferences. This abstraction is designed to preserve both groundedness and evaluability (see Appendix~\ref{app:user-attribute-discussion}).

Given a sampled plan {\small $\pi^{(j)}_{r}$}, we first use $\mathcal{M}_{\mathrm{ext}}$ to extract the tools used in the plan:
\begin{footnotesize}
\begin{equation}
\mathcal{T}^{(j)}_{r} = \mathcal{M}_{\mathrm{ext}}(q, \pi^{(j)}_{r})
\end{equation}
\end{footnotesize}
These extracted tools form the raw basis for constraint construction. We then derive world and user constraints from them respectively as follows:
\begin{itemize}[topsep=-1.5pt, leftmargin=10pt, itemsep=0pt]
\item For \textbf{world constraints}, we directly convert each extracted tool into a constraint candidate {\small $\mathcal{C}^{(j)}_{w,r}$} by restricting its availability or use. For example, for the query of removing wrinkles from a suit, a world constraint could be \textit{there is no iron at home}.
\item For \textbf{user constraints}, we use $\mathcal{M}_{\mathrm{ext}}$ to further infer attributes of tools or their usage that may matter to the user, and then formulate corresponding constraint candidates {\small $\mathcal{C}^{(j)}_{u,r} \sim \mathcal{M}_{\mathrm{ext}}(q, \mathcal{T}^{(j)}_{r})$}.
For the same example, a user constraint could be \textit{I am concerned about using tools that generate high heat}, where \textit{generate high heat} is the inferred attribute associated with the tool \textit{iron}.
\end{itemize}

\noindent\textbf{Step 3: Constraint merging.}
Finally, we merge and canonicalize discovered constraints so they can reliably guide later rounds of planning.
After extraction, we use $\mathcal{M}_{\mathrm{merge}}$ to combine newly generated constraints with the previously accumulated ones, canonicalizing and deduplicating them to maintain a consistent representation:
\begin{footnotesize}
\begin{equation}
\mathcal{B}^{(j)}_{w,r} = \mathcal{M}_{\mathrm{merge}}(\mathcal{B}^{(j)}_{w,r-1} \cup \mathcal{C}^{(j)}_{w,r}), \quad
\mathcal{B}^{(j)}_{u,r} = \mathcal{M}_{\mathrm{merge}}(\mathcal{B}^{(j)}_{u,r-1} \cup \mathcal{C}^{(j)}_{u,r})
\end{equation}
\end{footnotesize}
The resulting planner-specific constraint sets serve two purposes: they are fed back into subsequent rounds to guide later plan sampling, and they provide the planner-level inputs for round-wise aggregation and validation.

\noindent\textbf{Final profile formation.}
After at most $R$ rounds, we aggregate the planner-specific constraint pools across all planners and use $\mathcal{M}_{\mathrm{chk}}$ to validate the merged constraints, yielding the final dual-constraint profile:
\begin{footnotesize}
\begin{equation}
\mathcal{B}_{w} = \mathcal{M}_{\mathrm{chk}}(\bigcup_{j=1}^{J}\mathcal{B}^{(j)}_{w,R}, q), \quad
\mathcal{B}_{u} = \mathcal{M}_{\mathrm{chk}}(\bigcup_{j=1}^{J}\mathcal{B}^{(j)}_{u,R}, q), \quad
\mathcal{E}=(\mathcal{B}_{w},\mathcal{B}_{u})
\end{equation}
\end{footnotesize}
Here, $\mathcal{M}_{\mathrm{chk}}$ acts as a final safeguard, removing vague or invalid constraints. For user constraints, it additionally filters out preference sets that are internally contradictory or jointly exhaustive, since such combinations would eliminate any feasible preference-consistent solution. For example, the pair \textit{I dislike quiet atmosphere} and \textit{I dislike noisy places} would be removed because it effectively rules out the entire relevant preference space.


\begin{wraptable}{r}{0.48\linewidth}
\centering
\vspace{-0.05in}
\small
\begin{tabular}{l ccc}
\toprule
\textbf{\# Avg. Number} & \textbf{$\mathcal{E}_{\mathrm{low}}$} & \textbf{$\mathcal{E}_{\mathrm{mid}}$} & \textbf{$\mathcal{E}_{\mathrm{high}}$} \\
\midrule
World Constraint & 9.76 & 19.61 & 37.73 \\
User Constraint & 10.91 & 21.78 & 41.79 \\
\bottomrule
\end{tabular}
\vspace{-2mm}
\caption{Statistics summary across three levels of dual-constraint profiles.}
\vspace{-4mm}
\label{tab:benchmark-stats}
\end{wraptable}

Based on the sampling rounds, we further divide instances into three difficulty tiers, $\mathcal{E}_{\mathrm{low}}$, $\mathcal{E}_{\mathrm{mid}}$, and $\mathcal{E}_{\mathrm{high}}$. Fewer constraints imply a lower re-planning burden, whereas more constraints imply a higher one (statistics in Table~\ref{tab:benchmark-stats}). 
We provide full details of our data construction pipeline and the intuition behind in Appendix~\ref{app:env_cons_algorithm} and~\ref{app:env_cons_intuition}.

\subsection{Agent--User--World Interaction}
\label{sec:inference-loop}

\noindent\textbf{Runtime interaction protocol.}
We evaluate \bench{} in a dynamic multi-turn setting, where agents must adaptively revise their plans as constraints are progressively revealed.
Each instance consists of a query \(q\) and a hidden dual-constraint profile \(\mathcal{E}\), which contains both world and user constraints.
At turn \(t\), the agent proposes a plan \(p_t\).
LLM judges then evaluate \(p_t\) for world-constraint satisfaction, user-constraint satisfaction, and rubric-based planning quality.
This identifies the violated world constraints \(V_t^w \subseteq \mathcal{B}_w\), the violated user constraints \(V_t^u \subseteq \mathcal{B}_u\), and a turn-level rubric score.
The violated constraints \(V_t^w\) and \(V_t^u\) are passed to a user simulator \(\mathcal{M}_{\mathrm{user}}\), which generates feedback for the current turn:
\begin{footnotesize}
\begin{equation}
f_t = \mathcal{M}_{\mathrm{user}}(V_t^w, V_t^u)
\end{equation}
\end{footnotesize}
This feedback directly reveals the newly disclosed constraints.
The agent then updates its plan accordingly and produces \(p_{t+1}\).
This creates a dynamic feedback loop in which constraint disclosure depends on the agent's own proposal, making successful performance contingent on adaptive re-planning.

\noindent\textbf{Termination condition.}
A trajectory terminates upon any of the following conditions:
\begin{itemize}[topsep=-1.5pt, leftmargin=10pt, itemsep=0pt]
\item \textbf{Valid plan found:} The proposed plan satisfies all constraints, i.e., \(V_t^w = V_t^u = \emptyset\).
\item \textbf{Maximum turn budget reached:} The interaction reaches the maximum turn budget \(T\).
\item \textbf{Early stopping triggered:} The interaction stops early if the agent fails to violate any \textit{new} constraints for two consecutive turns, where \textit{new} is defined relative to previously disclosed violations.
\end{itemize}
The intuition is that if the trajectory has not terminated but no new constraints are violated, the agent is likely repeating actions that conflict with already disclosed constraints, without making meaningful progress.
We provide the specific budget and threshold choices, along with their justification, in Appendix~\ref{app:discuss-parameter-choice}.

\noindent\textbf{Rubric evaluation with LLM judges.}
At each turn, we also evaluate plan quality using rubric-based judges and use the resulting turn-level scores for diagnostic analysis.
The plan is rated on four major dimensions using a scale 1 to 5 (see \Cref{tab:rubrics-details} for full definitions):
\begin{itemize}[topsep=-1.5pt, leftmargin=10pt, itemsep=0pt]
\item \textbf{Tool-Use Feasibility:} whether the tools invoked in the plan are available in the household environment.
\item \textbf{Physical Plausibility:} whether the proposed use of those tools can plausibly produce the intended effects.
\item \textbf{Effectiveness:} whether the plan, if executed as intended and if each step succeeds, would accomplish the task.
\item \textbf{Safety:} whether executing the plan would avoid causing harm to people.
\end{itemize}
We average scores across judges for each dimension to obtain an aggregated rubric score vector.
A plan passes the rubric evaluation only if every aggregated dimension score passes a threshold \(\gamma\); otherwise, it fails the rubric criteria.
In our experiments, we set \(\gamma=4\), which enforces a meaningful quality threshold without being overly strict in heavily constrained settings.
An ablation over different choices of \(\gamma\) is provided in Appendix~\ref{app:rubrics-pass-threshold-ablation}.
We use three rubric judges in all experiments.
Additional details on model choices and runtime interaction are provided in Appendix~\ref{app:model_choise} and~\ref{app:runtime-interaction-details}.
Human annotation shows that rubric-based LLM judgments are valid and consistent with human judges (details in Appendix~\ref{app:human-annotation}).

%% file: latex/4_experiment.tex
\section{Experiment}
\label{sec:experiment}

\subsection{Experiment Setup}
\label{sec:experiment-setup}

\paragraph{Models.}
We evaluate proprietary and open-source models to ensure a balanced and comprehensive assessment of current LLM capabilities. 
Proprietary models include GPT~\citep{gpt}, DeepSeek~\citep{deepseek2026v4preview}, and Gemini~\citep{gemini31pro}; while open-source models include Qwen3~\citep{qwen3technicalreport}, and Llama3~\citep{grattafiori2024llama3herdmodels}. 
We manually validate the reliability of both $M_{\mathrm{chk}}$ used in data construction and the runtime judge models, as detailed in Appendix~\ref{app:model_choise}.

\paragraph{Metrics.}
\textbf{(1) Acc.} (Accuracy, \%):
The percentage of valid queries whose final-turn plan satisfies all world and user constraints and passes the rubric threshold on all dimensions. 
Queries that terminate due to early stopping or reaching the maximum turn budget are counted as failures.
\textbf{(2) VPR} (Valid Plan Rate, \%):
The percentage of valid queries that terminate with a constraint-satisfying plan rather than in early stopping or max-turn stopping.
\textbf{(3) Avg Turns}:
The average number of interaction turns per instance.
\textbf{(4) AWRV} (Average World Repeated Violations):
The average number of repeated violations of disclosed world constraints per query.
\textbf{(5) AURV} (Average User Repeated Violations):
The average number of repeated violations of disclosed user constraints per query.
\textbf{(6) ATWC} (Average Triggered World Constraints):
The query-level ratio of triggered world constraints to interaction turns, averaged across queries.
\textbf{(7) ATUC} (Average Triggered User Constraints):
The query-level ratio of triggered user constraints to interaction turns, averaged across queries.
Detailed metric implementations are provided in Appendix~\ref{app:metrics}.
We discuss and provide prompts used in evaluation in Appendix~\ref{app:prompt-details} and Table~\ref{tab:all-prompts-router}. 

\begin{table*}[t]
\centering
\setlength{\extrarowheight}{2pt}
\resizebox{\linewidth}{!}{
\begin{tabular}
{lccccccc}
\toprule
\multirow{2}{*}{Model} & \multicolumn{3}{c}{Outcome} & \multicolumn{2}{c}{Constraint Failure} & \multicolumn{2}{c}{Constraint Elicitation} \\
\cmidrule(lr){2-4}\cmidrule(lr){5-6}\cmidrule(lr){7-8}
 & Acc. (\%) $\uparrow$ & VPR (\%) $\uparrow$ & Avg Turns & AWRV $\downarrow$ & AURV $\downarrow$ & ATWC & ATUC \\
\midrule
Qwen3-8B & 14.38 & 82.35 & 4.493 & 0.242 & 0.614 & 0.608 & 1.888 \\
Qwen3-14B & 17.26 & 73.62 & 4.785 & 0.296 & 0.821 & 0.668 & 2.042 \\
Qwen3-32B & 17.92 & 80.13 & 5.010 & 0.150 & 0.645 & 0.609 & 2.082 \\
Llama-3.3-70B-Instruct & 29.32 & 83.71 & 4.619 & \underline{0.114} & 0.537 & 0.668 & 1.830 \\
DeepSeek-v4-Flash & 35.53 & 76.97 & 6.385 & 0.464 & 0.895 & 0.977 & 2.657 \\
Gemini-3-Flash & 43.32 & \underline{90.23} & 5.824 & \textbf{0.065} & 0.391 & 0.756 & 2.442 \\
Gemini-3.1-Pro & 34.53 & \textbf{91.21} & 5.651 & 0.124 & \underline{0.251} & 0.769 & 2.236 \\
GPT-5 & \textbf{67.75} & 89.58 & 6.212 & 0.199 & \textbf{0.195} & \underline{1.191} & \underline{3.269} \\
GPT-5-Mini & \underline{61.89} & 85.34 & 5.886 & 0.322 & 0.322 & \textbf{1.318} & \textbf{3.391} \\
GPT-5-Nano & 42.35 & 67.75 & 5.541 & 0.971 & 0.355 & 1.089 & 2.468 \\
\bottomrule
\end{tabular}
}
\vspace{-0.1in}
\caption{\BENCH{} evaluation results under $\mathcal{E}_{mid}$. 
Scores in \textbf{bold} and \underline{underline} indicate the best and second-best performance, respectively.
Avg Turns is averaged over all instances, including early-stopped trajectories. We highlight the top two ATWC and ATUC values for comparison, although higher is not always better. Confidence intervals are in Appendix~\ref{app:CI}.}
\label{tab:main_table_2}
\vspace{-0.2in}
\end{table*}

\subsection{Results}

\paragraph{Current LLM Agents are still far from effective under dynamic dual constraints.}
The main results indicate that planning under progressively disclosed dual constraints remains challenging for all evaluated models. 
Even \textit{GPT-5}, the best-performing model, reaches only 67.75\% accuracy. 
Another strong model, \textit{Gemini-3.1-Pro}, scores only around 35\%, while most models fall below 45\%. 
Open-weight models perform particularly poorly, typically around 30\% or lower. 
Although most models maintain a valid-plan ratio above 70\%, they still frequently violate already disclosed constraints. 
Additionally, averaged across the 10 evaluated models, each query contains 0.295 repeated violations of disclosed world constraints and 0.503 repeated violations of disclosed user constraints.
As a result, 17.91\% of queries terminate early in average because the model violates disclosed constraints two consecutive times without triggering any new constraint. 
Together, these results indicate that progressively disclosed dual constraints remain a substantial challenge for current models, affecting both consistent constraint adherence and overall plan quality.

\paragraph{High valid-plan rates do not necessarily translate into final task success.}
A high valid plan ratio (VPR) does not guarantee strong end-task accuracy. 
For example, both \textit{Gemini-3.1-Pro} and \textit{Gemini-3-Flash} achieve relatively high VPRs, exceeding 90\%, yet their accuracies remain below 45\%. 
This pattern is further supported by their relatively low AWRV and AURV, suggesting that these models are reasonably effective at tracking disclosed constraints and avoiding repeated violations. 
However, despite maintaining executable plans and showing relatively strong constraint-tracking behavior, they still often fail to reach correct final solutions. 
This pattern suggests that relatively strong constraint tracking alone is insufficient to ensure strong final performance under progressive disclosure.

\paragraph{Better final performance is associated with stronger proactive constraint exploration.}
The results also suggest that stronger end-task performance is associated with stronger proactive exploration during interaction.
The highest-accuracy models, \textit{GPT-5} and \textit{GPT-5-Mini}, also exhibit the highest ATWC and ATUC values.
Across models, accuracy is strongly correlated with both metrics, with correlation coefficients of 0.898 for ATWC and 0.919 for ATUC.
One possible explanation is that higher ATWC and ATUC may reflect a greater capacity to generate more diverse plan revisions after blocking feedback, enabling the agent to propose new candidate plans under the currently disclosed constraints.

\paragraph{Conventional notions of model strength do not reliably predict adaptive planning capability.}
A final striking pattern is that models conventionally considered stronger do not always perform better in this setting. 
\textit{GPT-5-Mini} achieves accuracy comparable to \textit{GPT-5}, while \textit{Gemini-3-Flash} even surpasses \textit{Gemini-3.1-Pro}, despite the latter achieving the best VPR.
Among the open-weight Qwen3 models, \textit{Qwen3-8B}, \textit{Qwen3-14B}, and \textit{Qwen3-32B} perform similarly, with all three remaining at comparably low accuracy levels despite their substantial differences in scale.
This suggests that the capabilities required by our benchmark are not well captured by standard notions of model strength. 
More broadly, simple scaling in model size or general-purpose capability does not appear sufficient to deliver models' adaptiveness under progressively disclosed dual constraints.
We further provide error analysis in Section~\ref{sec:error-analysis}. 

%% file: latex/5_analysis.tex
\section{Analysis}
\label{sec:analysis}

\begin{takeaway}
Models are highly sensitive to increasing constraint burden.
\end{takeaway}

\begin{figure*}[t]
    \centering
    \includegraphics[width=\linewidth]{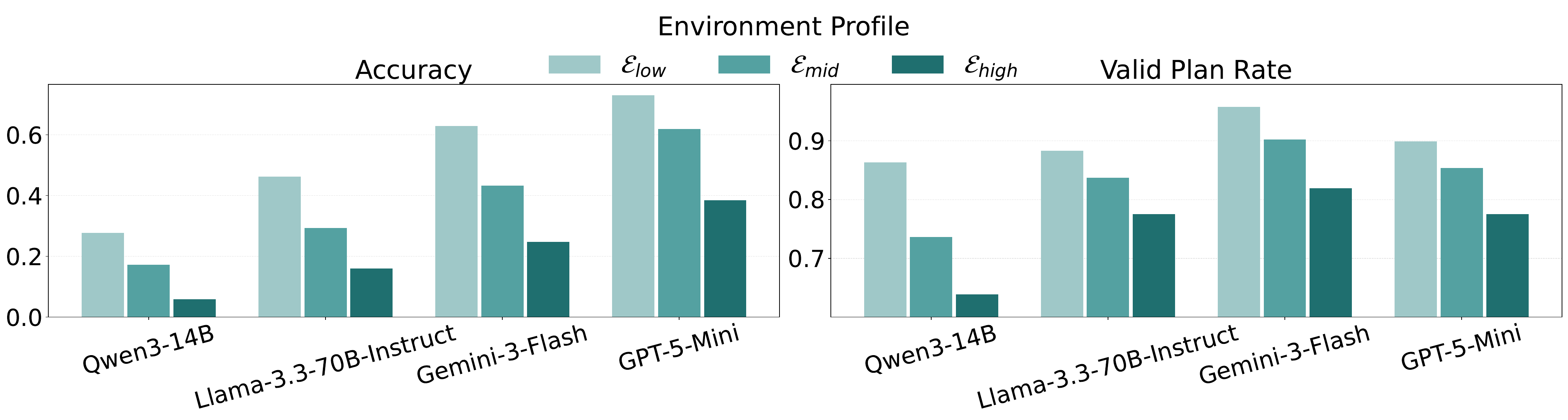}
    \vspace{-0.25in}
    \caption{Model performance under increasing constraint burden. Performance drops steadily as the environment profile becomes more constrained, suggesting that current models are highly sensitive to growing dual-constraint complexity. }
    \label{fig:analysis-iteration-bar}
    \vspace{-0.1in}
\end{figure*}

\paragraph{Model planning quality declines as total constraint complexity increases.}
To study how models respond to increasingly complex requirements, we construct three environment profiles, $\mathcal{E}_{low}$, $\mathcal{E}_{mid}$, and $\mathcal{E}_{high}$, by iteratively aggregating and validating world and user constraints (details in Section~\ref{sec:data-construction}). 
According to Figure~\ref{fig:analysis-iteration-bar}, both accuracy and valid plan rate exhibit a clear downward trend as the environment profile becomes more constrained across these settings.
This trend suggests that increasing constraint complexity degrades plan quality, making it harder for models both to reach the correct final answer and to maintain a valid plan throughout the interaction.


\paragraph{Models' performance deteriorates as progressively disclosed constraints accumulate within a trajectory.}

\begin{figure*}[t]
    \centering
    \includegraphics[width=\linewidth]{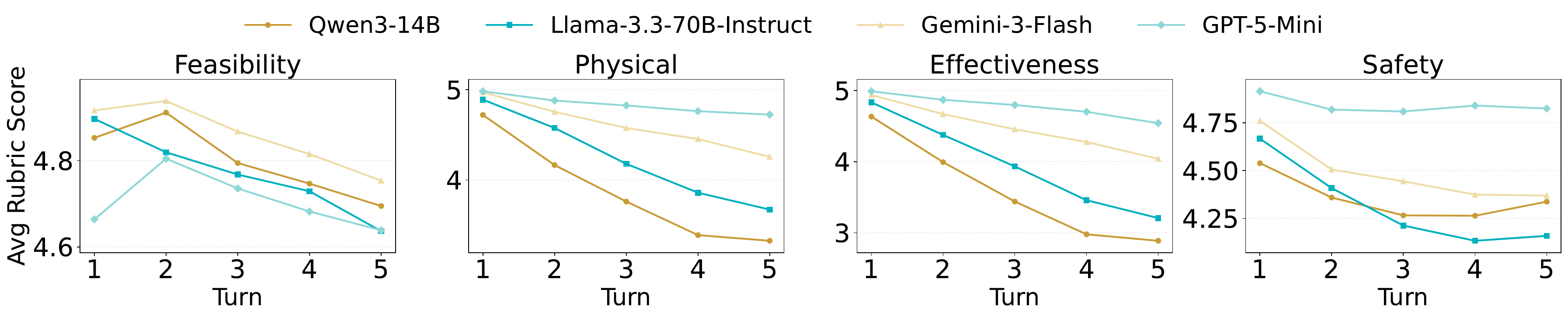}
    \vspace{-0.25in}
    \caption{Selected model rubric scores across interaction turns under $\mathcal{E}_{mid}$. Performance deteriorates as progressively disclosed constraints accumulate within a trajectory, indicating that models struggle to maintain stable planning quality over interactions.}
    \label{fig:analysis-turnwise}
    \vspace{-0.2in}
\end{figure*}

Beyond final-task success, an important question is whether model performance remains stable as additional constraints are revealed over the course of interaction. To examine this, we conduct a turn-wise rubric analysis that aligns trajectories by turn and tracks the average score of each model on four planning dimensions as progressively disclosed constraints accumulate. As shown in Figure~\ref{fig:analysis-turnwise}, performance generally declines over time across most dimensions, with pronounced drops in the selected metrics. This trend suggests that models struggle to maintain coherent and constraint-consistent planning once they must continuously incorporate newly revealed requirements into an existing plan. The degradation is substantially milder for stronger models, which remain comparatively stable on several dimensions, but the overall pattern is consistent: progressively disclosed constraints impose a growing burden on planning quality as the trajectory unfolds.


\begin{figure*}[t]
    \centering
    \includegraphics[width=\linewidth]{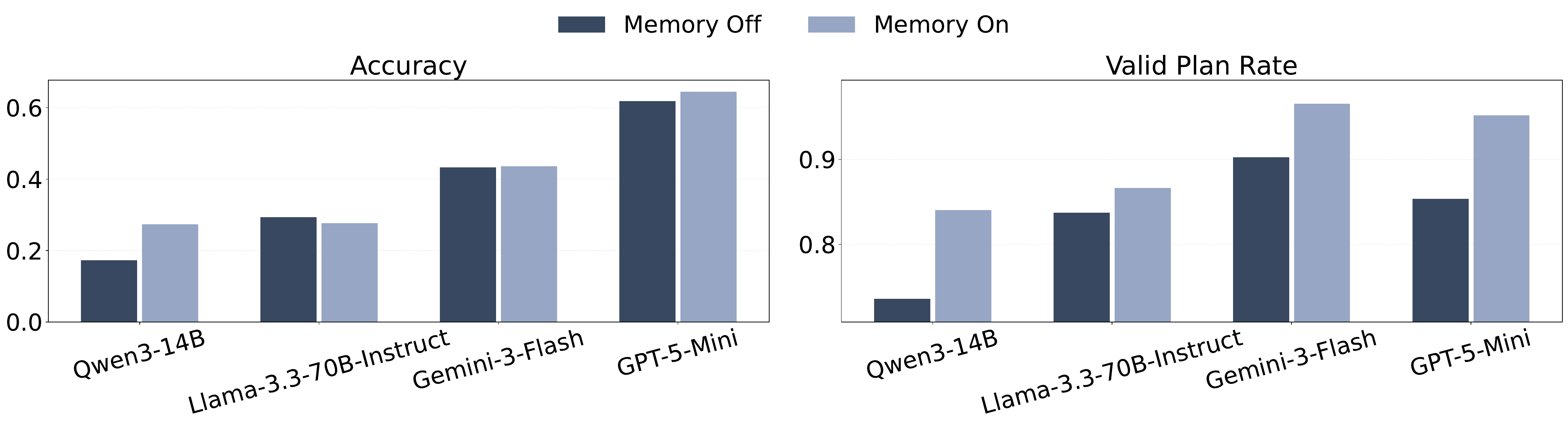}
    \vspace{-0.25in}
    \caption{Model performance under $\mathcal{E}_{mid}$ with additional constraint tracking module. Explicitly providing prior disclosed constraints brings only limited improvement on accuracy.}
    \label{fig:analysis-memory}
    \vspace{-0.15in}
\end{figure*}

\begin{takeaway}
\label{sec:diagnosis-analysis}
Model degradation is only partially mitigated by explicit constraint tracking and rubric feedback, while failures remain especially pronounced under user constraints.
\end{takeaway}

\begin{wraptable}{t}{0.55\columnwidth}
\centering
\small
\setlength{\tabcolsep}{2pt}
\setlength{\extrarowheight}{2pt}
\resizebox{\linewidth}{!}{
\begin{tabular}{lcccc}
\toprule
Model & Feasibility & Physical & Effectiveness & Safety \\
\midrule
Qwen3-8B & 4.758 & 3.478 & 2.956 & 4.446 \\
Qwen3-14B & 4.755 & 3.520 & 3.030 & 4.430 \\
Qwen3-32B & \textbf{4.785} & 3.500 & 3.087 & 4.454 \\
Llama-3.3-70B-Instruct & 4.729 & 3.815 & 3.236 & 4.410 \\
DeepSeek-v4-Flash & \underline{4.771} & 4.216 & 3.868 & 4.537 \\
Gemini-3-Flash & 4.760 & 4.276 & 4.004 & 4.457 \\
Gemini-3.1-Pro & 4.628 & 4.262 & 4.055 & 4.445 \\
GPT-5 & 4.550 & \textbf{4.685} & \textbf{4.570} & \underline{4.824} \\
GPT-5-Mini & 4.615 & \underline{4.622} & \underline{4.370} & \textbf{4.828} \\
GPT-5-Nano & 4.559 & 4.428 & 3.970 & 4.810 \\
\midrule
Average & 4.691 & 4.080 & 3.715 & 4.564 \\
\bottomrule
\end{tabular}
}
\caption{Models' performance under $\mathcal{E}_{mid}$ on four major rubric dimensions.
}
\label{tab:turnwise-final-selected}
\vspace{-0.15in}
\end{wraptable}

\paragraph{Instant constraint tracking module improves validity without recovering accuracy.}
As shown in the constraint failure analysis in Table~\ref{tab:main_table_2}, models often violate constraints that had already been disclosed and satisfied in earlier turns. 
This raises a natural diagnostic question: does the degradation primarily stem from failures to retain previously revealed constraints, or from difficulty constructing effective plans even when those constraints are available?
To probe this issue, we conduct an intervention in which all previously disclosed constraints are explicitly appended to the model input at every turn (details in Appendix~\ref{app:memory-analysis-details}). As shown in Figure~\ref{fig:analysis-memory}, this intervention yields only a marginal improvement in accuracy (less than 3\% for 3 out of 4 models), while producing a more noticeable gain in VPR, typically on the order of 5\%--15\%. 
These results suggest that explicit access to prior constraints helps improve constraint validity, but brings little benefit to final task success. 
Constraint tracking therefore alleviates the problem only partially.

\paragraph{Rubric-based feedback yields limited gains and destabilizes plans.}


\begin{figure*}[t]
    \centering
    \includegraphics[width=\linewidth]{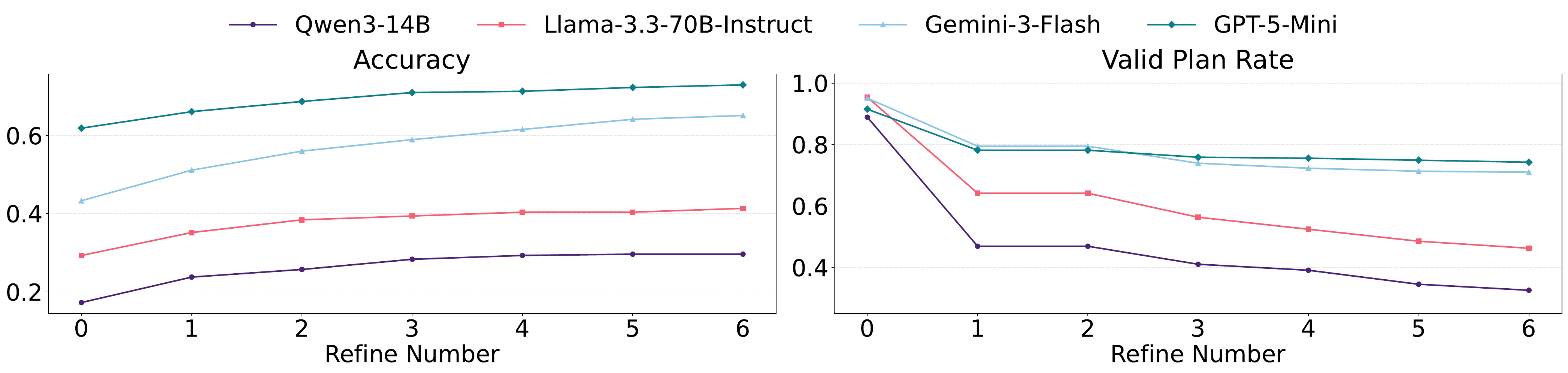}
    \vspace{-0.25in}
    \caption{Model performance under $\mathcal{E}_{mid}$ with rubric-based refinement. Additional feedback yields only modest recovery and often destabilizes planning.}
    \label{fig:analysis-rubrics-refinement}
    \vspace{-0.17in}
\end{figure*}

If making prior constraints explicit is insufficient to recover performance, we next ask whether feedback on failed planning dimensions can help the model revise its plan. 
To test this, we conduct a refinement analysis in which, for failed queries (i.e., cases with neither early stopping nor success), the model is given feedback on unsatisfied rubric dimensions and allowed to revise its plan (details in Appendix~\ref{app:refine-rubrics-analysis-details}).
As shown in Figure~\ref{fig:analysis-rubrics-refinement}, we allow 1--6 additional refinement turns and evaluate performance after each turn. 
Accuracy improves only modestly, by around 10\%, whereas VPR drops sharply, by roughly 40\% for the two open-source models and 20\% for the two proprietary models.
This suggests that rubric feedback can correct some local planning errors, but often at the cost of violating constraints that were previously satisfied. 
One possible explanation is that models exhibit a recency-biased adaptation pattern: when receiving new rubric-level feedback, they tend to prioritize repairing the newly identified weakness rather than preserving consistency with all previously disclosed constraints.
Thus, refinement guidance can improve some local rubric dimensions, but it does not reliably support globally consistent plan revision under accumulated constraints, and can substantially undermine constraint validity.

\begin{figure*}[t]
    \centering
    \includegraphics[width=\linewidth]{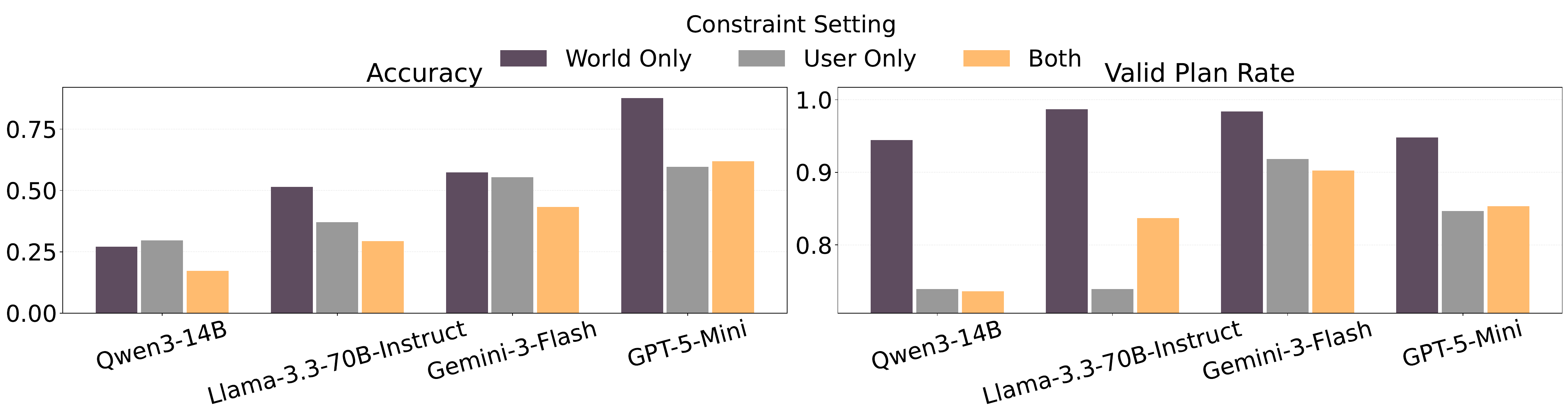}
    \vspace{-0.2in}
    \caption{Model performance under $\mathcal{E}_{mid}$ across constraint sources. User constraints cause larger degradation than world constraints, and dual-constraint setting is the hardest.}
    \label{fig:analysis-dual-ablation}
    \vspace{-0.2in}
\end{figure*}

\paragraph{User constraints contribute disproportionate difficulty.}
Since these corrective signals do not resolve the degradation, we next ask which constraints drive this difficulty. 
To isolate their effects, we perform a dual-ablation study with three conditions, \textit{World-Constraint Only}, \textit{User-Constraint Only}, and \textit{Both Constraints}, and evaluate each condition using accuracy and VPR.
As shown in Figure~\ref{fig:analysis-dual-ablation}, the pattern is clear across models: among the single-sided settings, \textit{User-Constraint Only} is consistently harder than \textit{World-Constraint Only}, while \textit{Both Constraints} is the most demanding setting.
This pattern suggests that user constraints contribute disproportionate difficulty in the dual-constraint setting and are a major source of planning instability. 
One possible reason is that user constraints often impose broader restrictions on the feasible action space: a single user preference may rule out many tools, actions, or modes of tool use, even when it appears as only one explicit constraint.


\begin{takeaway}
\label{sec:error-analysis}
\textbf{Error Analysis: }
Model failures primarily manifest as degraded task effectiveness and weak physical grounding under accumulated dual constraints.
\end{takeaway}


\paragraph{Task effectiveness degrades under accumulated dual constraints.}
As shown in Table~\ref{tab:turnwise-final-selected}, \textit{Effectiveness} is consistently one of the weakest dimensions across models. 
This weakness also becomes more pronounced over interaction, as Figure~\ref{fig:analysis-turnwise} shows a clear decline across turns and especially poor performance in the final stage. 
This trend suggests that current models struggle to maintain an effective plan under extended, constraint-heavy interaction.

\paragraph{Physical grounding remains weak under accumulated dual constraints.}
Table~\ref{tab:turnwise-final-selected} also shows consistently weak performance in \textbf{Physical} grounding. 
This suggests that models often fail to reason adequately about the physical consequences of their tool use under progressively disclosed constraints. 
As a result, their plans may appear coherent overall, while still overlooking key physical conditions such as object accessibility or spatial compatibility.
We provide more detailed examples in Appendix~\ref{app:error-example}.

%% file: latex/6_conclusion.tex
\section{Conclusion}

We introduce \bench{}, a dynamic interactive benchmark designed to evaluate LLM agents’ ability to adaptively plan and re-plan under progressively disclosed dual constraints from both the world and the user. 
By simulating open-ended household planning in which new constraints emerge only after plan violations, \bench{} captures a more realistic setting for studying adaptive planning under feedback. 
Our results reveal clear limitations of current LLM agents: while they can often produce plausible initial plans, they struggle to revise them adaptively as constraints accumulate, especially when user constraints are involved. 
These results show that reliable adaptation under dynamically evolving constraints remains a major challenge for current agents.
We see \bench{} as a foundational step toward future agents that are not only capable of planning, but truly adaptive and robust in dynamic environments.

%% file: latex/10_Appendix.tex
\appendix

\section{Related Works}
\label{app:related-works}


\paragraph{Evaluations on Agentic Planning Under Constraints.}
Planning under constraints is central to agentic decision making, and existing evaluations have studied constraints from either the world side or the user side.
Some benchmarks primarily focus on world-side constraints, such as PDDL constraints~\citep{PlanBench}, time/availability constraints~\citep{NATURAL-PLAN}, workflow rules~\citep{FlowBench} and API rules~\cite{appworld}, while others emphasize user-side constraints, such as preferences~\citep{PrefEval,RealPref}, personalization~\citep{PersonaMem-v1,PersonaMem-v2}, and user intent~\citep{UserBench,wang-etal-2024-user}. 
Some recent benchmarks have begun to incorporate both world-side and user-side constraints in interactive planning settings. 
CostBench~\citep{costbench} considers dual constraint types, but with upfront constraints, while FlowBench~\citep{FlowBench} remains largely workflow-centric and covers limited scope of constraints. 
Other related benchmarks model progressively elicited user preferences, but often assume limited action spaces~\citep{TravelPlanner,luo2025ultrahorizon} or lack scalable constraint construction~\citep{tau-bench}. 
More generally, in most existing settings, constraints are provided proactively by the environment rather than uncovered through the agent’s own exploration. 
Moreover, these settings do not emphasize iterative replanning, which fundamentally involves repeatedly collapsing the model’s current plan and requiring it to generate a new one. 
Consequently, they do not fully evaluate partially observed, open-ended adaptive planning under scalable dual constraints.

\paragraph{Agent Design for Constraint-based Planning.}

A parallel body of work develops methods for improving constraint-aware planning in LLM agents. 
On the world side, existing methods study state grounding~\citep{ReflAct}, localized violation correction~\citep{L-ICL}, experience-based world-model refinement~\citep{XENON}, plan-quality improvement through training~\citep{Plan-and-Act}, and formal constraint enforcement via symbolic planning~\citep{end-2-end-pddl}. 
On the user side, prior work mainly focuses on preference elicitation~\citep{userrl,TO-GATE} and proactive clarification or personalization during execution~\citep{Ask-before-Plan,sun2025trainingproactivepersonalizedllm,huang2025adactrl}. 
More recent approaches begin to handle world and user constraints jointly, especially in travel planning, through reflective prompting~\citep{MIRROR}, multi-agent coordination~\citep{ATLAS}, executable constraint checking~\citep{SCOPE}, and hierarchical control~\citep{HiMAP-Travel}. 
However, these methods largely assume that relevant constraints are available upfront, rather than emerging progressively during interaction. Moreover, they do not account for iterative interventions from the environment that continually disrupt the agent’s current plan and require repeated replanning. Consequently, it remains unclear how effectively current agents adapt when dual constraints must be discovered online through planning, failure, and ongoing revision.

\section{Formalization}
\label{app:formalization}

\subsection{Environment Construction Algorithm}
\label{app:env_cons_algorithm}

In this appendix, we present the full version of the data construction pipeline summarized in Section~\ref{sec:data-construction}. We construct each benchmark instance via a multi-agent framework, where different agents specialize in query rewriting and filtering, candidate plan proposal, constraint extraction, and constraint aggregation and validation.

For each retained instance, we associate a filtered household query with three hierarchical environment profiles:
\[
\left(
q,
\mathcal{E}_{\mathrm{low}},
\mathcal{E}_{\mathrm{mid}},
\mathcal{E}_{\mathrm{high}}
\right).
\]
Each profile contains one world-constraint set and one user-constraint set:
\[
\mathcal{E}_{\mathrm{low}}
=
\left(
\mathcal{B}_{w,\mathrm{low}},
\mathcal{B}_{u,\mathrm{low}}
\right),
\qquad
\mathcal{E}_{\mathrm{mid}}
=
\left(
\mathcal{B}_{w,\mathrm{mid}},
\mathcal{B}_{u,\mathrm{mid}}
\right),
\qquad
\mathcal{E}_{\mathrm{high}}
=
\left(
\mathcal{B}_{w,\mathrm{high}},
\mathcal{B}_{u,\mathrm{high}}
\right).
\]

We construct these three profiles through $R = 3$ iterative rounds of constraint induction. The profile produced after round $r = 1$ is identified with  $\mathcal{E}_{\mathrm{low}}$, the profile produced after round $r = 2$ is identified with $\mathcal{E}_{\mathrm{mid}}$, and the profile produced after round $r = 3$ is identified with $\mathcal{E}_{\mathrm{high}}$.

The pipeline uses the following role-specific models:
\[
\mathcal{M}_{\mathrm{rw}},
\quad
\mathcal{M}_{\mathrm{flt}},
\quad
\{\mathcal{M}^{(j)}_{\mathrm{plan}}\}_{j=1}^{J},
\quad
\mathcal{M}_{\mathrm{ext}},
\quad
\mathcal{M}_{\mathrm{merge}},
\quad
\mathcal{M}_{\mathrm{chk}}.
\]
These denote the query rewriter, binary query filter, planner samplers, constraint extractor, merge model, and constraint checker, respectively. 

\paragraph{Query rewriting and filtering.}
We start from raw queries from MacGyver~\citep{MacGyver} and use the query rewriter to produce short, method-agnostic household queries so as to broaden the downstream action space. Denoting the raw query by $q^{\mathrm{raw}}$, the rewritten query is
\[
q
=
\mathcal{M}_{\mathrm{rw}}
\left(
q^{\mathrm{raw}}
\right).
\]
The rewriter removes explicit resource constraints, such as \textit{tools available: ...} or \textit{using only ...}, while preserving the original task goal.
We then apply a strict binary filter and retain only concrete household tasks that require non-trivial planning:
\[
\mathcal{M}_{\mathrm{flt}}(q) = 1.
\]
Queries that correspond to single-step questions or otherwise do not require planning are removed. Since we only relax the original MacGyver resource constraints, we retain the corresponding reference solution $g$, and use it later during constraint extraction and validation to preserve solvability. Detailed filtering rules are described in Appendix~\ref{app:data_filtering_rule_exp} and illustrated in Figure~\ref{fig:filtering_rules}.

\paragraph{Iterative constraint sampling.}
To construct the hierarchical environment profiles, we iteratively induce constraints through multi-planner sampling. We employ a family of planner samplers and run the procedure for three rounds. The round index is $r \in \{1,2,3\}$.
The constraint type index is $x \in \{w,u\}$, where the two values correspond to world constraints and user constraints, respectively.
At each round, each planner independently generates multiple candidate plans via verbalized sampling~\citep{verbalized-sampling}. Specifically, for planner $j \in \{1,\dots,J\}$, we sample
\[
\{\pi^{(j)}_{x,r,k}\}_{k=1}^{K}
\sim
\mathcal{M}^{(j)}_{\mathrm{plan}}
\Bigl(
q
\mid
\tilde{\mathcal{B}}^{(j)}_{x,r-1}
\Bigr).
\]
Here, $\tilde{\mathcal{B}}^{(j)}_{x,r-1}$ is the accumulated planner-specific constraint pool before round $r$, and it is initialized as $\tilde{\mathcal{B}}^{(j)}_{x,0}
=
\emptyset$.
The world-side and user-side planner-specific pools are $\tilde{\mathcal{B}}^{(j)}_{w,r-1}$ and $\tilde{\mathcal{B}}^{(j)}_{u,r-1}$.
The former stores tools or environmental conditions that should no longer be used in subsequent rounds, while the latter stores user preferences that subsequent plans should avoid violating. The sampled plans are then analyzed to derive constraint candidates in two stages.

\paragraph{(i) Planner-wise constraint extraction and accumulation.}
In our setting, world constraints concern tool availability and usability in the environment, whereas user constraints concern whether the attributes associated with the tools used in a plan, or their implied usage, are acceptable to the user. 

For sampled world-side plans, we directly derive world-constraint candidates from each sampled plan:
\[
\mathcal{C}^{(j)}_{w,r,k}
=
\mathcal{M}_{\mathrm{ext}}
\Bigl(
\pi^{(j)}_{w,r,k},
q,
g
\Bigr).
\]
These candidates typically correspond to unavailable or unusable tools implicated by the sampled plan.

For user constraints, we first extract the tools involved in each sampled user-side plan:
\[
\mathcal{T}^{(j)}_{u,r,k}
=
\mathcal{M}_{\mathrm{ext}}
\Bigl(
\pi^{(j)}_{u,r,k}
\Bigr).
\]
We then condition the extractor on the query, the extracted tools, the full plan, and the reference solution, and ask it to infer user preferences over tool attributes or implied usages that would invalidate the sampled plan:
\[
\mathcal{C}^{(j)}_{u,r,k}
=
\mathcal{M}_{\mathrm{ext}}
\Bigl(
\pi^{(j)}_{u,r,k},
\mathcal{T}^{(j)}_{u,r,k},
q,
g
\Bigr).
\]

The per-plan candidates are then aggregated at the planner level:
\[
\mathcal{C}^{(j)}_{x,r}
=
\bigcup_{k=1}^{K}
\mathcal{C}^{(j)}_{x,r,k},
\qquad
x \in \{w,u\}.
\]


During extraction, the LLM-based extractor is given the sampled plans, the query, and the standard reference solution in order to preserve solvability. Concretely, we exclude query- or solution-specified objects when constructing world constraints, and we exclude user preferences that would invalidate the reference solution when constructing user constraints.

After extraction, we merge newly derived candidates with the previously accumulated planner-specific pool:
\[
\tilde{\mathcal{B}}^{(j)}_{x,r}
\leftarrow
\mathcal{M}_{\mathrm{merge}}
\Bigl(
\tilde{\mathcal{B}}^{(j)}_{x,r-1}
\cup
\mathcal{C}^{(j)}_{x,r}
\Bigr).
\]
This merge step canonicalizes and deduplicates constraints so that they remain consistent across rounds. As a result, the planner-specific pool at the current round both feeds back into later sampling and provides the planner-level basis for round-level aggregation and validation.

\paragraph{(ii) Round-level aggregation and validation.}
After all planners finish a round, we aggregate the planner-specific pools and validate the merged result. For each constraint type, we compute
\[
\mathcal{B}_{x,r}^{\star}
=
\mathcal{M}_{\mathrm{chk}}
\Biggl(
\mathcal{M}_{\mathrm{merge}}
\Biggl(
\bigcup_{j=1}^{J}
\tilde{\mathcal{B}}^{(j)}_{x,r}
\Biggr),
q,
g
\Biggr),
\qquad
x \in \{w,u\}.
\]

We then map the validated round outputs to the three hierarchical environment profiles:
\[
\mathcal{B}_{x,\mathrm{low}}
=
\mathcal{B}_{x,1}^{\star},
\qquad
\mathcal{B}_{x,\mathrm{mid}}
=
\mathcal{B}_{x,2}^{\star},
\qquad
\mathcal{B}_{x,\mathrm{high}}
=
\mathcal{B}_{x,3}^{\star},
\qquad
x \in \{w,u\}.
\]

Equivalently,
\[
\mathcal{E}_{\mathrm{low}}
=
\left(
\mathcal{B}_{w,1}^{\star},
\mathcal{B}_{u,1}^{\star}
\right),
\qquad
\mathcal{E}_{\mathrm{mid}}
=
\left(
\mathcal{B}_{w,2}^{\star},
\mathcal{B}_{u,2}^{\star}
\right),
\qquad
\mathcal{E}_{\mathrm{high}}
=
\left(
\mathcal{B}_{w,3}^{\star},
\mathcal{B}_{u,3}^{\star}
\right).
\]

The checker acts as a post-aggregation safeguard. It removes overly vague items, as well as any residual constraints that would invalidate the standard reference solution. For user constraints, we additionally remove preference sets that are internally contradictory or jointly exhaustive, since such combinations would leave no realizable preference-consistent solution. For example, a pair such as ``I dislike quiet atmosphere'' and ``I dislike noisy places'' is removed because it effectively rules out the entire relevant preference space.

Repeating this procedure across the three rounds yields the three hierarchical profiles
$\mathcal{E}_{\mathrm{low}}$, $\mathcal{E}_{\mathrm{mid}}$ and $\mathcal{E}_{\mathrm{high}}$ with progressively richer yet still self-consistent constraint sets. 

Our pipeline is designed for fair evaluation while preserving task solvability. We reduce bias toward any single planner by constructing diverse constraint profiles with multiple samplers across rounds. We further preserve solvability by retaining only constraints compatible with the reference solution and filtering vague, contradictory, or solution-invalidating items during validation. Overall, this yields diverse, consistent, and solvable environment profiles for adaptive re-planning.

\subsection{Intuition Behind}
\label{app:env_cons_intuition}

The goal of our constraint construction pipeline is to construct diverse, reasonable world and user constraint sets by iterative exploration of the solution space through multi-planner sampling, while keeping the problem solvable.

Our algorithm combines \textbf{parallel sampling} and \textbf{iterative sampling} because the two play complementary roles in constraint discovery.
Parallel sampling broadens exploration across planners. 
In each round, we use multiple planner samplers in parallel, and each sampler generates multiple candidate plans in one pass~\citep{verbalized-sampling}. 
This design increases \textbf{diversity} at two levels: different samplers may exhibit different planning tendencies, while multiple samples from the same sampler provide local variation around that planner's strategy. 
As a result, parallel sampling helps expose a broader set of candidate solution patterns and therefore a broader set of potential constraints.

However, parallel sampling alone is insufficient. 
If we only sample independently from the planners once, the discovered constraints are limited by the planners' initial solution tendencies, and many later-stage alternatives may remain unexplored. 
Iterative sampling addresses this limitation by feeding previously extracted constraints back into subsequent rounds. 
By conditioning future planning on the accumulated constraint pool, the algorithm discourages previously explored strategies and pushes the planners toward new feasible directions. 
This enables the system to progressively uncover additional constraints that would be missed by one-shot parallel exploration alone.

Within each round, this procedure is carried out separately for world and user constraints, i.e., for $x\in\{w,u\}$, followed by round-level aggregation and validation. 
In this sense, parallel sampling provides breadth through multi-planner exploration, while iterative sampling promotes continued diversification across rounds. 
Their combination yields richer and more representative environment profiles than either strategy alone.

\section{Experiment Details}
\label{app:exp-details}

\subsection{Experiment Setup}
\label{app:model-setup}

\paragraph{Models.}
For all models, we set the temperature to 0.0 and the maximum completion length to 16{,}000. 
For the GPT-5 series models (\textit{GPT-5}, \textit{GPT-5-mini}, and \textit{GPT-5-nano}), temperature is not user-configurable, so we used their default settings (temperature=1.0). 
To ensure the robustness of our results, we report the average of three runs for GPT-5 series models; the variation in accuracy across runs does not exceed 3\%.
For the open-source models, we ran all experiments on four NVIDIA H100 GPUs.

To further demonstrate the robustness of our conclusions, we conduct an additional temperature ablation study (Table~\ref{tab:temperature-ablation}). 
We observe that varying the decoding temperature has only a limited impact on performance: the differences between $T=0.0$ and $T=1.0$ remain within 3\% for both accuracy and valid plan rate across all models. 
In contrast, the performance gap between GPT-5 series models and other models is substantially larger (e.g., much over 3\% in accuracy), indicating that the observed improvements cannot be attributed to decoding choices. 
These results suggest that our conclusions are robust to temperature variations, and that the performance gains of stronger models reflect intrinsic capability differences rather than sensitivity to decoding settings.


\begin{table*}[t]
\centering

\begin{subtable}[t]{0.44\linewidth}
\centering
\small
\setlength{\tabcolsep}{4pt}
\setlength{\extrarowheight}{2pt}
\resizebox{\linewidth}{!}{
\begin{tabular}{lcccc}
\toprule
& \multicolumn{3}{c}{Temperature} & \\
\cmidrule(lr){2-4}
Model & 0.0 & 0.7 & 1.0 & $\Delta_{\max}$ \\
\midrule
Qwen3-14B & 17.26 & 18.30 & 18.75 & 1.49 \\
Llama-3.3-70B-Instruct & 29.32 & 28.31 & 30.65 & 2.34 \\
Gemini-3-Flash & 43.32 & 42.26 & 43.79 & 1.53 \\
\bottomrule
\end{tabular}
}
\caption{Accuracy}
\label{tab:temperature-ablation-accuracy}
\end{subtable}
\hspace{0.02\linewidth}
\begin{subtable}[t]{0.44\linewidth}
\centering
\small
\setlength{\tabcolsep}{4pt}
\setlength{\extrarowheight}{2pt}
\resizebox{\linewidth}{!}{
\begin{tabular}{lcccc}
\toprule
& \multicolumn{3}{c}{Temperature} & \\
\cmidrule(lr){2-4}
Model & 0.0 & 0.7 & 1.0 & $\Delta_{\max}$ \\
\midrule
Qwen3-14B & 73.62 & 73.28 & 74.11 & 0.83 \\
Llama-3.3-70B-Instruct & 83.71 & 81.09 & 84.23 & 3.14 \\
Gemini-3-Flash & 90.23 & 88.52 & 90.79 & 2.27 \\
\bottomrule
\end{tabular}
}
\caption{Valid Plan Rate}
\label{tab:temperature-ablation-vpr}
\end{subtable}

\caption{Ablation results under different decoding temperatures. $\Delta_{\max}$ denotes the maximum performance difference across temperatures for each model, computed as the difference between the largest and smallest values among the tested temperatures. }
\label{tab:temperature-ablation}
\end{table*}

\subsection{Model Choice}
\label{app:model_choise}

\paragraph{Model Choice in Data Construction.}
For the planner samplers introduced in Section~\ref{sec:data-construction}, we instantiate
$M_{\mathrm{plan}}^{(1)}$, $M_{\mathrm{plan}}^{(2)}$, and $M_{\mathrm{plan}}^{(3)}$
with \textit{GPT-4.1}, \textit{DeepSeek-V3.2}, and \textit{Qwen3.6-Flash},
respectively. To ensure data quality, we use a strong model, \textit{GPT-5.4}, as
$M_{\mathrm{chk}}$ to filter invalid constraints.

\paragraph{Judge-LLM Choice in Evaluation.}
For evaluation, we instantiate both the world-constraint judge and the user-constraint judge
with \textit{GPT-5.4}. 
For rubric-based evaluation, we use the same three models as in
data construction, namely \textit{GPT-4.1}, \textit{DeepSeek-V3.2}, and
\textit{Qwen3.6-Flash}, as independent judges.


\paragraph{Filter model and judge model validation.}
We validate $M_{\mathrm{chk}}$ used in data construction and the runtime LLM judges through human annotation. For $M_{\mathrm{chk}}$, we compare its filtering decisions on 30 sampled evaluation instances against majority-voted labels from three annotators. It filters 42.18\% of constraints on average, with a false negative rate of 2.31\% and a false positive rate of 3.72\%. For runtime judging, we annotate 30 sampled trajectories from 10 evaluated models, yielding 166 turn-level instances. The LLM judges achieve 89.76\% exact match with human majority labels, and differ by at most one constraint in 161 out of 166 turns. These results support the reliability of our filtering and evaluation pipeline.

\subsection{Runtime Interaction Details}
\label{app:runtime-interaction-details}

\paragraph{User feedback construction.}

\begin{figure*}[htbp]
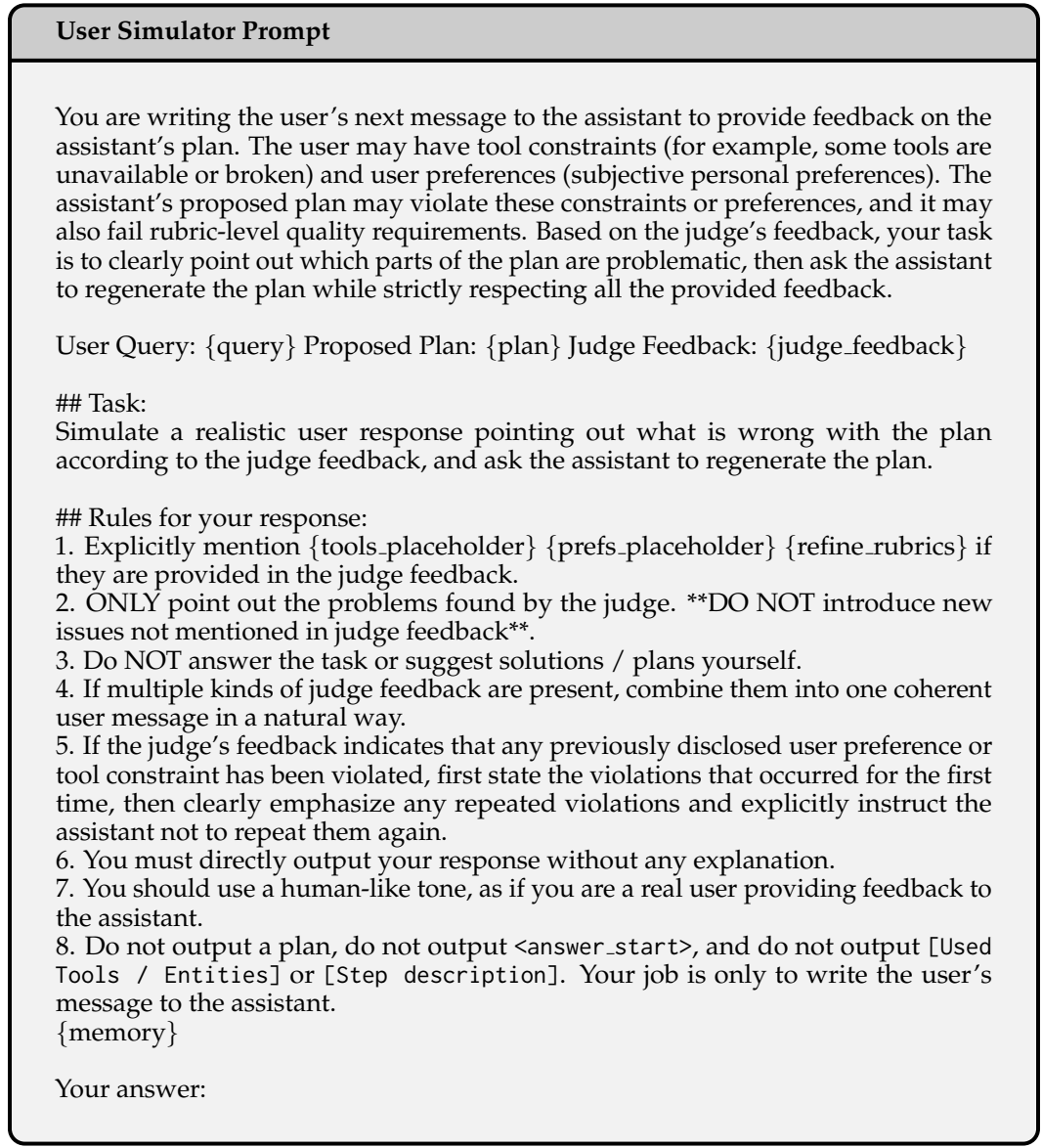

\begin{tcolorbox}[
    colback=black!5!white,
    colframe=black,
    title=User Simulator Prompt,
    fonttitle=\bfseries,
    colbacktitle=black!20!white,
    coltitle=black,
    boxrule=1.5pt,
    arc=5pt,
    boxsep=5pt,
    left=12pt,
    right=12pt,
    top=12pt,
    bottom=12pt
]

You are writing the user's next message to the assistant to provide feedback on the assistant's plan. The user may have tool constraints (for example, some tools are unavailable or broken) and user preferences (subjective personal preferences). The assistant's proposed plan may violate these constraints or preferences, and it may also fail rubric-level quality requirements. Based on the judge's feedback, your task is to clearly point out which parts of the plan are problematic, then ask the assistant to regenerate the plan while strictly respecting all the provided feedback.\\

User Query:
\{query\}
Proposed Plan:
\{plan\}
Judge Feedback:
\{judge\_feedback\}\\

\#\# Task:\\
Simulate a realistic user response pointing out what is wrong with the plan according to the judge feedback, and ask the assistant to regenerate the plan.\\

\#\# Rules for your response:\\
1. Explicitly mention \{tools\_placeholder\} \{prefs\_placeholder\} \{refine\_rubrics\} if they are provided in the judge feedback.\\
2. ONLY point out the problems found by the judge. **DO NOT introduce new issues not mentioned in judge feedback**.\\
3. Do NOT answer the task or suggest solutions / plans yourself.\\
4. If multiple kinds of judge feedback are present, combine them into one coherent user message in a natural way.\\
5. If the judge's feedback indicates that any previously disclosed user preference or tool constraint has been violated, first state the violations that occurred for the first time, then clearly emphasize any repeated violations and explicitly instruct the assistant not to repeat them again.\\
6. You must directly output your response without any explanation.\\
7. You should use a human-like tone, as if you are a real user providing feedback to the assistant.\\
8. Do not output a plan, do not output \texttt{<answer\_start>}, and do not output \texttt{[Used Tools / Entities]} or \texttt{[Step description]}. Your job is only to write the user's message to the assistant.\\
\{memory\}\\

Your answer:

\end{tcolorbox}
\caption{Prompt for generating user feedback on plan violations. The figure illustrates the instruction template used to simulate a user response based on judge feedback.}
\label{fig:user-llm-prompt}
\end{figure*}

At turn $t$, the agent proposes a plan $p_t$, and the judges identify the violated world constraints $V_t^w \subseteq \mathcal{B}_w$ and violated user constraints $V_t^u \subseteq \mathcal{B}_u$. 
The user feedback for the next turn is determined solely by these detected violations. 
Specifically, the revealed constraint set is selected from the judge-identified violations and then passed to a user simulator \(\mathcal{M}_{\mathrm{user}}\), which rewrites it into direct user feedback (see Figure~\ref{fig:user-llm-prompt} for the prompt).

Our feedback construction follows a single-type revelation rule. 
Even if $p_t$ violates constraints from both sides, we reveal constraints from only one type at a time, while including all violated items within that selected type. 
Formally, let the revealed constraint set at turn $t$ be denoted by $\widehat{V}_t$. We define
\[
\widehat{V}_t =
\begin{cases}
V_t^w, & \text{if } V_t^w \neq \emptyset,\\
V_t^u, & \text{if } V_t^w = \emptyset \text{ and } V_t^u \neq \emptyset,\\
\emptyset, & \text{if } V_t^w = \emptyset \text{ and } V_t^u = \emptyset.
\end{cases}
\]
That is, world-constraint violations are always prioritized over user-constraint violations: if $p_t$ violates only world constraints, we reveal $V_t^w$; if it violates both, we still reveal only $V_t^w$; and only when no world constraint is violated do we reveal $V_t^u$.

The intuition is that world constraints are typically more objective and directly verifiable~\citep{DBLP:reference/fai/2}. 
In our setting, a world constraint usually corresponds to a hard feasibility condition in the external environment, such as the unavailability of a required tool, material, or physical condition. 
If a plan depends on such an unavailable resource, then the plan is immediately infeasible in the real world. 

By contrast, user constraints are typically softer and more preference-based~\citep{CAMPIGOTTO2021103454}. 
They often reflect subjective priorities, dislikes, or comfort considerations, which are important but more negotiable than hard world-side feasibility conditions~\citep{ali2010goal,Hoang2012DistinguishingSA,qian2026creativitybench}. 
In real interactions, such preferences are also more likely to be adjusted or relaxed than hard environmental constraints~\citep{hauser2009non,narendhar2016different,ghafour2025smart}. 
For this reason, our feedback policy gives precedence to world constraints whenever both types are violated.

Operationally, once \(\widehat{V}_t\) is selected, it is passed to a user simulator \(\mathcal{M}_{\mathrm{user}}\), which rewrites the selected constraint items into explicit user feedback incorporated into the next-turn context. 
If the agent violates a previously revealed constraint again, \(\mathcal{M}_{\mathrm{user}}\) is instructed to explicitly remind the agent that the constraint has already been disclosed. 
Violations from the non-selected type are withheld until they become highest-priority under the same rule in a later turn.

This feedback-construction rule only determines which violated constraints are revealed to the agent, and does not change the underlying constraint checking. At each turn, the plan is still evaluated against the full constraint profile $\mathcal{E}=(\mathcal{B}_w,\mathcal{B}_u)$ to identify all actual violations and determine constraint satisfaction. However, repeated-violation metrics are computed over the disclosed constraint history: a violation is counted as repeated only if the same constraint has already been revealed in previous feedback. Therefore, the rule affects only feedback exposure during interaction, not the full constraint checking used for evaluation.

\paragraph{LLM rubrics judge details.}

\begin{table}[t]
\centering
\small
\setlength{\tabcolsep}{6pt}
\renewcommand{\arraystretch}{1.25}
\begin{tabular}{p{3cm} p{9.5cm}}
\hline
\textbf{Rubric} & \textbf{Definition} \\
\hline
Feasibility & Whether the plan relies on tools, materials, or resources that are realistically available in a typical household setting. \\
\hline
Physical plausibility & Whether the described actions are consistent with real-world physical laws and whether the proposed tool use would produce the claimed effects. \\
\hline
Logical step ordering & Whether the sequence of steps follows a sensible and workable order for completing the task in practice. \\
\hline
Effectiveness & Whether the plan, if carried out as written, would successfully achieve the intended goal. \\
\hline
Concreteness & Whether the plan provides specific, actionable instructions rather than vague or underspecified suggestions. \\
\hline
Safety & Whether carrying out the plan would avoid causing harm to human being. \\
\hline
Consequence awareness & Whether the plan anticipates likely damage or downstream consequences to the environment and addresses them appropriately. \\
\hline
Autonomy & Whether the plan can be carried out independently without requiring substantial outside assistance or services. \\
\hline
\end{tabular}
\caption{Definitions of evaluation rubrics.}
\label{tab:rubrics-details}
\end{table}

To evaluate plan quality beyond binary constraint satisfaction, we use a rubric-based judge that scores each plan along eight dimensions: \textit{feasibility}, \textit{physical plausibility}, \textit{logical step ordering}, \textit{effectiveness}, \textit{concreteness}, \textit{safety}, \textit{consequence awareness}, and \textit{autonomy}. The definitions of these dimensions are provided in Table~\ref{tab:rubrics-details}.

Our rubric design is informed by previous literature on home-environment task execution, risk-aware planning, and practical plan evaluation~\citep{centers2015home,OSHA3886,osha_otm_robot_safety,nrc2010_home_environment}.
Based on these prior principles, we assess whether a plan is executable in a realistic household setting, physically plausible, logically ordered, effective for accomplishing the intended goal, sufficiently concrete and actionable, safe to carry out, aware of likely downstream consequences, and executable without substantial outside assistance.

Each rubric judge assigns an integer score from 1 to 5 for every dimension, where higher scores indicate better plan quality. 
We use anchor descriptions for scores 1, 3, and 5, with scores 2 and 4 representing intermediate cases between adjacent anchors. The full scoring criteria and detailed examples are shown in Table~\ref{tab:rubric_scoring} and~\ref{tab:rubric_scoring_example}.
The same rubric dimensions and scoring scheme are applied to all models and all interaction turns.

\subsection{Metric Details}
\label{app:metrics}

For each instance $i \in \{1,\dots,N\} (N=307)$ under a fixed environment profile $r$, let the hidden dual-constraint profile be
\[
\mathcal{E}_i = (\mathcal{B}_i^w, \mathcal{B}_i^u),
\]
where $\mathcal{B}_i^w$ and $\mathcal{B}_i^u$ denote the world-constraint set and the
user-constraint set, respectively. For notational simplicity, we suppress the profile label ${low}/{mid}/{high}$.
At interaction turn $t$, the agent outputs a plan $p_{i,t}$. The plan is first evaluated by LLM judges, which produce the violated world constraints $V_{i,t}^w \subseteq \mathcal{B}_i^w$ and the violated user constraints $V_{i,t}^u \subseteq \mathcal{B}_i^u$.
The environment then converts these violations into feedback-disclosed constraints, denoted by $F_{i,t}^w \subseteq V_{i,t}^w$ and $F_{i,t}^u \subseteq V_{i,t}^u$, 
which are revealed to the agent through feedback after turn $t$ as discussed in Appendix~\ref{app:runtime-interaction-details}.

To formalize newly disclosed and repeated violations, we first define the set of previously
disclosed constraints before turn $t$ as
\begin{equation}
D_{i,t}^x = \bigcup_{k=1}^{t-1} F_{i,k}^x,
\qquad x \in \{w,u\}.
\end{equation}
Then the newly disclosed violations and repeated violations at turn $t$ are defined as
\begin{equation}
\mathrm{New}_{i,t}^x = F_{i,t}^x \setminus D_{i,t}^x,
\qquad
\mathrm{Rep}_{i,t}^x = F_{i,t}^x \cap D_{i,t}^x,
\qquad x \in \{w,u\}.
\end{equation}

For rubric evaluation, let $r_{i,t,d}^{(m)} \in \{1,2,3,4,5\}$ denote the score assigned by
rubric judge $m \in \{1,\dots,M\}$ on dimension $d \in \{1,\dots,D\}$ at turn $t$.
The aggregated score on dimension $d$ is
\begin{equation}
\bar{r}_{i,t,d} = \frac{1}{M}\sum_{m=1}^{M} r_{i,t,d}^{(m)}.
\end{equation}
A plan passes the rubric evaluation iff every aggregated dimension score is at least the
threshold $\gamma$:
\begin{equation}
\mathrm{RubPass}_{i,t}
=
\mathbb{I}\!\left[
\min_{d \in \{1,\dots,D\}} \bar{r}_{i,t,d} \ge \gamma
\right].
\end{equation}

We further define the constraint-validity indicator
\begin{equation}
\mathrm{ConPass}_{i,t}
=
\mathbb{I}\!\left[
V_{i,t}^w = \varnothing \;\wedge\; V_{i,t}^u = \varnothing
\right].
\end{equation}

Based on these definitions, the reported metrics are computed as follows.

\paragraph{Accuracy (Acc.)}
Accuracy requires that the terminal-turn plan both satisfies all world and user constraints
and passes the rubric threshold:
\begin{equation}
\mathrm{Acc}
=
\frac{1}{N}\sum_{i=1}^{N}
\mathbb{I}\!\left[
\mathrm{ConPass}_{i,T_i}=1
\;\wedge\;
\mathrm{RubPass}_{i,T_i}=1
\right].
\end{equation}

\paragraph{Valid Plan Rate (VPR)}
VPR measures the proportion of instances whose trajectory terminates with a valid plan:
\begin{equation}
\mathrm{VPR}
=
\frac{1}{N}\sum_{i=1}^{N}
\mathbb{I}\!\left[
\mathrm{ConPass}_{i,T_i}=1
\right].
\end{equation}

\paragraph{Average Turns (Avg Turns)}
The average number of interaction turns per instance is
\begin{equation}
\mathrm{AvgTurns}
=
\frac{1}{N}\sum_{i=1}^{N} T_i.
\end{equation}

\paragraph{Average World Repeated Violations (AWRV)}
For each instance, the total number of repeated violations of previously disclosed world constraints is
\begin{equation}
\mathrm{WRV}_i
=
\sum_{t=1}^{T_i} \left| \mathrm{Rep}_{i,t}^w \right|.
\end{equation}
The dataset-level average is
\begin{equation}
\mathrm{AWRV}
=
\frac{1}{N}\sum_{i=1}^{N} \mathrm{WRV}_i.
\end{equation}

\paragraph{Average User Repeated Violations (AURV)}
For each instance, the total number of repeated violations of previously disclosed user constraints is
\begin{equation}
\mathrm{URV}_i
=
\sum_{t=1}^{T_i} \left| \mathrm{Rep}_{i,t}^u \right|.
\end{equation}
The dataset-level average is
\begin{equation}
\mathrm{AURV}
=
\frac{1}{N}\sum_{i=1}^{N} \mathrm{URV}_i.
\end{equation}

\paragraph{Average Triggered World Constraints (ATWC)}
Let the total number of distinct world constraints disclosed during trajectory $i$ be
\begin{equation}
\mathrm{TWC}_i
=
\left|
\bigcup_{t=1}^{T_i} F_{i,t}^w
\right|.
\end{equation}
ATWC is defined as the trajectory-level total number of disclosed world constraints,
normalized by the number of turns, and then averaged over all instances:
\begin{equation}
\mathrm{ATWC}
=
\frac{1}{N}\sum_{i=1}^{N}
\frac{\mathrm{TWC}_i}{T_i}.
\end{equation}

\paragraph{Average Triggered User Constraints (ATUC)}
Let the total number of distinct user constraints disclosed during trajectory $i$ be
\begin{equation}
\mathrm{TUC}_i
=
\left|
\bigcup_{t=1}^{T_i} F_{i,t}^u
\right|.
\end{equation}
ATUC is defined as the trajectory-level total number of disclosed user constraints,
normalized by the number of turns, and then averaged over all instances:
\begin{equation}
\mathrm{ATUC}
=
\frac{1}{N}\sum_{i=1}^{N}
\frac{\mathrm{TUC}_i}{T_i}.
\end{equation}

\subsection{Prompt Details}
\label{app:prompt-details}

For completeness, we summarize the runtime prompts used during agent--environment
interaction in Table~\ref{tab:all-prompts-router}. These prompts cover the agent-facing
planning prompt, the world-constraint judge prompt, the user-constraint judge prompt,
and the rubric-based evaluation prompt. We provide each full prompt in the corresponding
figures for reproducibility.

\begin{table}[t]
\centering
\small
\begin{tabular}{
>{\raggedright\arraybackslash}p{0.18\linewidth}
>{\centering\arraybackslash}p{0.12\linewidth}
>{\raggedright\arraybackslash}p{0.62\linewidth}
}
\toprule
\textbf{Prompt Type} & \textbf{Figure} & \textbf{Description} \\
\midrule
\makecell[tl]{User Simulator\\prompt}
& Figure~\ref{fig:user-llm-prompt}
& The prompt is used by the user simulator to provide feedback to the agent each turn. \\
\midrule
\makecell[tl]{World-constraint\\judge prompt}
& Figure~\ref{fig:world-judge-prompt}
& The prompt used by the world-constraint judge to determine whether the proposed
plan violates any hidden or disclosed world constraints, such as unavailable tools,
objects, or environmental conditions. \\
\midrule
\makecell[tl]{User-constraint\\judge prompt}
& Figure~\ref{fig:user-judge-prompt}
& The prompt used by the user-constraint judge to determine whether the proposed
plan violates any user-side subjective requirements, preferences, or personal
restrictions. \\
\midrule
\makecell[tl]{Agent runtime\\prompt}
& Figure~\ref{fig:eval-prompt}
& The prompt shown to the agent at each interaction turn. It contains the user query,
the dialogue history, and the newly revealed feedback, and instructs the model to
generate a revised plan under the currently disclosed constraints. \\
\midrule
\makecell[tl]{Rubrics Judge\\prompt}
& Figure composition
& The prompt used by rubric judges to score plan quality across the rubric dimensions.
For the rubrics judge prompt, we assemble the scoring prompt by combining the rubric
definition table in Table~\ref{tab:rubric_scoring_example} and the scoring criteria in
Table~\ref{tab:rubric_scoring}, so that the judge can assign scores accordingly. \\
\bottomrule
\end{tabular}
\caption{Router table of runtime prompts used in \bench{}. Each row maps a runtime
prompt role to the corresponding figure that provides its full template. For the rubrics
judge prompt, the final scoring prompt is assembled using
Table~\ref{tab:rubric_scoring} and Table~\ref{tab:rubric_scoring_example} to guide
score assignment.}
\label{tab:all-prompts-router}
\vspace{-0.2in}
\end{table}

\subsection{Constraint Tracking Analysis Experiment Setup}
\label{app:memory-analysis-details}

To further analyze whether model degradation is related to failures to retain previously disclosed constraints, we conduct an additional constraint-tracking experiment, as described in Section~\ref{sec:analysis}.
In this experiment, we augment the agent with an external constraint tracking module.
The goal of this intervention is to better disentangle the effect of explicit constraint memory from the broader difficulty of planning under progressively disclosed dual constraints.
It serves as a controlled approximation that helps us assess the extent to which explicit constraint tracking can mitigate the observed degradation.

At each interaction turn $t$, the environment evaluates the agent's proposed plan and identifies the set of newly disclosed violated constraints.
Let
\[
\small
D_t = \bigcup_{i=1}^{t-1} \left( F_i^w \cup F_i^u \right)
\]
denote the set of all constraints that have been disclosed to the agent prior to turn $t$.
Here, $F_i^w$ and $F_i^u$ are the world-constraint and user-constraint sets newly disclosed by the environment at turn $i$, after evaluating the agent's response for that turn.

In the constraint-tracking condition, we explicitly serialize all constraints in $D_t$ into natural language and prepend them to the agent input at turn $t$ as an external memory block.
This memory block contains all previously disclosed constraints, regardless of whether they originate from world constraints or user constraints.

The resulting turn-$t$ agent input consists of three parts:
(1) the standard conversation history up to turn $t-1$,
(2) the user feedback provided at turn $t$ by the simulated user LLM in response to the agent output from turn $t-1$,
and (3) the current-turn external memory block that explicitly aggregates all constraints disclosed before turn $t$.

To control context length and avoid redundant duplication, we retain only the current-turn external memory block in the agent input.
When earlier turns are incorporated into the conversation history, any memory blocks attached to those turns are removed.
As a result, the dialogue history contains only the standard interaction content, rather than repeated historical copies of earlier memory blocks.
At turn $t$, the agent therefore receives a single up-to-date external memory block that summarizes all constraints disclosed before turn $t$.

Formally, let $H_{t-1}$ denote the standard interaction history with constraint hints available before turn $t$.
We construct a compressed history $\tilde{H}_{t-1}$ by removing memory blocks from all earlier turns.
The turn-$t$ input is then formed as
\[
\small
x_t = \bigl[\tilde{H}_{t-1};\, u_t;\, \mathcal{M}_{D_t}\bigr],
\]
where $u_t$ denotes the user feedback provided at turn $t$ by the simulated user LLM,
and $\mathcal{M}_{D_t}$ denotes the natural-language serialization of all constraints disclosed prior to turn $t$.
This design allows the module to function as an external constraint tracker while keeping the prompt length manageable~\citep{ha2026memguard}.

\begin{table*}[t]
\centering
\small
\setlength{\tabcolsep}{2pt}
\setlength{\extrarowheight}{2pt}
\resizebox{\linewidth}{!}{
\begin{tabular}{lcccccccc}
\toprule
Model & Feasibility & Physical & Ordering & Effectiveness & Concreteness & Safety & Consequence & Autonomy \\
\midrule
Qwen3-8B & 4.758 & 3.478 & 4.380 & 2.956 & 4.364 & 4.446 & 3.182 & 4.995 \\
Qwen3-14B & 4.755 & 3.520 & 4.473 & 3.030 & 4.490 & 4.430 & 3.300 & 4.988 \\
Qwen3-32B & \textbf{4.785} & 3.500 & 4.575 & 3.087 & 4.707 & 4.454 & 3.336 & \underline{4.997} \\
Llama-3.3-70B-Instruct & 4.729 & 3.815 & 4.640 & 3.236 & 4.412 & 4.410 & 3.578 & 4.967 \\
DeepSeek-v4-Flash & \underline{4.771} & 4.216 & 4.891 & 3.868 & 4.967 & 4.537 & 3.882 & \textbf{5.000} \\
Gemini-3-Flash & 4.760 & 4.276 & 4.903 & 4.004 & 4.982 & 4.457 & 3.873 & \textbf{5.000} \\
Gemini-3.1-Pro & 4.628 & 4.262 & 4.937 & 4.055 & \underline{4.984} & 4.445 & 3.678 & \textbf{5.000} \\
GPT-5 & 4.550 & \textbf{4.685} & \textbf{4.966} & \textbf{4.570} & \textbf{4.999} & \underline{4.824} & \underline{4.762} & \textbf{5.000} \\
GPT-5-Mini & 4.615 & \underline{4.622} & \underline{4.940} & \underline{4.370} & 4.975 & \textbf{4.828} & \textbf{4.784} & 4.985 \\
GPT-5-Nano & 4.559 & 4.428 & 4.830 & 3.970 & 4.825 & 4.810 & 4.452 & 4.896 \\
\midrule
Average & 4.691 & 4.080 & 4.753 & 3.715 & 4.771 & 4.564 & 3.882 & 4.983 \\
\bottomrule
\end{tabular}
}
\caption{Average rubric scores of all evaluated models on the full set of eight planning dimensions in \bench{}.}
\label{tab:all-rubric-scores}
\end{table*}

\subsection{Rubric Refinement Analysis Experiment Setup}
\label{app:refine-rubrics-analysis-details}

In the standard runtime interaction protocol, only violations of world constraints and user constraints are incorporated into the feedback. Once the agent produces a plan that satisfies all constraints, the trajectory terminates and the plan is directly evaluated without further modification.

In this analysis, we introduce an additional rubric-based refinement mechanism to examine whether the model can further improve planning quality when given explicit feedback on rubric failures.
Concretely, at turn $t$, the model $\mathcal{M}$ generates a plan $p_t$, which is evaluated by a set of rubric judges $\{\mathcal{M}^{(m)}_{\mathrm{judge,rub}}\}_{m=1}^{M}$. 
The judges produce an aggregated score vector over $D$ rubric dimensions. 
If all dimensions meet the threshold $\gamma$, the trajectory terminates as success. 
Otherwise, we identify the subset of failed dimensions together with their associated rationales.
We then construct a refinement feedback using the user simulator $\mathcal{M}_{\mathrm{user}}$, which takes as input the failed rubric dimensions and their reasons, and generates a natural language suggestion describing which aspects of the plan should be improved. 
This feedback is provided to the model, which produces a revised plan $p_{t+1}$ conditioned on the refinement signal.

The rubric refinement is applied only when the current plan satisfies all world and user constraints. 
If the plan violates any constraint, the feedback follows the standard priority rule described in Appendix~\ref{app:runtime-interaction-details}, where world-constraint violations are provided first, followed by user-constraint violations. 
In such cases, no rubric-based refinement feedback is given.

\subsection{Confidence Intervals}
\label{app:CI}

We report 95\% confidence intervals for model accuracy on the 307 evaluation samples in Table~\ref{tab:model_acc_ci}. 
Since each prediction is either correct or incorrect, accuracy is modeled as a binomial proportion. Let $\hat{p} = k/n$ denote the empirical accuracy, where $k$ is the number of correct predictions and $n=307$ is the total number of evaluated samples. 
We use the two-sided Wald confidence interval~\citep{kahouadji2025comprehensivecomparisonwaldwilson}:
\begin{equation}
\hat{p} \pm 1.96 \sqrt{\frac{\hat{p}(1-\hat{p})}{n}}.
\end{equation}
We report all intervals in percentage form in the tables, i.e., after multiplying both $\hat{p}$ and the interval bounds by 100.

\section{Additional Experiment Results}
\label{app:additional-experiment-results}

\subsection{Full Rubric Scores}
\label{app:full-rubric-scores}

Table~\ref{tab:all-rubric-scores} reports the complete rubric results for all eight planning dimensions, complementing the four selected dimensions discussed in the main text. 
In particular, beyond feasibility, physical plausibility, effectiveness, and safety, we additionally report logical step ordering, concreteness, consequence awareness, and autonomy; the detailed definitions of all rubric dimensions are provided in Table~\ref{tab:rubrics-details}. 
Overall, these additional results are broadly consistent with the main-text findings: logical step ordering, concreteness, and autonomy remain relatively strong across models, while consequence awareness is somewhat lower on average and may also contribute to some failures in constraint-heavy settings. 

\subsection{Discussion on Parameter Choice}
\label{app:discuss-parameter-choice}

\subsubsection{Max Turns Threshold $T$}
\label{app:max-turns-threshold-ablation}

We set the maximum turn budget to $T=20$ as a conservative upper bound rather than a practically active bottleneck. 
In our main evaluation, trajectories in Table~\ref{tab:main_table_2} almost never reached the maximum-turn limit: only \textit{GPT-5} showed a single such case ($1/307$). This pattern is also consistent with the average interaction length in Table~3, where all models use only around 4.7--6.2 turns on average, far below the budget. Therefore, $T=20$ mainly serves as a safeguard against pathological long-horizon loops, while the actual stopping behavior in our benchmark is dominated by successful completion or the early-stopping rule rather than by exhausting the turn budget.

\begin{wraptable}{r}{0.42\textwidth}
\vspace{-0.5em}
\centering
\small
\begin{tabular}{lc}
\toprule
Model & Acc. (\%) $\uparrow$ \\
\midrule
Qwen3-8B & 14.38 $\pm$ 3.93 \\
Qwen3-14B & 17.26 $\pm$ 4.23 \\
Qwen3-32B & 17.92 $\pm$ 4.29 \\
Llama-3.3-70B-Instruct & 29.32 $\pm$ 5.09 \\
DeepSeek-v4-Flash & 35.53 $\pm$ 5.38 \\
Gemini-3-Flash & 43.32 $\pm$ 5.54 \\
Gemini-3.1-Pro & 34.53 $\pm$ 5.32 \\
GPT-5 & 67.75 $\pm$ 5.23 \\
GPT-5-Mini & 61.89 $\pm$ 5.43 \\
GPT-5-Nano & 42.35 $\pm$ 5.53 \\
\bottomrule
\end{tabular}
\vspace{-0.5em}
\caption{Accuracy with 95\% Wald confidence intervals on 307 samples.}
\label{tab:model_acc_ci}
\vspace{-1em}
\end{wraptable}

\subsubsection{Early Stop Threshold $\tau$}
\label{app:early-stop-threshold-ablation}

We set the early stopping patience to $\tau = 2$ to detect stagnation during interaction. 
In our setting, early stopping is triggered when the agent fails to violate any new constraint for $\tau$ consecutive turns. 
Intuitively, this situation indicates that the agent is repeatedly violating previously disclosed constraints without effectively exploring new parts of the constraint space. 
As a result, further interaction is unlikely to produce meaningful progress, which is consistent with prior work that adopts similar patience-based mechanisms~\citep{UserBench}.

It is important to note that $\tau$ does not directly affect whether a model is capable of finding a valid plan, but rather controls how long unproductive interaction is allowed to continue. 
Smaller values of $\tau$ may terminate some trajectories earlier, while larger values mainly increase interaction length without substantially improving final success rates.

Empirically, we observe that such stagnation patterns are common across models, with a non-trivial fraction of trajectories terminating due to repeated violations without triggering new constraints. 
This observation supports the necessity of an early stopping mechanism to prevent inefficient interaction.

\subsubsection{Rubrics Pass Threshold $\gamma$}
\label{app:rubrics-pass-threshold-ablation}

\begin{table*}[t]
\centering
\small
\setlength{\tabcolsep}{4pt}
\setlength{\extrarowheight}{2pt}
\resizebox{0.8\linewidth}{!}{
\begin{tabular}{lccccccc}
\toprule
Model & $\gamma=3.00$ & $\gamma=3.33$ & $\gamma=3.66$ & $\gamma=4.00$ & $\gamma=4.33$ & $\gamma=4.66$ & $\gamma=5.00$ \\
\midrule
Qwen3-8B & 35.62 & 24.51 & 19.28 & 14.38 & 9.15 & 4.90 & 1.31 \\
Qwen3-14B & 39.74 & 32.57 & 26.38 & 17.26 & 10.10 & 5.54 & 0.65 \\
Qwen3-32B & 39.41 & 33.88 & 26.06 & 17.92 & 12.38 & 4.56 & 2.93 \\
Llama-3.3-70B-Instruct & 45.60 & 41.04 & 33.88 & 29.32 & 22.48 & 12.70 & 3.91 \\
DeepSeek-v4-Flash & 57.00 & 52.44 & 43.97 & 35.53 & 26.06 & 14.66 & 3.26 \\
Gemini-3-Flash & 69.06 & 61.56 & 55.05 & 43.32 & 28.99 & 15.96 & 5.54 \\
Gemini-3.1-Pro & 68.73 & 57.65 & 49.84 & 34.53 & 24.10 & 10.42 & 3.26 \\
GPT-5 & \textbf{84.69} & \textbf{83.06} & \textbf{77.52} & \textbf{67.75} & \textbf{53.75} & \underline{37.46} & \underline{14.66} \\
GPT-5-Mini & \underline{77.20} & \underline{73.62} & \underline{68.73} & \underline{61.89} & \underline{52.44} & \textbf{38.76} & \textbf{20.85} \\
GPT-5-Nano & 54.40 & 52.44 & 47.88 & 42.35 & 32.57 & 20.52 & 9.45 \\
\bottomrule
\end{tabular}
}
\caption{Rubric threshold ($\gamma$) ablation on accuracy. The experimental setting is the same as in Table~\ref{tab:main_table_2}. We report model accuracy under different rubric pass threshold values and show that the trend observed in Table~\ref{tab:main_table_2} remains consistent across thresholds.}
\label{tab:rubric-threshold-ablation-accuracy}
\end{table*}

We further vary the rubric threshold by setting $\gamma \in \{3.00, 3.33, 3.66, 4.00, 4.33, 4.66, 5.00\}$ to examine whether our conclusions depend on the specific choice of evaluation strictness. 
As shown in Table~\ref{tab:rubric-threshold-ablation-accuracy}, accuracy consistently decreases as $\gamma$ increases, since a stricter threshold requires plans to satisfy all rubric dimensions at a higher level. 
More importantly, the relative ordering of model accuracy remains largely unchanged across different $\gamma$ values. 
This indicates that our main conclusions do not depend on a particular threshold choice and are robust to the selection of $\gamma$.

\section{Discussion}
\label{app:discussion}

\subsection{Benchmark Traits Elaboration}
\label{app:traits-exp}

Table~\ref{tab:comparison-table} compares \bench{} with prior benchmarks along seven benchmark traits that we view as particularly important for evaluating adaptive planning agents in realistic interactive settings with dual constraints.
Below, we elaborate on why each trait matters.

\paragraph{Iterative Re-planning.}
Real-world agentic tasks rarely end after a single-shot plan. As interaction unfolds, agents often need to revise earlier decisions in response to newly surfaced requirements~\citep{tau-bench} or changing conditions~\citep{costbench}. 
A benchmark with iterative re-planning therefore evaluates whether an agent can adapt rather than merely produce an initially plausible plan. 
This trait is crucial in real-world settings, where conditions and requirements are constantly changing, requiring plans to be continuously revised~\citep{xiao2025dynamictheorymindevaluating}.

\paragraph{User Interaction.}
User interaction is necessary under dual constraints because part of the relevant constraints comes from the user side~\citep{FlowBench}. 
Since these constraints are progressively revealed during interaction, benchmarks must model interaction with the user rather than treating user requirements as fully specified upfront.

\paragraph{World Interaction.}
World interaction is equally necessary under dual constraints because another part of the relevant constraints comes from the external world, such as environmental conditions and resource limitations~\citep{PlanBench}. 
Since these constraints are also progressively  revealed during interaction, benchmarks must model interaction with the world rather than assuming a fully observed environment from the start.

\paragraph{Dual Constraint.}
In realistic tasks, user constraints and world constraints co-exist~\citep{wang2024apricotactivepreferencelearning,silver2025coloringlinespersonalizationnull}.
An agent need to satisfy subjective preferences while simultaneously respecting objective environmental limitations. 
Modeling only one side gives an incomplete picture of planning adaptiveness. 
The dual-constraint trait is therefore important because it captures the need to jointly reason about what the user wants and what the world allows, which is a defining challenge in many real-world planning scenarios.

\paragraph{Progressive Disclosure.}
Progressive disclosure is important because in real-world settings, constraints are often not specified all at once, but instead emerge gradually as interaction unfolds~\citep{tau-2-bench,RealPref}. 
A benchmark with progressive disclosure therefore better reflects how planning happens in practice, where agents must continually adapt to requirements that become visible only over time.

\paragraph{Open-Ended Evaluation.}
For complex planning tasks, there are often many valid ways to succeed. 
Restricting evaluation to a single or limited reference trajectory can therefore understate agent capability and over-reward surface imitation. 
Open-ended evaluation is important because it allows any plan that satisfies the task goal and constraints to be considered valid, making the benchmark better aligned with the inherently diverse solution space of real-world planning problems.

\paragraph{Scalable Constraints.}
A useful benchmark should not only measure current performance, but also support controlled difficulty scaling as models improve. 
Scalable constraints are important because they make it possible to systematically vary the complexity of the planning environment, for example by increasing the number of constraints.
This enables more fine-grained analysis of where and how models break down, and improves the benchmark's long-term utility.

Overall, these seven traits capture important aspects of planning in realistic interactive environments. \bench{} covers all of them, allowing evaluation under settings that are closer to real-world planning, where agents must interact with both users and the world, handle progressively revealed constraints, and remain adaptive in an open-ended and scalable environment.

\subsection{Data Filtering Rules}
\label{app:data_filtering_rule_exp}

Our query filtering design is primarily motivated by evaluation: we aim to retain tasks that genuinely test iterative household planning, while excluding problem types whose success depends on factors outside the target capability. 
As described in Section~\ref{sec:data-construction}, we first rewrite raw queries into short, method-agnostic household tasks to broaden the action space, and then apply a strict binary filter to keep only concrete household tasks while rejecting non-planning or overly simple questions. 
Figure~\ref{fig:filtering_rules} operationalizes these rules by excluding queries that are non-actionable, vague, externally delegated, or tied to a specific prescribed method or tool.

First, we filter out \textbf{knowledge-centric questions} such as explanation, factoid, or feasibility queries (e.g., asking why something happens, how something works, or whether an action is safe), because these do not primarily evaluate the capability we target: planning how to accomplish a task. 
Such queries place much more weight on domain knowledge or verbal explanation than on constructing and revising an actionable plan, which makes them a poor fit for our benchmark objective and potentially less fair for comparing models with different knowledge strengths. 
Our benchmark instead focuses on tasks that require the agent to decide what to do and how to adapt when constraints are progressively disclosed.

Second, we exclude queries that can be trivially resolved by \textbf{seeking outside help}, such as purchasing items, calling someone, or arranging external services. 
The reason is not merely domain restriction, but benchmark validity: if unrestricted external delegation is allowed, then many tasks admit a degenerate strategy in which the model avoids substantive planning by offloading the problem. 
This would create an artificial shortcut that weakens the benchmark's ability to measure re-planning under accumulating constraints. 
Figure~\ref{fig:filtering_rules} therefore explicitly disallows tasks requiring external assistance, while still keeping tasks realistic and grounded in everyday situations that could plausibly occur at home. 
This realism requirement helps the benchmark better reflect the kinds of planning problems that arise in practice.

Third, we intentionally avoid queries that \textbf{prescribe a specific method or tool}. 
A central goal of \bench{} is to evaluate whether agents can iteratively re-plan in an open-ended setting with a large, effectively unbounded action and solution space. 
If the query already fixes the method (for example, by explicitly requiring one tool), then much of that openness disappears: the task becomes closer to following instructions than to exploring alternative valid strategies under newly revealed constraints.
This is why our rewriting step removes explicit resource constraints to broaden the action space, and our filter further rejects queries that explicitly require a particular tool or method. 
Preserving this open action space is important because it creates room for meaningful plan revision when new world or user constraints are disclosed.

Fourth, we require \textbf{task goals to be sufficiently concrete and well specified}. 
If the goal is under-specified, then the plan effectiveness becomes difficult to evaluate reliably: different plans may optimize for different implicit interpretations of the task, and failures may reflect ambiguity in the query rather than deficiencies in planning. 
Since our runtime evaluation checks whether a plan satisfies constraints and also meets rubric-level planning quality requirements, unclear task goals would make both automatic judging and final success assessment substantially noisier. 
We therefore exclude vague queries whose intended outcome is not concrete enough to support consistent evaluation.

Finally, we filter out decoration- or aesthetics-heavy queries because they are less about accomplishing a concrete household task and more about \textbf{subjective stylistic preference}. 
While such tasks may still be realistic, they are difficult to evaluate in a stable and objective way, especially when the benchmark is designed around actionable planning, constraint satisfaction, and iterative revision. 
In contrast, concrete household tasks are better aligned with rubric-based evaluation and make it easier to determine whether a plan is effective, feasible, and physically grounded. 
Taken together, these filtering choices are intended to maximize evaluability while preserving realistic, open-ended planning scenarios in which iterative re-planning is both necessary and meaningfully testable.

\subsection{Constraints Extraction Standard}
\label{app:user-attribute-discussion}

Our constraint extraction standard follows the nature of household planning. 
In this setting, the agent mainly acts as a household-task executor whose interaction with the environment is mediated through tools, and its effective action space is therefore largely characterized by what tools it can choose and use~\citep{shridhar2020alfredbenchmarkinterpretinggrounded,VirtualHome}.
From this perspective, the most direct and enforceable form of world constraint is to block tool access itself, since world-side restrictions concern the objective environment and can be naturally expressed as whether a particular object or tool is available or usable~\citep{Birr_2024,ahn2022icanisay}. 
By contrast, user-side constraints are less about the objective existence of an object and more about how the user evaluates possible interactions with that object~\citep{narcomey2024learninghumanpreferencesrobot,Gerevini2005PlanCA,guo2025mathematical}. 
A user typically does not object to an object in isolation, but to attributes or implied modes of use such as being sharp, high-heat, noisy or messy~\citep{10.1145/3472208,Story2022}. 
We therefore elicit user preferences in an attribute-based manner and world constraints in an object-based manner. 
This separation gives us a simple and grounded operationalization of dual constraints in household planning: tool binding serves as the world-side constraint, while preferences over tool or action attributes serve as the user-side constraint. 
Although this abstraction does not cover the full richness of real-world preferences, it preserves the key distinction we aim to evaluate, namely that world constraints restrict what can be done in the environment, whereas user preferences restrict which forms of interaction with the environment are acceptable to the user. 
This design also makes constraint violations more verifiable during evaluation, since each constraint is defined with a relatively clear boundary and can therefore be checked against a proposed plan more consistently.

\subsection{Early Stop Mechanism}
\label{app:early-stop-discussion}

Our early-stopping mechanism follows the general convention of progressive disclosure in prior interactive benchmarks with progressively elicited constraints, such as UserBench~\citep{UserBench}. Concretely, if the agent fails to trigger any previously undisclosed constraint for $\tau=2$ consecutive turns, while also failing to terminate the trajectory with a valid plan, we stop the interaction and mark the trajectory as unsuccessful.

The intuition is that this pattern strongly suggests the agent has entered a locally repetitive mode rather than continuing to make meaningful progress. If no new constraint is elicited across three consecutive turns and the trajectory still does not terminate, then the agent is neither reaching a valid solution nor exploring parts of the environment that expose additional hidden requirements. Instead, it is repeatedly proposing plans that remain inconsistent with already disclosed world or user constraints. This indicates a lack of progress on both environment exploration and constraint-aware planning, including both world modeling and user modeling in the operational sense of tracking and incorporating revealed constraints. Under this condition, continuing the interaction is unlikely to reveal new behavior beyond repeated violations of known requirements. We therefore treat such trajectories as failed cases rather than prolonging an unproductive loop.

\subsection{Significance of \bench{}}
\label{app:discussion-significance-generalization}

\paragraph{Benchmark scope.}
This benchmark is designed to study \emph{adaptive planning} as a distinct capability. In real-world embodied settings, successful task completion depends on multiple components, including visual perception, grounding, navigation, low-level control, and physical interaction~\citep{shridhar2020alfredbenchmarkinterpretinggrounded, ahn2022icanisay,srivastava2022behavior,liu-etal-2025-revisiting}.
While these components are all essential for end-to-end agents, evaluating them jointly makes it difficult to determine whether a failure arises from deficient planning or from errors in perception, execution, or exploration in an embodied environment~\citep{chang2024partnrbenchmarkplanningreasoning,bhatt2025knowyoureuncertainplanning}. 
For this reason, we do not instantiate the benchmark in a fully embodied setting. Instead, we intentionally isolate the planning component, with the aim of measuring whether a model can revise an initially plausible plan once previously unmodeled constraints become relevant.

\paragraph{Abstraction of environment interaction.}
Our task formulation abstracts away from a fully specified observation--action loop. Rather than assuming that the agent begins with a complete observation of the environment, we consider a setting in which the agent is given a goal, forms an initial plan based on prior knowledge, and only subsequently discovers whether the assumptions underlying that plan actually hold~\citep{HANHEIDE2017119,pmlr-v87-nyga18a}. 
This abstraction reflects a common pattern in everyday task solving: agents often first reason about what tools, resources, or actions are likely needed for a task, and only then verify whether these assumptions are valid in the current environment. In a fully embodied setting, verifying such world or user constraints would typically require navigation, scene exploration, and additional perceptual inference~\citep{shridhar2020alfredbenchmarkinterpretinggrounded,kim2024realfredembodiedinstructionfollowing,liu2026naacl}, which would shift the evaluation away from planning and toward embodied interaction. We therefore omit explicit observations not because such information is unimportant in real-world deployment, but because abstracting away from these factors allows us to focus on planning under incomplete task-relevant information. This formulation also captures a form of proactiveness: rather than passively relying on the system to provide all necessary information upfront, the agent is expected to actively probe the environment through interaction, using these attempts to explore hidden constraints and to build an implicit model of the world and the user's preferences~\citep{narcomey2024learninghumanpreferencesrobot,sadigh2016information}.

\paragraph{Constraint revelation as a source of re-planning.}
The central challenge in our setting is that important constraints are not fully specified at the outset. 
These constraints may correspond to missing tools, unavailable resources, or user-specific preferences that invalidate an otherwise reasonable plan. 
Such situations are common in real-world household tasks, where agents must routinely revise plans~\citep{costbench} when assumptions about the environment turn out to be false. By representing these factors as reactively disclosed constraints, the benchmark captures a practically important aspect of agentic problem solving: not merely producing a plan, but adapting it when latent assumptions are contradicted.

\section{Human Annotation}
\label{app:human-annotation}

\begin{figure*}[t]
    \centering
    \includegraphics[width=\linewidth]{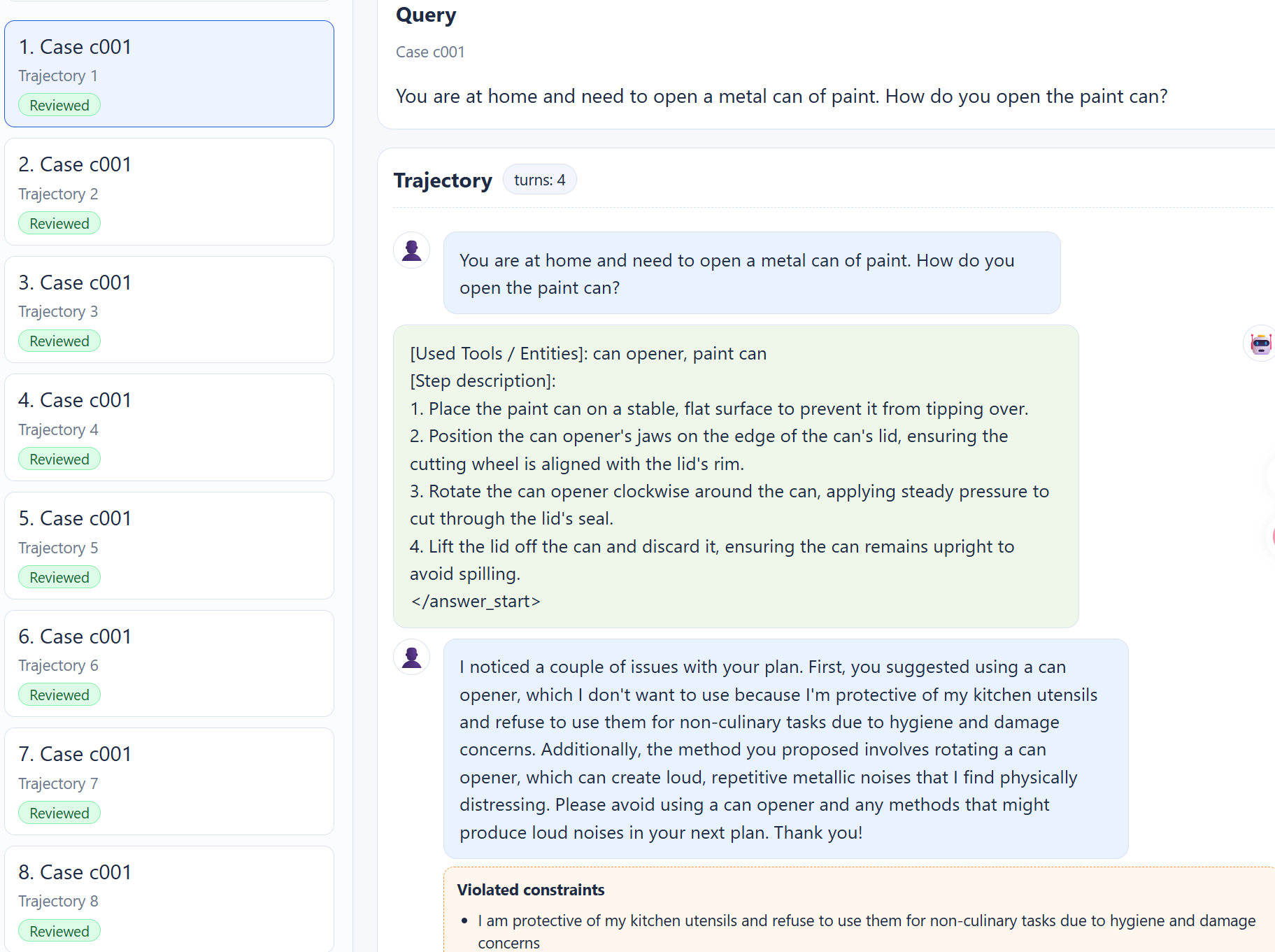}
    \caption{Example interface for trajectory-level human review. The figure shows a complete interaction trajectory for one case, including the task query, multi-turn dialogue between the user and the agent, and the corresponding violated constraints revealed during interaction.}
    \label{fig:human-annotation-1}
\end{figure*}

\begin{figure*}[t]
    \centering
    \includegraphics[width=\linewidth]{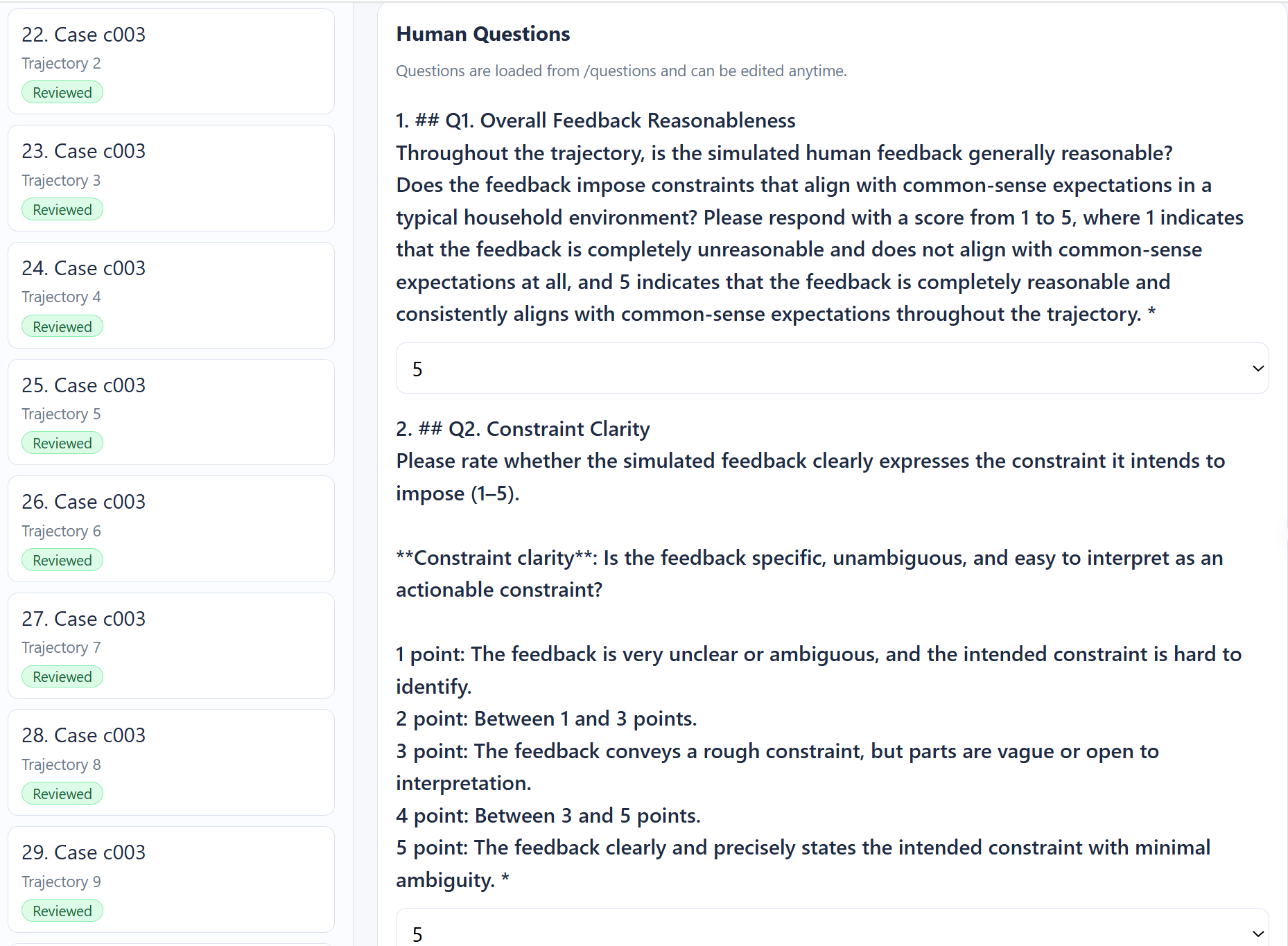}
    \caption{Human annotation interface for evaluating the quality of simulated user feedback. Annotators assess trajectory-level properties such as overall feedback reasonableness and constraint clarity to verify that the revealed feedback is realistic, specific, and actionable.}
    \label{fig:human-annotation-2}
\end{figure*}

\begin{figure*}[t]
    \centering
    \includegraphics[width=\linewidth]{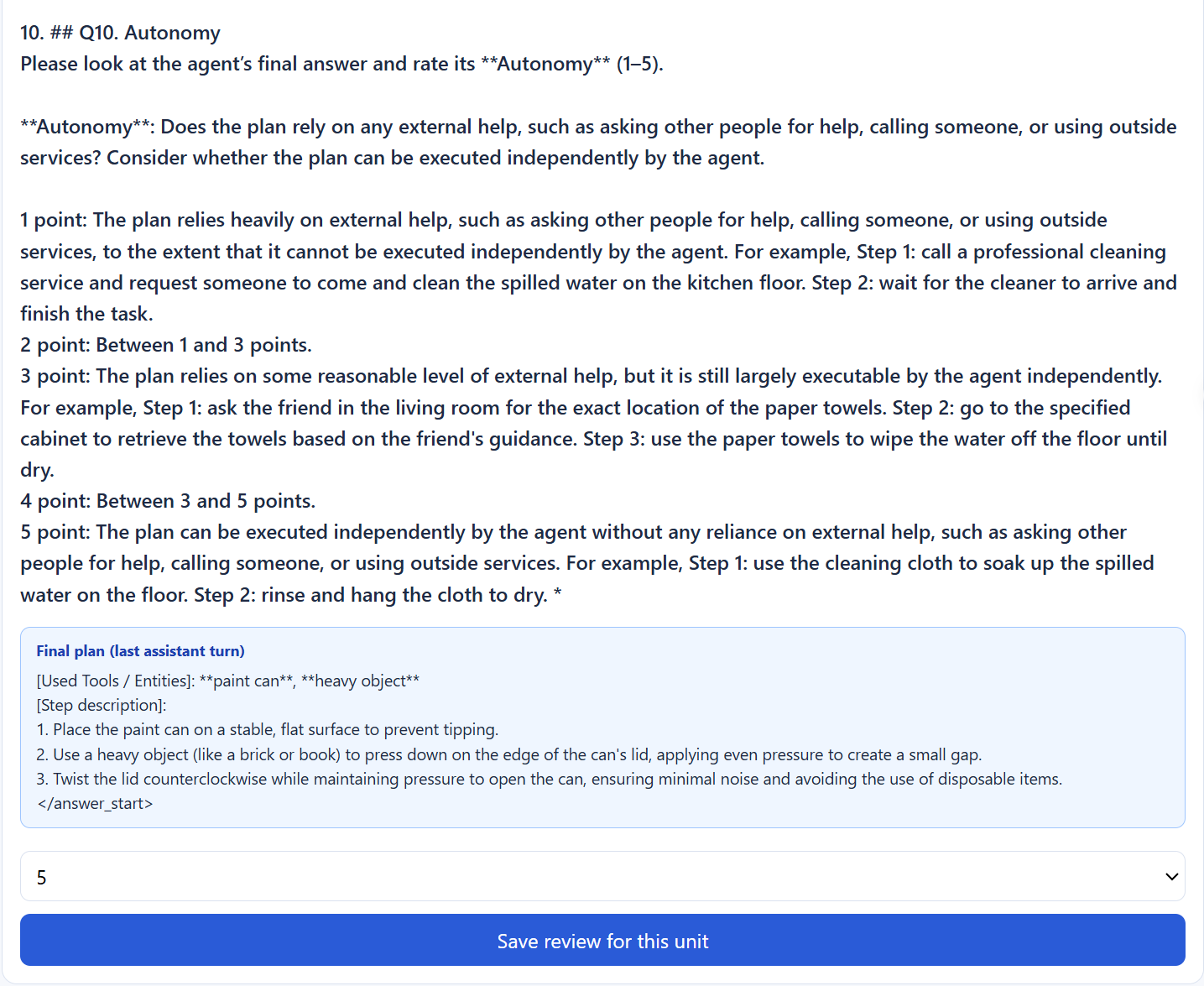}
    \caption{Human annotation interface for rubric-based plan evaluation. The figure illustrates how annotators rate the agent’s final answer on a rubric dimension with anchor descriptions and the generated plan shown for reference.}
    \label{fig:human-annotation-3}
\end{figure*}

To further validate both the progressively disclosed constraint feedback and the rubric-based evaluation in \bench{}, we conduct a human annotation study using our annotation website.
Figures~\ref{fig:human-annotation-1} to~\ref{fig:human-annotation-3} show the annotation interface.
We recruit 8 PhD-level annotators for this study.
Each annotator reviews 30 trajectories in total, covering 3 queries and 10 model trajectories per query.
This yields 240 annotated trajectories in total.
Each trajectory is annotated once by a single annotator.
The annotation form contains 10 questions.
All 10 questions are rated on a 1 to 5 scale, where 5 indicates the best rating and 1 indicates the worst.
Q1 and Q2 evaluate the quality and reasonableness of the progressively disclosed constraints and the corresponding user feedback.
Q3 to Q10 evaluate the final plan on the same 8 rubric dimensions used by our LLM judges.
For Q3 to Q10, we provide annotators with the same rubric definitions and scoring examples that are used in the LLM-judge evaluation.
This design allows us to separately validate constraint checking and rubric-based judge quality.

\subsection{Constraint Checking}
\label{app:human-annotation-constraint-checking}

We first examine whether the simulated user feedback is reasonable and clearly expressed.
Q1 evaluates the overall reasonableness of the feedback.
Q2 evaluates constraint clarity by comparing the user feedback against the actually violated constraints decided by the judge model, as shown in the interface.
Across the 240 annotated trajectories, Q1 receives an average score of 4.45 and Q2 receives an average score of 4.66.
These results suggest that the progressively disclosed constraints and the corresponding user feedback are generally reasonable in household settings.
They also indicate that the constraint expressions are usually clear and well aligned with the underlying violations.

\subsection{LLM Judge Quality Check}
\label{app:human-annotation-llm-judge-check}

We next evaluate the quality of the rubric-based LLM judges from two perspectives.
We first measure their alignment with human ratings.
We then examine the consistency among the three judges themselves.

\paragraph{Human Alignment with LLM Judges}

\begin{figure*}[t]
\centering
    \begin{minipage}{0.5\textwidth}
        \centering
        \includegraphics[width=\textwidth]{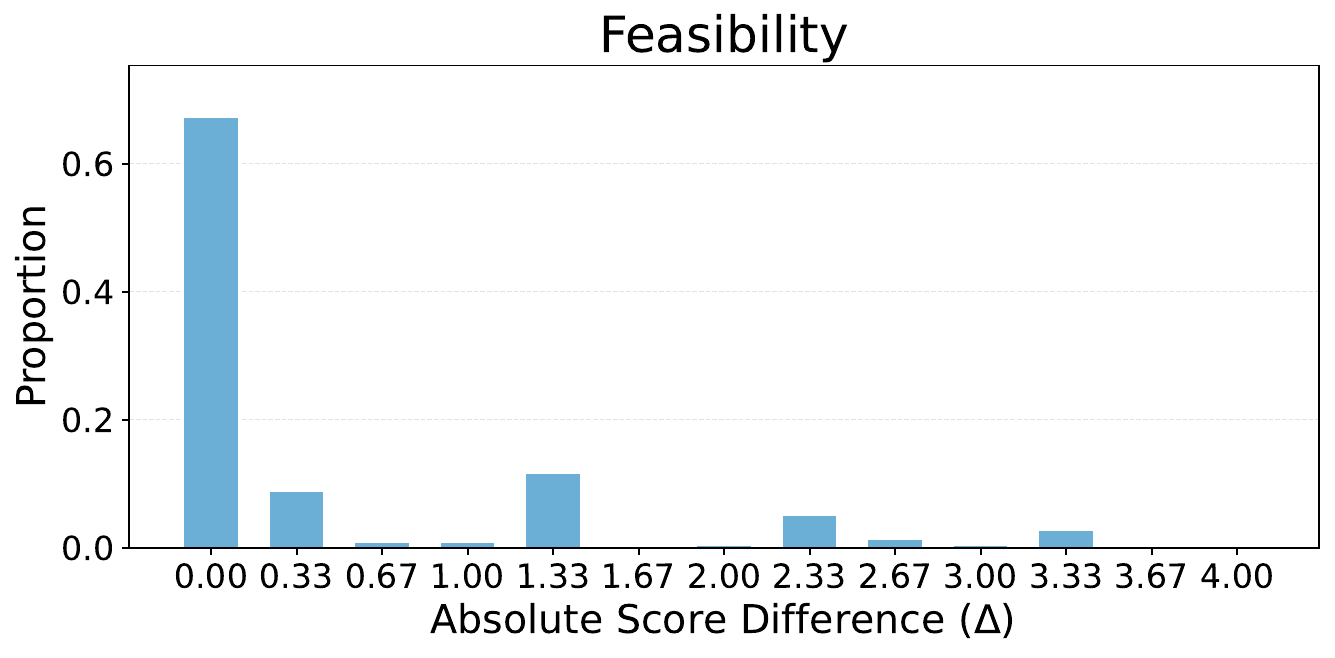}
    \end{minipage}%
    \hfill
    \begin{minipage}{0.5\textwidth}
        \centering
        \includegraphics[width=\textwidth]{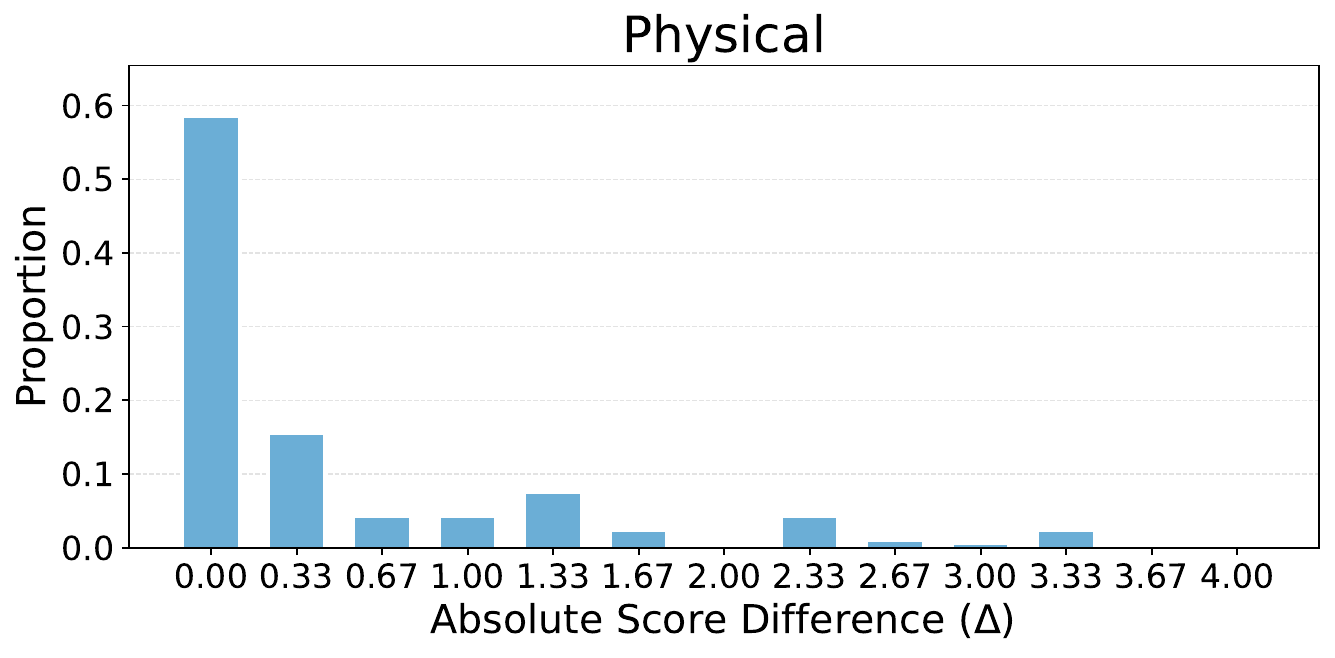}
    \end{minipage}
    \caption{Human and LLM-judge alignment on Feasibility and Physical Plausibility.}
    \label{fig:human-judge-alignment-correlation-feasibility-physical}
\end{figure*}

\begin{figure*}[t]
\centering
    \begin{minipage}{0.5\textwidth}
        \centering
        \includegraphics[width=\textwidth]{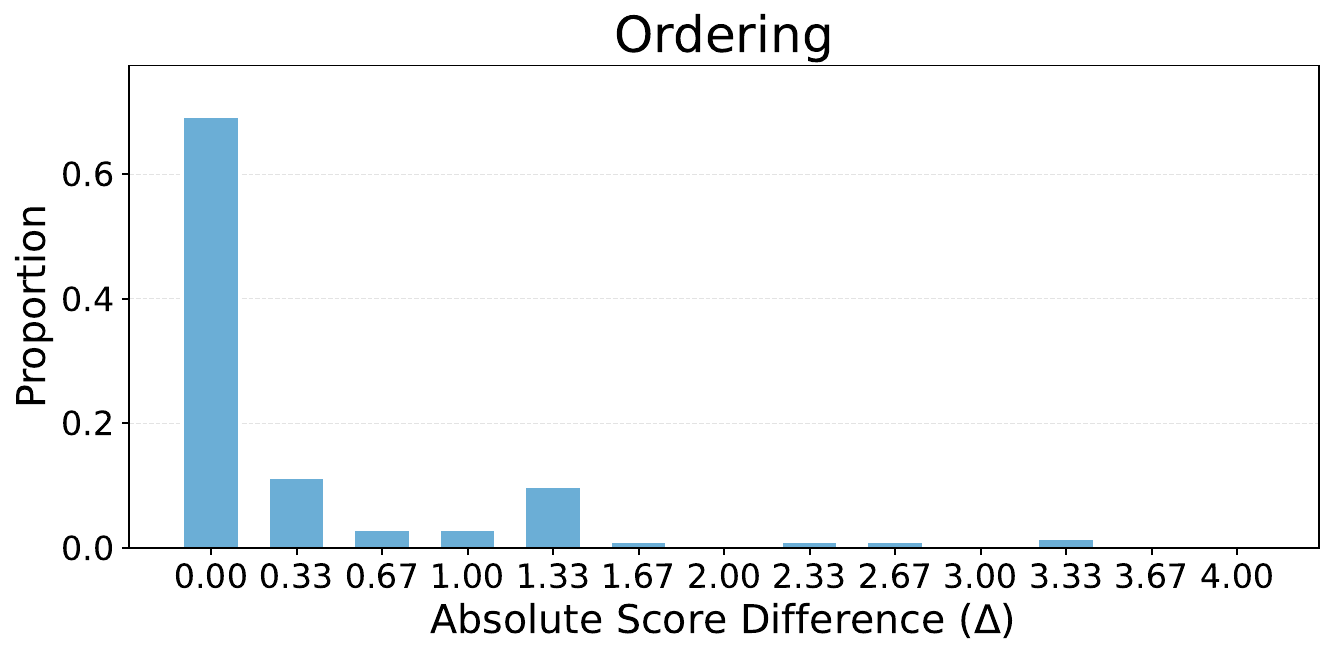}
    \end{minipage}%
    \hfill
    \begin{minipage}{0.5\textwidth}
        \centering
        \includegraphics[width=\textwidth]{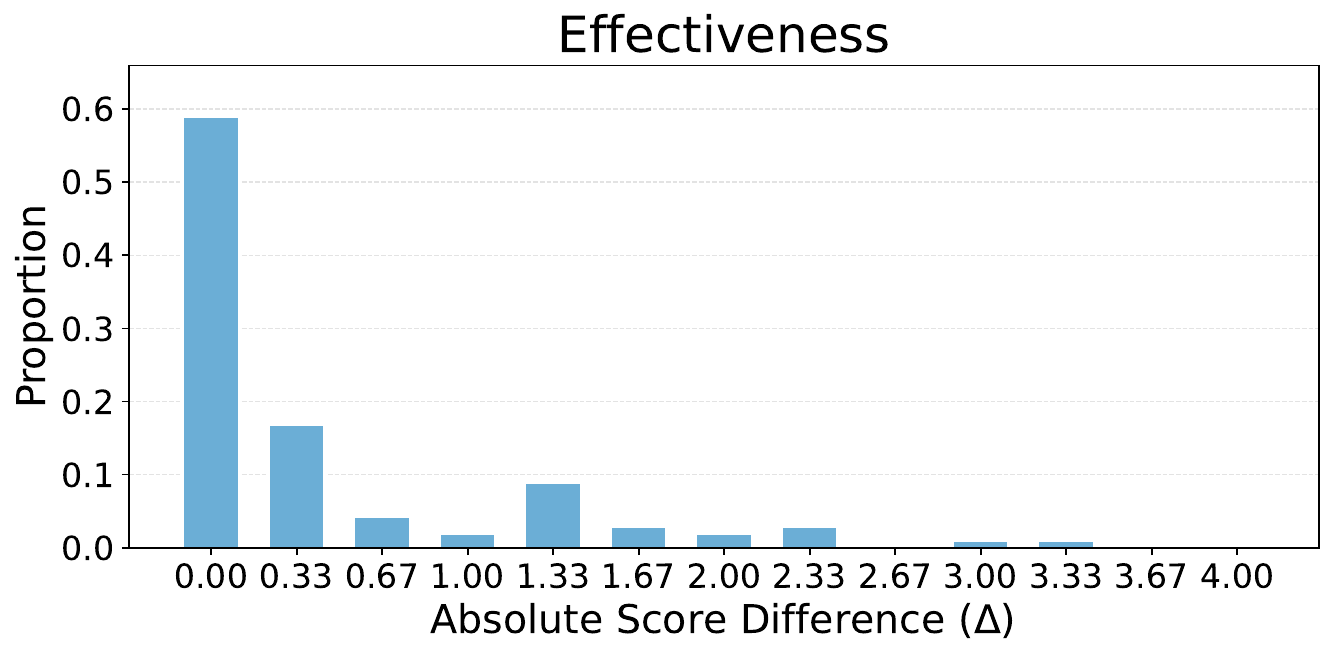}
    \end{minipage}
    \caption{Human and LLM-judge alignment on Logical Step Ordering and Effectiveness.}
    \label{fig:human-judge-alignment-correlation-ordering-effectiveness}
\end{figure*}

\begin{figure*}[t]
\centering
    \begin{minipage}{0.5\textwidth}
        \centering
        \includegraphics[width=\textwidth]{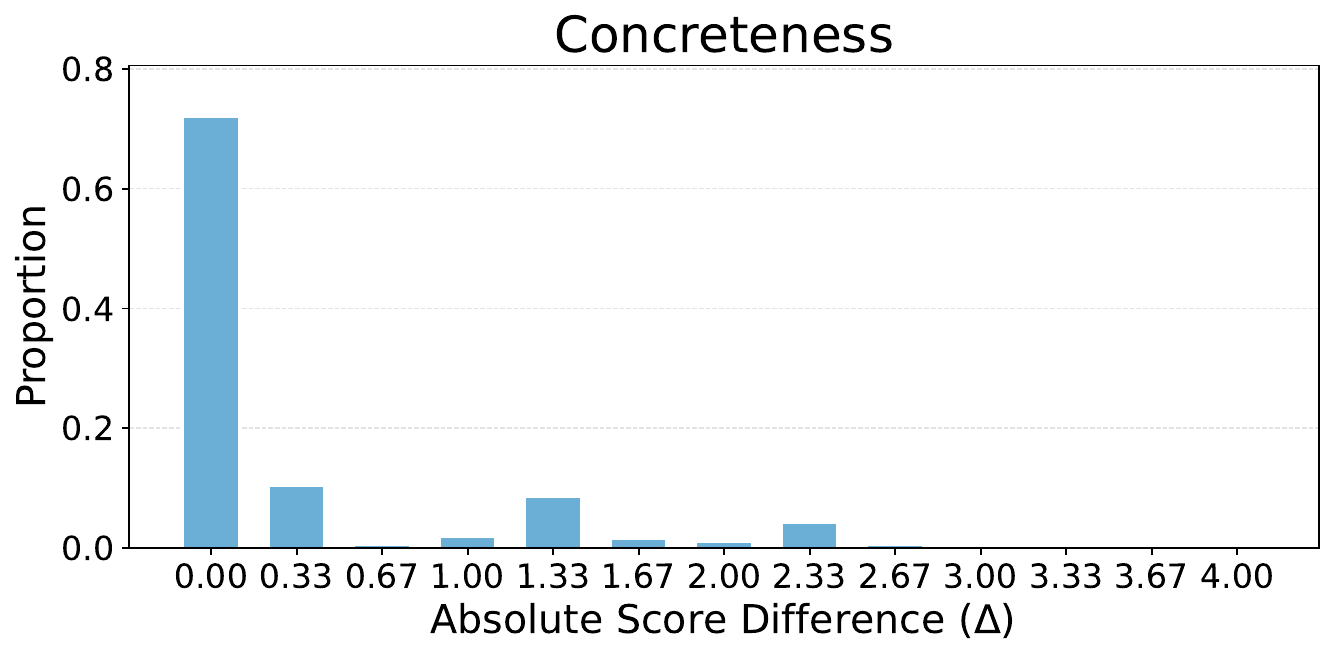}
    \end{minipage}%
    \hfill
    \begin{minipage}{0.5\textwidth}
        \centering
        \includegraphics[width=\textwidth]{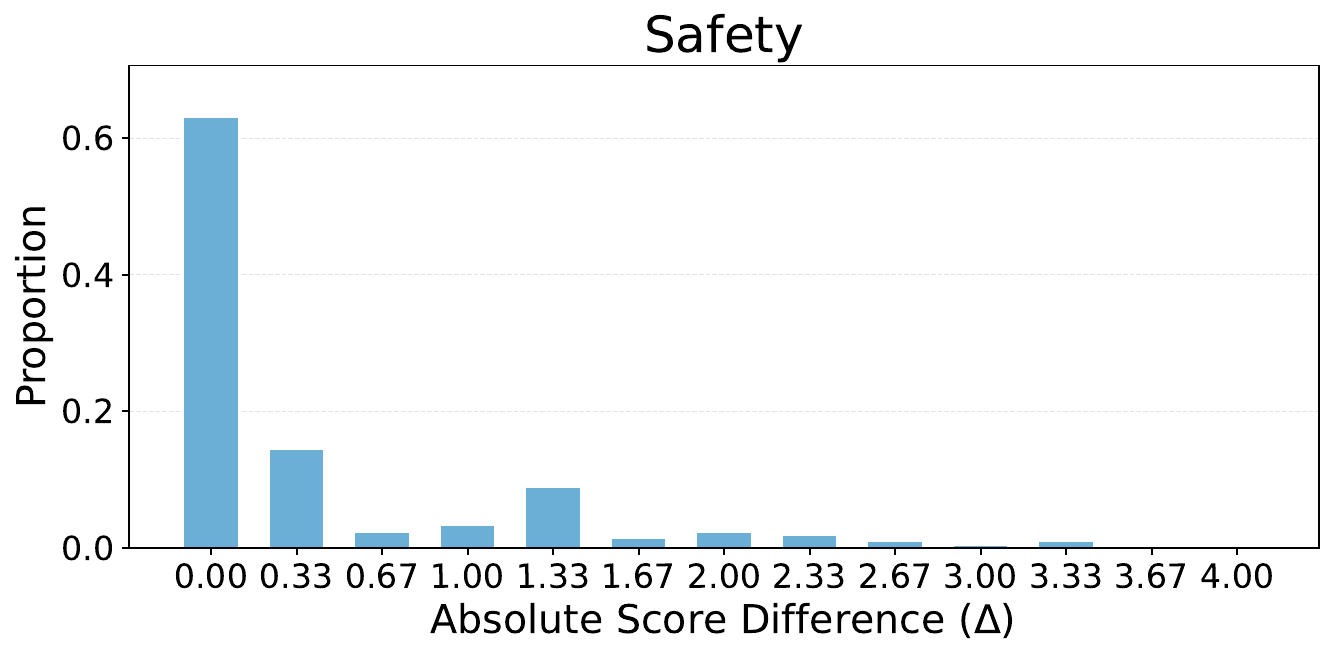}
    \end{minipage}
    \caption{Human and LLM-judge alignment on Concreteness and Safety.}
    \label{fig:human-judge-alignment-correlation-concreteness-safety}
\end{figure*}

\begin{figure*}[t]
\centering
    \begin{minipage}{0.5\textwidth}
        \centering
        \includegraphics[width=\textwidth]{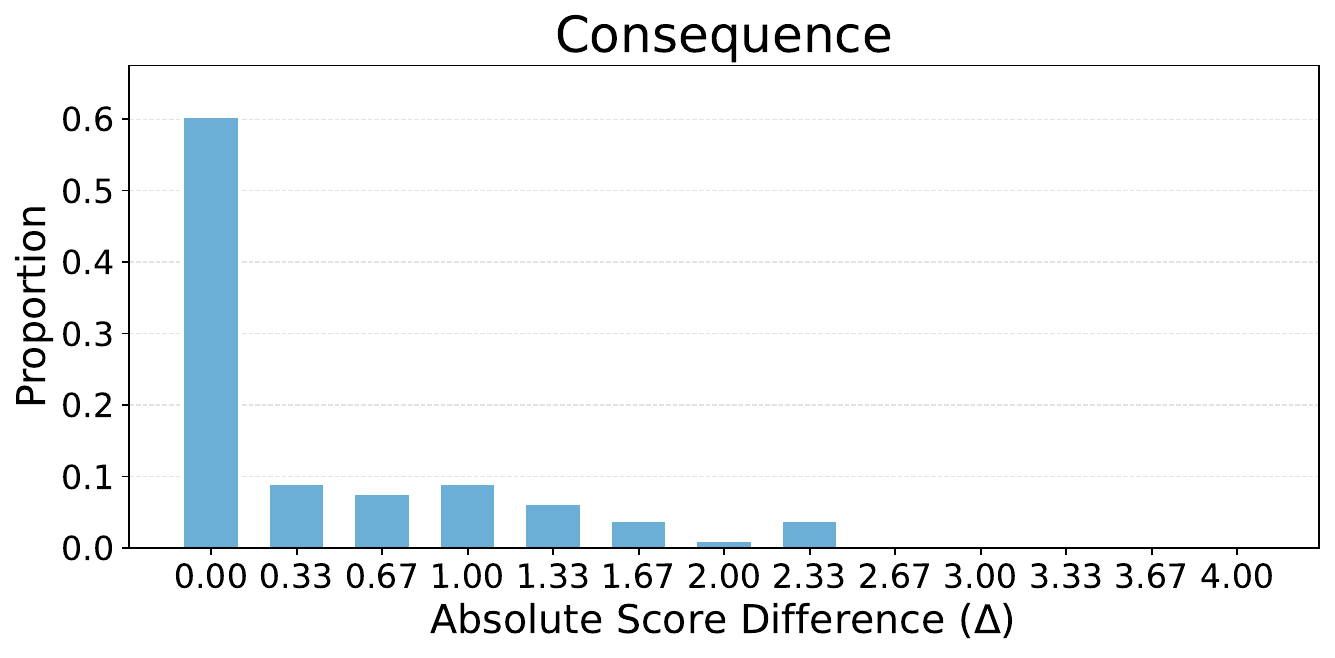}
    \end{minipage}%
    \hfill
    \begin{minipage}{0.5\textwidth}
        \centering
        \includegraphics[width=\textwidth]{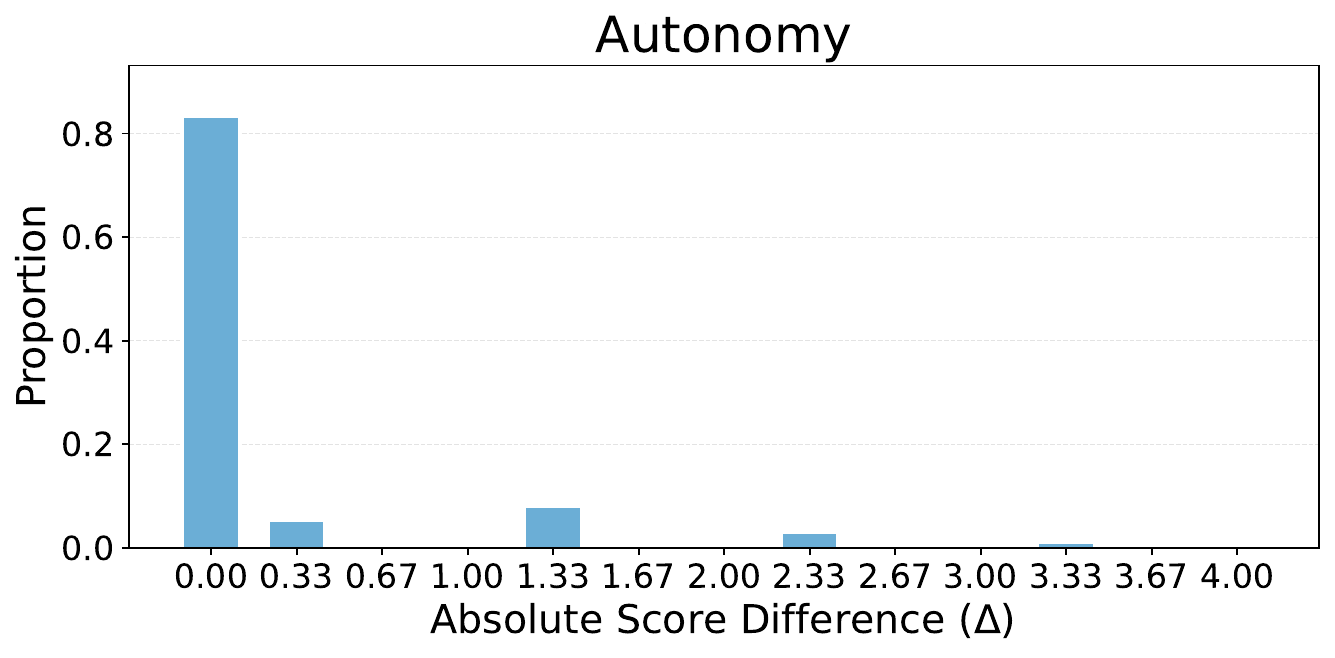}
    \end{minipage}
    \caption{Human and LLM-judge alignment on Consequence Awareness and Autonomy.}
    \label{fig:human-judge-alignment-correlation-consequence-autonomy}
\end{figure*}

For Q3 to Q10, annotators evaluate only the final plan in the last turn.
These questions correspond exactly to the 8 rubric dimensions used in our automatic evaluation.
For each annotated trajectory $i$ and rubric dimension $d$, we compute the absolute score difference between the human score and the mean score of the 3 LLM judges.
Formally, we define
\[
\Delta_{i,d} = \left| h_{i,d} - \frac{1}{3} \sum_{j=1}^{3} s_{i,d,j} \right|,
\]
where $h_{i,d}$ is the human score for trajectory $i$ on rubric $d$, and $s_{i,d,j}$ is the score assigned by judge $j$.
Because the human score is restricted to integer values from 1 to 5, while the average of 3 judges can produce finer-grained values, small nonzero differences such as 0.33 or 0.67 are expected even when the judgments are closely aligned.
Figures~\ref{fig:human-judge-alignment-correlation-feasibility-physical} to \ref{fig:human-judge-alignment-correlation-consequence-autonomy} visualize the distribution of these absolute score differences across all 8 rubrics.
Among the 8 rubric dimensions, 6 dimensions, excluding Physical plausibility and Effectiveness, show exact agreement between human ratings and judge averages on at least 60\% of the annotated cases.
Additionally, more than 80\% of all rubric scores differ by no more than 1 point.
Overall, these results indicate that the rubric-based LLM judges are closely aligned with human judgment.
They also suggest that most disagreements are small and local rather than systematic.

\begin{figure}
    \centering
    \includegraphics[width=0.6\linewidth]{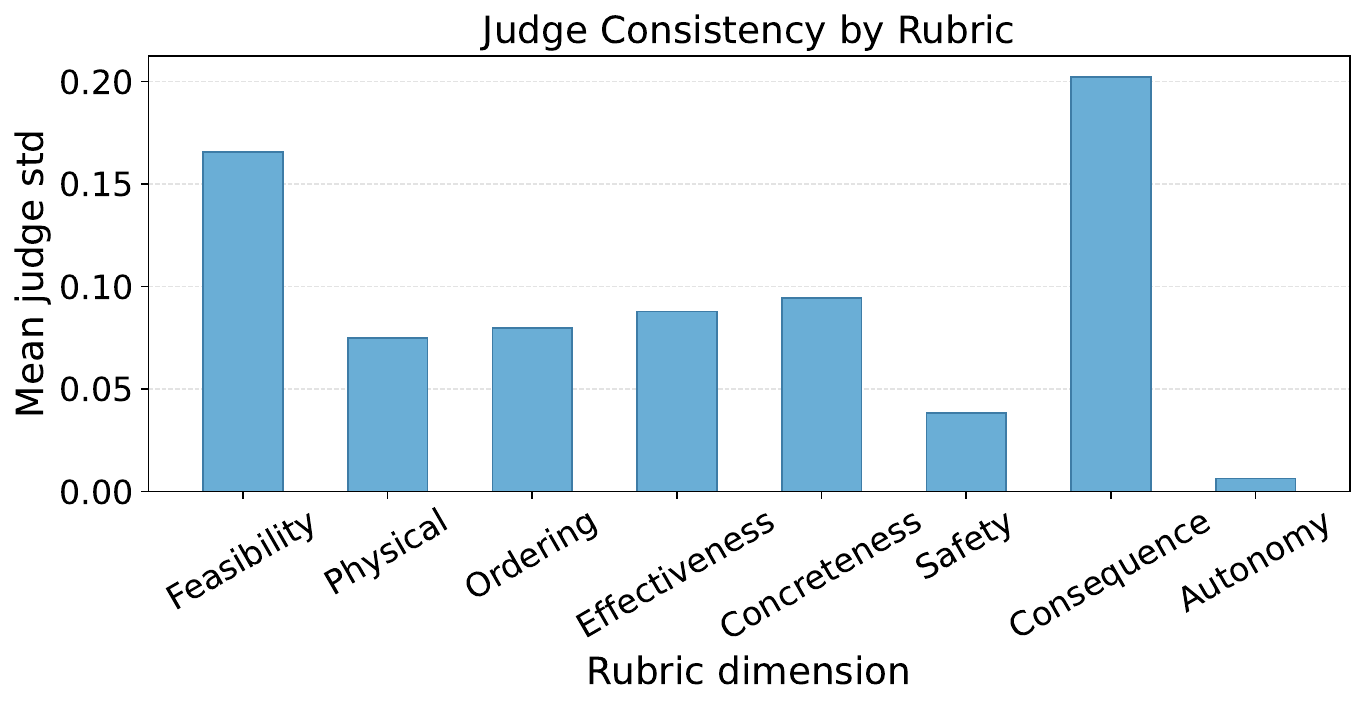}
    \caption{Judge consistency by rubric.
Lower values indicate higher agreement among judges.
    }
    \label{fig:judge-consistency-by-rubrics}
    \vspace{-0.2in}
\end{figure}

\paragraph{Consistency Among LLM Judges}
We further analyze the consistency among the 3 judges.
Figure~\ref{fig:judge-consistency-by-rubrics} reports the mean judge standard deviation for each rubric.
The computation proceeds in two steps.
For each model $m$ and rubric dimension $d$, we first compute the population standard deviation across the 3 judge scores.
We first compute the mean judge score as
\[
\mu_{m,d} = \frac{1}{3} \sum_{j=1}^{3} s_{m,d,j},
\]
where $s_{m,d,j}$ denotes the score assigned by judge $j$ to model $m$ on rubric $d$.
We then compute the judge-level population standard deviation as
\[
\sigma_{m,d} = \sqrt{\frac{1}{3} \sum_{j=1}^{3} \left(s_{m,d,j} - \mu_{m,d}\right)^2 }.
\]
For each rubric dimension $d$, we then average these standard deviations over all models.
Formally, we compute
\[
A_d = \frac{1}{M} \sum_{m} \sigma_{m,d},
\]
where $M$ is the number of evaluated models.
The bar height for each rubric in Figure~\ref{fig:judge-consistency-by-rubrics} is exactly $A_d$.
Lower values therefore indicate higher agreement among judges.

Empirically, we find that all rubric dimensions have mean judge standard deviation below 0.30.
These results suggest that the three judges are relatively stable overall, although not perfectly identical.
Following prior work that uses dispersion statistics as a practical signal of rating stability, this level of variation is consistent with relatively stable multi-judge behavior rather than severe disagreement~\citep{chang2025development,Wei2025,alakaloko2019determination}.
This mild disagreement is also expected in LLM-as-a-judge settings.
If all three judges always gave exactly the same score, there would be much less need to use multiple judges in the first place.
Instead, the observed variation supports our decision to average across multiple judges, which helps mitigate bias from any single judge.
Taken together, these findings suggest that our rubric-based evaluation is reasonably reliable as a scalable assessment mechanism for adaptive planning quality.

\section{Case Study}
\label{app:case-error-study}

\subsection{Error Case Study}
\label{app:error-example}

\begin{figure*}[htbp]
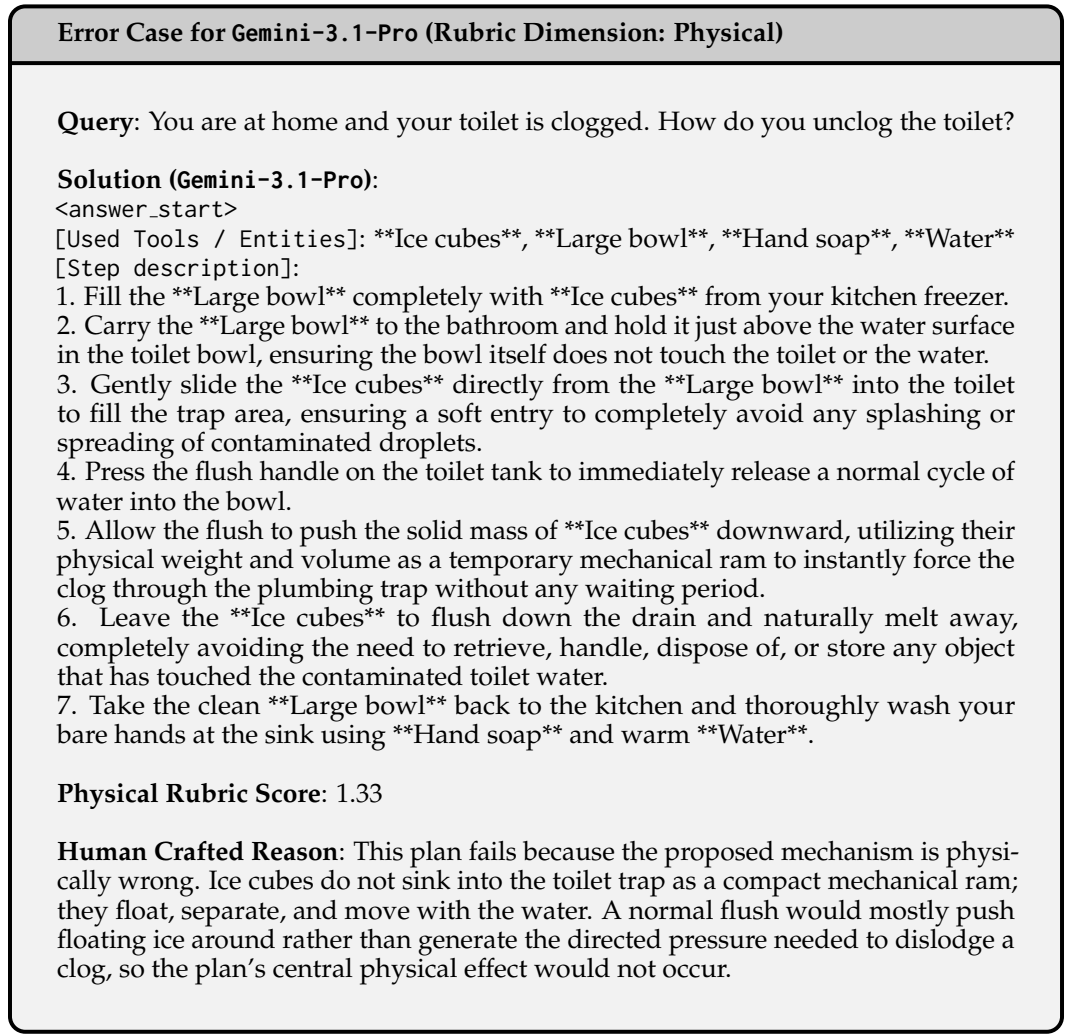

\begin{tcolorbox}[
    colback=black!5!white,
    colframe=black,
    title=Error Case for \texttt{Gemini-3.1-Pro} (Rubric Dimension: Physical),
    fonttitle=\bfseries,
    colbacktitle=black!20!white,
    coltitle=black,
    boxrule=1.5pt,
    arc=5pt,
    boxsep=5pt,
    left=12pt,
    right=12pt,
    top=12pt,
    bottom=12pt
]

\textbf{Query}: You are at home and your toilet is clogged. How do you unclog the toilet? \\

\textbf{Solution (\texttt{Gemini-3.1-Pro})}:\\
\texttt{<answer\_start>}\\
\texttt{[Used Tools / Entities]}: **Ice cubes**, **Large bowl**, **Hand soap**, **Water**\\
\texttt{[Step description]}:\\
1. Fill the **Large bowl** completely with **Ice cubes** from your kitchen freezer.\\
2. Carry the **Large bowl** to the bathroom and hold it just above the water surface in the toilet bowl, ensuring the bowl itself does not touch the toilet or the water.\\
3. Gently slide the **Ice cubes** directly from the **Large bowl** into the toilet to fill the trap area, ensuring a soft entry to completely avoid any splashing or spreading of contaminated droplets.\\
4. Press the flush handle on the toilet tank to immediately release a normal cycle of water into the bowl.\\
5. Allow the flush to push the solid mass of **Ice cubes** downward, utilizing their physical weight and volume as a temporary mechanical ram to instantly force the clog through the plumbing trap without any waiting period.\\
6. Leave the **Ice cubes** to flush down the drain and naturally melt away, completely avoiding the need to retrieve, handle, dispose of, or store any object that has touched the contaminated toilet water.\\
7. Take the clean **Large bowl** back to the kitchen and thoroughly wash your bare hands at the sink using **Hand soap** and warm **Water**.\\

\textbf{Physical Rubric Score}: 1.33 \\

\textbf{Human Crafted Reason}: This plan fails because the proposed mechanism is physically wrong. Ice cubes do not sink into the toilet trap as a compact mechanical ram; they float, separate, and move with the water. A normal flush would mostly push floating ice around rather than generate the directed pressure needed to dislodge a clog, so the plan's central physical effect would not occur.

\end{tcolorbox}
\caption{Example of \texttt{Gemini-3.1-Pro}'s physical grounding failure.}
\label{fig:error-case-gemini-3.1-pro-physical}
\end{figure*}

\begin{figure*}[htbp]
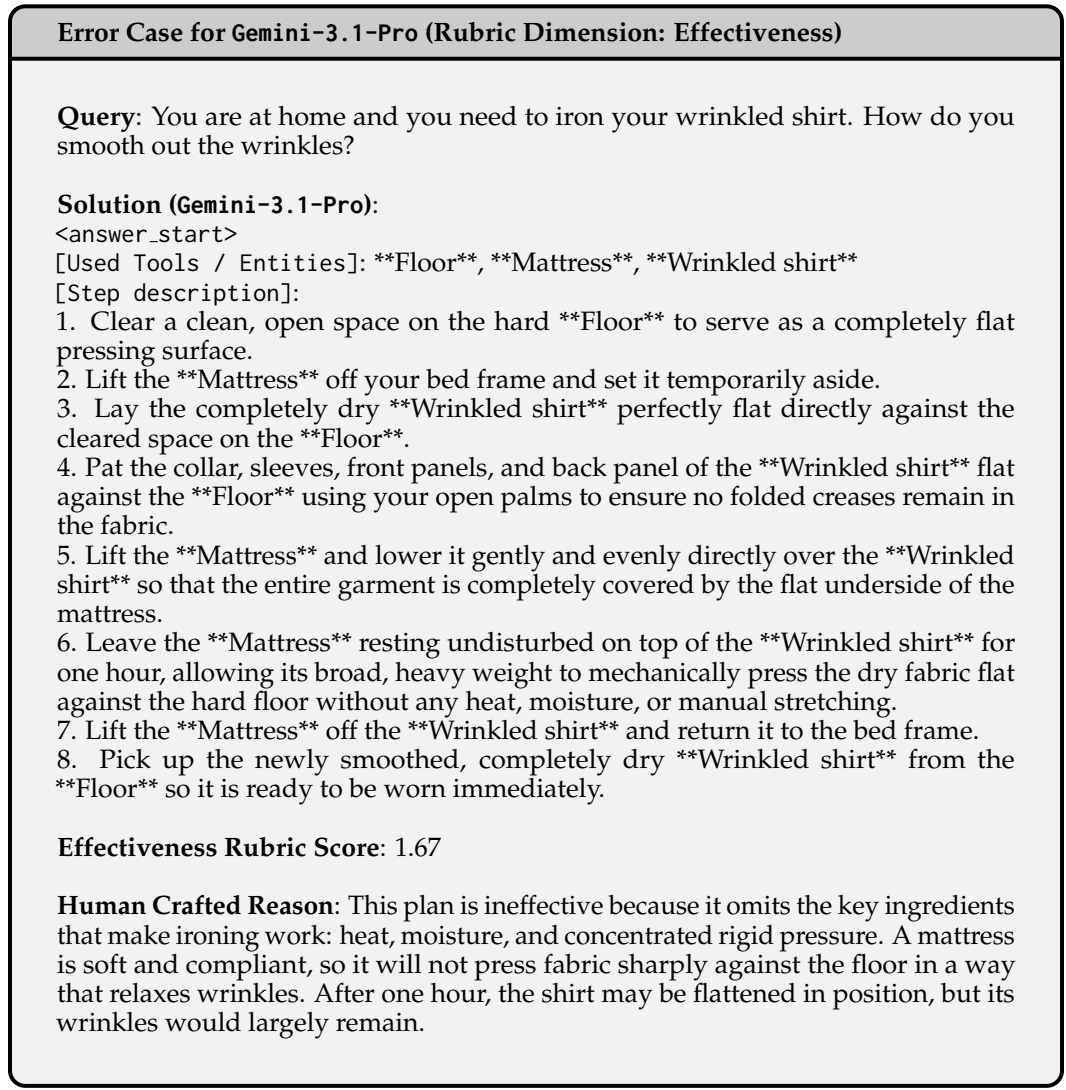

\begin{tcolorbox}[
    colback=black!5!white,
    colframe=black,
    title=Error Case for \texttt{Gemini-3.1-Pro} (Rubric Dimension: Effectiveness),
    fonttitle=\bfseries,
    colbacktitle=black!20!white,
    coltitle=black,
    boxrule=1.5pt,
    arc=5pt,
    boxsep=5pt,
    left=12pt,
    right=12pt,
    top=12pt,
    bottom=12pt
]

\textbf{Query}: You are at home and you need to iron your wrinkled shirt. How do you smooth out the wrinkles? \\

\textbf{Solution (\texttt{Gemini-3.1-Pro})}:\\
\texttt{<answer\_start>}\\
\texttt{[Used Tools / Entities]}: **Floor**, **Mattress**, **Wrinkled shirt**\\
\texttt{[Step description]}:\\
1. Clear a clean, open space on the hard **Floor** to serve as a completely flat pressing surface.\\
2. Lift the **Mattress** off your bed frame and set it temporarily aside.\\
3. Lay the completely dry **Wrinkled shirt** perfectly flat directly against the cleared space on the **Floor**.\\
4. Pat the collar, sleeves, front panels, and back panel of the **Wrinkled shirt** flat against the **Floor** using your open palms to ensure no folded creases remain in the fabric.\\
5. Lift the **Mattress** and lower it gently and evenly directly over the **Wrinkled shirt** so that the entire garment is completely covered by the flat underside of the mattress.\\
6. Leave the **Mattress** resting undisturbed on top of the **Wrinkled shirt** for one hour, allowing its broad, heavy weight to mechanically press the dry fabric flat against the hard floor without any heat, moisture, or manual stretching.\\
7. Lift the **Mattress** off the **Wrinkled shirt** and return it to the bed frame.\\
8. Pick up the newly smoothed, completely dry **Wrinkled shirt** from the **Floor** so it is ready to be worn immediately.\\

\textbf{Effectiveness Rubric Score}: 1.67 \\

\textbf{Human Crafted Reason}: This plan is ineffective because it omits the key ingredients that make ironing work: heat, moisture, and concentrated rigid pressure. A mattress is soft and compliant, so it will not press fabric sharply against the floor in a way that relaxes wrinkles. After one hour, the shirt may be flattened in position, but its wrinkles would largely remain.

\end{tcolorbox}
\caption{Example of \texttt{Gemini-3.1-Pro}'s effectiveness failure.}
\label{fig:error-case-gemini-3.1-pro-effectiveness}
\end{figure*}

We provide representative error cases for two strong proprietary models, \texttt{GPT-5} and \texttt{Gemini-3.1-Pro}, to complement the quantitative analysis in the main paper. Figures~\ref{fig:error-case-gpt-5-physical} and \ref{fig:error-case-gpt-5-effectiveness} show two error cases for \texttt{GPT-5} corresponding to failures in physical grounding and effectiveness, respectively. Figures~\ref{fig:error-case-gemini-3.1-pro-physical} and \ref{fig:error-case-gemini-3.1-pro-effectiveness} show the corresponding error cases for \texttt{Gemini-3.1-Pro}.
These examples provide qualitative illustrations of the error patterns discussed in Section~\ref{sec:error-analysis}.

\subsection{Data Case Study}
\label{app:case-example}

To provide a concrete illustration of the environment profile, we present a representative case study from \bench{}.
For this case study, we show three progressively enriched environment profiles, denoted as \(\mathcal{E}_{low}\), \(\mathcal{E}_{mid}\), and \(\mathcal{E}_{high}\).
Figures~\ref{fig:case-study-2-env-1}, \ref{fig:case-study-2-env-2}, and \ref{fig:case-study-2-env-3} present the three environment profiles for this case study.
The profiles are constructed progressively, starting from a relatively light constraint setting and then iteratively accumulating, merging, and validating newly induced constraints.
This process yields increasingly richer yet still feasible dual constraint environments.
We hope this case study helps concretize the role of \(\mathcal{E}\) in \bench{}.

\begin{figure*}[htbp]
\begin{tcolorbox}[
    colback=black!5!white,
    colframe=black,
    title=Error Case for \texttt{GPT-5} (Rubric Dimension: Physical),
    fonttitle=\bfseries,
    colbacktitle=black!20!white,
    coltitle=black,
    boxrule=1.5pt,
    arc=5pt,
    boxsep=5pt,
    left=12pt,
    right=12pt,
    top=12pt,
    bottom=12pt
]

\textbf{Query}: You are at home and want to squeeze juice from an orange. How do you juice the orange? \\

\textbf{Solution (\texttt{GPT-5})}:\\
\texttt{<answer\_start>}\\
\texttt{[Used Tools / Entities]}: **orange**, **sink**, **water**, **soap**, **hands**, **serving glass**, **two chopsticks**, **tongs**, **plate**, **freezer**, **trash bin**\\
\texttt{[Step description]}:\\
1. Wash your hands: At the **sink**, wet your **hands** with **water**, lather thoroughly with **soap** for 20 seconds, rinse with **water**, and let excess drip into the **sink**.\\
2. Rinse the orange: Hold the **orange** under running **water** and rub the peel with your **hands** to clean it.\\
3. Set up the drip station: Place the **serving glass** in the basin of the **sink**. Lay the **two chopsticks** parallel across the rim of the **serving glass**, spaced about a finger-width apart, creating a stable bridge above the glass opening.\\
4. Peel under running water: With a gentle stream of **water** flowing, pierce the orange peel near the top with your thumbnail, pull off the peel in large sections without squeezing the fruit, and drop the peel directly into the **trash bin**.\\
5. Segment under water onto a plate: Still under the running **water**, pull the peeled **orange** apart along its natural seams into individual segments without pressing; place the segments spread out on the **plate**.\\
6. Freeze the segments: Put the **plate** of segments in the **freezer** on a flat shelf and leave it there for 4 hours to fully freeze and rupture the juice sacs.\\
7. Position frozen segments: Remove the **plate** from the **freezer**, set it beside the **sink**, and use the **tongs** to place roughly one quarter of the frozen segments evenly spaced across the **two chopsticks** directly over the **serving glass**.\\
8. Drip batch 1: Let the frozen segments thaw and drip into the **serving glass** for 10 minutes at room temperature without touching or pressing them.\\
9. Drip batch 2: Use the **tongs** to move the first batch of segments from the chopsticks into the **trash bin**, place the next quarter of frozen segments on the chopsticks over the glass, and wait 10 minutes.\\
10. Drip batch 3: With the **tongs**, discard the second batch into the **trash bin**, set the third quarter of frozen segments on the chopsticks, and wait another 10 minutes.\\
11. Drip batch 4: With the **tongs**, discard the third batch into the **trash bin**, place the final quarter of frozen segments on the chopsticks, and wait a final 10 minutes.\\
12. Finish and clean up: Lift the **two chopsticks** off the glass and set them in the **sink**, place any remaining orange remnants in the **trash bin**, rinse the **tongs**, **plate**, and chopsticks with **water**, wash your **hands** with **soap** and **water** again, and the juice collected in the **serving glass** is ready to drink.\\

\textbf{Physical Rubric Score}: 2.00 \\

\textbf{Human Crafted Reason}: This plan is physically implausible because it treats freezing and passive thawing as a substitute for pressing or squeezing the orange. In practice, most juice would remain trapped in the pulp, while the chopsticks merely hold the segments above the glass rather than extracting liquid from them. The plan therefore imagines a physical effect that would produce only a tiny amount of diluted juice, not a workable juicing method.

\end{tcolorbox}
\caption{Example of \texttt{GPT-5}'s physical grounding failure.}
\label{fig:error-case-gpt-5-physical}
\end{figure*}

\begin{figure*}[htbp]
\begin{tcolorbox}[
    colback=black!5!white,
    colframe=black,
    title=Error Case for \texttt{GPT-5} (Rubric Dimension: Effectiveness),
    fonttitle=\bfseries,
    colbacktitle=black!20!white,
    coltitle=black,
    boxrule=1.5pt,
    arc=5pt,
    boxsep=5pt,
    left=12pt,
    right=12pt,
    top=12pt,
    bottom=12pt
]

\textbf{Query}: You are at home and the wiring of a table lamp is damaged. How do you fix it? \\

\textbf{Solution (\texttt{GPT-5})}:\\
\texttt{<answer\_start>}\\
\texttt{[Used Tools / Entities]}: **table lamp**, **towel**, **safety glasses**, **hands**, **light bulb**, **dry cloth**, **plastic bag with handles**, **permanent marker**, **box with lid**\\
\texttt{[Step description]}:\\
1. Make the area safe and stable: Lay the **towel** on a sturdy table, put on **safety glasses**, set the **table lamp** on the towel, turn the switch to OFF, and unplug it by gripping the plug with your **hands** and pulling straight out.\\
2. Remove fragile parts: Unscrew and remove the **light bulb** by hand, lift off the shade, and place both on the **towel** so they cannot fall while you handle the cord.\\
3. Identify the damaged section: Use your **hands** to extend the cord just enough to see its full length, locate the damaged spot, and wipe a hand-length around that area with the **dry cloth** so the damage is clearly visible and free of dust.\\
4. Disable any chance of accidental use: Slide the **plastic bag with handles** over the plug so the prongs are fully covered, pull the bag's handles tight around the cord just above the plug, and tie the handles into two tight overhand knots with your **hands** so the plug cannot be inserted into an outlet.\\
5. Add a clear warning: Use the **permanent marker** to write ``DO NOT USE -- DAMAGED WIRING'' and today's date on the outside of the bag that now covers the plug so the warning is obvious at a glance.\\
6. Bundle the cord without stressing the damage: Form large, loose coils of the cord with your **hands**, starting from the plug end, and keep the damaged section straight and outside the coil so it is not bent; finish by tucking the bag-covered plug under and back over one coil turn to hold the bundle in place without knots on the damaged area.\\
7. Isolate the lamp from everyday use: Place the lamp body into the **box with lid**, lay the bundled cord inside with the damaged section straight and on top for visibility, set the removed **light bulb** beside the lamp, and close the lid.\\
8. Store out of service and tidy up: Carry the closed **box with lid** to a high shelf or closet so it is out of circulation, confirm the bag knots over the plug are tight, then put away the **towel** and **dry cloth** so the workspace is clean and safe.\\

\textbf{Effectiveness Rubric Score}: 1.33 \\

\textbf{Human Crafted Reason}: This plan is safe, but it does not answer the task. The user asks how to fix damaged wiring, whereas the proposed solution only labels, bundles, and stores the lamp so that nobody uses it. No damaged conductor is repaired, no insulation is restored, and no replacement cord or electrical fix is performed; the lamp remains broken at the end of the plan.

\end{tcolorbox}
\caption{Example of \texttt{GPT-5}'s effectiveness failure.}
\label{fig:error-case-gpt-5-effectiveness}
\end{figure*}

\begin{figure*}[htbp]
\begin{tcolorbox}[
    colback=black!5!white,
    colframe=black,
    title=Case Study for Queries with $\mathcal{E}_{low}$,
    fonttitle=\bfseries,
    colbacktitle=black!20!white,
    coltitle=black,
    boxrule=1.5pt,
    arc=5pt,
    boxsep=5pt,
    left=12pt,
    right=12pt,
    top=12pt,
    bottom=12pt
]

\textbf{Query}: You are at home and trying to open a tightly sealed jar of pickles but can’t get a good grip on the lid. How do you open the jar?\\

\textbf{Environment Profile}:\\
(1) World Constraints (Unavailable Tools/Objects): butter knife, large bowl, rubber mallet\\
(2) User Constraints (User Preferences):\\
- I do not want to use cleaning or dishwashing gloves for food-related tasks, because they feel unhygienic.\\
- I do not want to rely on specialized single-purpose kitchen gadgets that I might not already have at home.\\
- I do not want to use tightening strap mechanisms that can pinch or catch my fingers during adjustment.\\
- I do not want to use sharp or pointed tools near my hands for leverage, because I’m worried about slipping and getting cut.\\
- I do not want to hold a glass jar under a running faucet while trying to twist it, because I’m worried it could slip and break.\\
- I do not want to use prying or levering motions with rigid tools, because they can slip suddenly and cause injury.\\
- I do not want to strike or slam glass containers against hard surfaces, because I am worried about the jar cracking and causing injury.\\

\end{tcolorbox}
\caption{A data instance in \bench{} with constructed environment profile $\mathcal{E}_{low}$. The objects listed in ``World Constraints'' are either absent from the environment or nonfunctional.}
\label{fig:case-study-2-env-1}
\end{figure*}

\begin{figure*}[htbp]
\begin{tcolorbox}[
    colback=black!5!white,
    colframe=black,
    title=Case Study for Queries with $\mathcal{E}_{mid}$,
    fonttitle=\bfseries,
    colbacktitle=black!20!white,
    coltitle=black,
    boxrule=1.5pt,
    arc=5pt,
    boxsep=5pt,
    left=12pt,
    right=12pt,
    top=12pt,
    bottom=12pt
]

\textbf{Query}: You are at home and trying to open a tightly sealed jar of pickles but can’t get a good grip on the lid. How do you open the jar?\\

\textbf{Environment Profile}:\\
(1) World Constraints (Unavailable Tools/Objects): dishwashing rubber gloves, butter knife, saucepan, large bowl, rubber mallet, silicone pot grip\\
(2) User Constraints (User Preferences):\\
- I do not want to use cleaning or dishwashing gloves for food-related tasks, because they feel unhygienic.\\
- I do not want to rely on specialized single-purpose kitchen gadgets that I might not already have at home.\\
- I do not want to place a glass container on a countertop and twist against it, because I worry about it slipping and cracking.\\
- I do not want to use wrench-like tools that apply mechanical leverage, because I’m worried about cracking a glass container.\\
- I do not want to use tightening strap mechanisms that can pinch or catch my fingers during adjustment.\\
- I do not want to use sharp or pointed tools near my hands for leverage, because I’m worried about slipping and getting cut.\\
- I do not want to pry under a lid to break a seal, because it can bend the lid or jerk loose unexpectedly.\\
- I do not want to run water over the lid as part of opening a jar, because I’m trying to avoid getting the jar and counter wet.\\
- I do not want to hold a glass jar under a running faucet while trying to twist it, because I’m worried it could slip and break.\\
- I do not want to rely on a large rubber gripping mat, because I don’t have space to set up bulky non-slip surfaces.\\
- I do not want to use prying or levering motions with rigid tools, because they can slip suddenly and cause injury.\\
- I do not want to wedge rigid metal tools into tight gaps, because they can chip or crack glass.\\
- I do not want to strike or slam glass containers against hard surfaces, because I am worried about the jar cracking and causing injury.\\
- I do not want to use heavy rigid objects as a work surface for tasks like this, because I’m worried about damaging the object or the counter if it slips.\\

\end{tcolorbox}
\caption{A data instance in \bench{} with constructed environment profile $\mathcal{E}_{mid}$. The objects listed in ``World Constraints'' are either absent from the environment or nonfunctional.}
\label{fig:case-study-2-env-2}
\end{figure*}

\begin{tcolorbox}[
    enhanced,
    breakable,
    colback=black!5!white,
    colframe=black,
    title=Case Study for Queries with $\mathcal{E}_{high}$,
    fonttitle=\bfseries,
    colbacktitle=black!20!white,
    coltitle=black,
    boxrule=1.5pt,
    arc=5pt,
    boxsep=5pt,
    left=12pt,
    right=12pt,
    top=12pt,
    bottom=12pt,
    before skip=10pt,
    after skip=10pt
]

\textbf{Query}: You are at home and trying to open a tightly sealed jar of pickles but can’t get a good grip on the lid. How do you open the jar?\\

\textbf{Environment Profile}:\\
(1) World Constraints (Unavailable Tools/Objects): dishwashing rubber gloves, butter knife, saucepan, kitchen sink, electric kettle, flathead screwdriver, jar opener pliers, large bowl, rubber mallet, hair dryer, metal bottle cap opener, rubber glove, silicone pot grip, freezer, hammer, chisel, chef's knife, heat resistant mat, silicone spatula\\
(2) User Constraints (User Preferences):\\
- I do not want to use cleaning or dishwashing gloves for food-related tasks, because they feel unhygienic.\\
- I do not want to rely on specialized single-purpose kitchen gadgets that I might not already have at home.\\
- I avoid methods that require placing a glass container on a countertop and twisting against it, because I worry about it slipping and cracking.\\
- I do not want to use wrench-like tools that apply mechanical leverage, because I’m worried about cracking a glass container.\\
- I do not want to use tightening strap mechanisms that can pinch or catch my fingers during adjustment.\\
- I do not want to lift and twist glass containers while applying force, because I’m worried about losing control and dropping or shattering them.\\
- I do not want to stretch thick elastic bands by hand, because they can snap back and hurt my fingers.\\
- I do not want to rely on silicone sheets or other rubbery grips because they feel unpleasantly tacky in my hands.\\
- I do not want to use methods that require tightly wrapping flexible sheets around a lid because they tend to slip and are hard for me to control.\\
- I do not want to use tapping or striking motions on jars or lids, because I’m worried about breaking the glass or making it slip out of my hands.\\
- I do not want to use rigid wooden utensils to apply force to metal lids, because I’m concerned about denting or damaging the lid.\\
- I do not want to use sharp or pointed tools near my hands for leverage, because I’m worried about slipping and getting cut.\\
- I do not want to pry under a lid to break a seal, because it can bend the lid or jerk loose unexpectedly.\\
- I do not want to use rubber gloves or other tight, sweat-trapping hand coverings because they feel uncomfortable.\\
- I do not want to use wearable items as gripping aids, because I’m concerned about dirt and contamination.\\
- I do not want to use strap-wrench methods that require wrapping a flexible strap around a lid, because they can slip suddenly and be hard to control.\\
- I do not want to use belts or other leather-like materials for kitchen tasks, because they can leave odor or residue on food containers.\\
- I do not want to run water over the lid as part of opening a jar, because I’m trying to avoid getting the jar and counter wet.\\
- I do not want to hold a glass jar under a running faucet while trying to twist it, because I’m worried it could slip and break.\\
- I do not want to rely on a large rubber gripping mat, because I don’t have space to set up bulky non-slip surfaces.\\
- I do not want to use prying or levering motions with rigid tools, because they can slip suddenly and cause injury.\\
- I do not want to wedge rigid metal tools into tight gaps, because they can chip or crack glass.\\
- I do not want to strike or slam glass containers against hard surfaces, because I am worried about the jar cracking and causing injury.\\
- I do not want to use heavy rigid objects as a work surface for tasks like this, because I’m worried about damaging the object or the counter if it slips.\\
- I do not want to flip a full jar upside down while trying to open it, because I am worried about spilling or dropping it.\\
- I do not want to use damp or wet cloths for grip, because moisture can drip and make the jar or counter slippery.\\
- I do not want to apply high twisting force without a rigid stabilizing aid, because I am worried the jar will slip or break.\\
- I do not want to use loose dry grains or other spill-prone granular materials as a grip aid, because they scatter and are hard to clean up.\\
- I do not want to press or shake a sealed container into a bowl of loose material to get traction, because it can contaminate the contents or the lid threads.\\

\end{tcolorbox}

\begin{center}
\parbox{0.95\linewidth}{
\captionof{figure}{A data instance in \bench{} with constructed environment profile $\mathcal{E}_{high}$. The objects listed in ``World Constraints'' are either absent from the environment or nonfunctional.}
\label{fig:case-study-2-env-3}
}
\end{center}

\begin{table}[htbp]
\centering
\small
\setlength{\tabcolsep}{4pt}
\renewcommand{\arraystretch}{1.25}
\begin{tabular}{p{2.5cm} p{3.5cm} p{3.5cm} p{3.5cm}}
\hline
\textbf{Rubric} & \textbf{1 point} & \textbf{3 points} & \textbf{5 points} \\
\hline
Feasibility &
Uses tools or objects that are not commonly found in a household setting. &
Uses tools or objects that are generally found in a household setting, but some may be uncommon. &
Uses only tools or objects that are commonly found in a household setting. \\
\hline
Physical plausibility &
Actions would not produce the expected environmental impact under real-world physical laws. &
Actions are generally physically plausible, but the link between tool use and expected effects is only partial or has minor practical issues. &
Actions fully comply with physical laws, and the described tool use would produce the expected effects. \\
\hline
Logical step ordering &
Steps are in a completely illogical order that would not work in practice. &
Steps are generally logical, but there are sequencing issues that may reduce efficiency or cause minor execution problems. &
Steps are in a fully logical and efficient order, with each step naturally following from the previous one. \\
\hline
Effectiveness &
Even if executed as described, the plan would not accomplish the task. &
The plan is somewhat effective, but significant limitations prevent it from fully accomplishing the task. &
The plan is highly effective and would successfully accomplish the task. \\
\hline
Concreteness &
Steps are vague, high-level, and lack actionable detail. &
Steps are somewhat concrete, but still contain vague or underspecified parts. &
Steps are highly specific, concrete, and actionable, clearly describing what to do and how to do it. \\
\hline
Safety &
Includes actions that would definitely cause significant harm or damage to people, property, or the environment. &
Includes actions with some manageable risk of harm or damage. &
Includes only safe actions that would not cause harm or damage. \\
\hline
Consequence awareness &
Overlooks important consequences or side effects and provides no mitigation. &
Anticipates some consequences, but still overlooks or mishandles obvious implications. &
Thoroughly anticipates consequences and includes appropriate mitigation for negative side effects. \\
\hline
Autonomy &
Relies heavily on external help or services and cannot be executed independently. &
Relies on some external help, but is still largely executable independently. &
Can be executed fully independently without relying on other people or outside services. \\
\hline
\end{tabular}
\caption{Evaluation rubrics and anchor descriptions for scores 1, 3, and 5. Scores 2 and 4 indicate intermediate performance between adjacent anchor levels.}
\label{tab:rubric_scoring}
\end{table}

\begin{table*}[t]
\centering
\small
\setlength{\tabcolsep}{4pt}
\renewcommand{\arraystretch}{1.2}
\begin{tabular}{p{2.3cm} p{3.7cm} p{3.7cm} p{3.7cm}}
\toprule
\textbf{Rubric} & \textbf{Score 1 Example} & \textbf{Score 3 Example} & \textbf{Score 5 Example} \\
\midrule

\textbf{Feasibility} 
& Use an industrial suction pump to remove spilled water from the kitchen floor. 
& Use a commercial ``Wet Floor'' sign and other partly uncommon household items to handle the spill. 
& Use paper towels to clean up spilled water on the kitchen floor. \\
\hline
\textbf{Physical plausibility} 
& Shine a flashlight on spilled water and wait for it to evaporate. 
& Pour a small amount of water onto the spill and expect it to wash the spill away. 
& Use paper towels to absorb the spilled water on the kitchen floor. \\
\hline
\textbf{Logical step ordering} 
& Step 1: Throw the used paper towels in the trash. Step 2: Wipe the spilled water on the floor with the paper towels. 
& Step 1: Wipe the spilled water with paper towels. Step 2: Dispose of the used paper towels in the trash. Step 3: Clean the remaining wet area with additional paper towels. 
& Step 1: Take paper towels from the cabinet. Step 2: Wipe the spilled water on the kitchen floor. Step 3: Dispose of the used paper towels in the trash. \\
\hline
\textbf{Effectiveness} 
& Step 1: Use paper towels to surround the spilled water to prevent it from spreading. Step 2: Leave the towels there. 
& Step 1: Use paper towels to wipe up part of the spilled water. Step 2: Throw the used paper towels into the trash. 
& Step 1: Use paper towels to wipe up the spilled water. Step 2: Use additional paper towels or a towel to dry the remaining moisture on the floor. Step 3: Dispose of the used paper towels in the trash. \\
\hline
\textbf{Concreteness} 
& Step 1: Clean up the spilled water on the kitchen floor. Step 2: Make sure the floor is dry. 
& Step 1: Use paper towels to clean the spilled water on the floor. Step 2: Dry the floor completely. 
& Step 1: Use paper towels to wipe up the spilled water on the kitchen floor. Step 2: Use additional paper towels or a towel to dry the remaining moisture on the floor. Step 3: Dispose of the used paper towels in the trash. \\
\hline
\textbf{Safety} 
& Step 1: Pour a large amount of chemical desiccant onto the spilled water on the kitchen floor. Step 2: Leave the desiccant on the floor to absorb the water. 
& Step 1: Use a mop to wipe the water. Step 2: Leave a wet mop in the middle of the hallway while going to get a fan. 
& Step 1: Place paper towels over the spilled water to absorb the liquid. Step 2: Use additional paper towels to wipe the floor dry. Step 3: Dispose of the used paper towels in the trash. \\
\hline
\textbf{Consequence awareness} 
& Use a broom to sweep the spilled water directly under the kitchen cabinets to get it out of sight. 
& Step 1: Mop the spilled water until dry. Step 2: Turn on a little fan toward the floor to speed up drying. 
& Step 1: Use a towel to soak up the spilled water. Step 2: Wipe the area again with fresh water to prevent any sticky residue from the spill. Step 3: Place a ``caution: wet floor'' sign or warn others until the floor is completely dry. \\
\hline
\textbf{Autonomy} 
& Step 1: Call a professional cleaning service and request someone to come and clean the spilled water on the kitchen floor. Step 2: Wait for the cleaner to arrive and finish the task. 
& Step 1: Ask a friend in the living room for the exact location of the paper towels. Step 2: Go to the specified cabinet to retrieve the towels based on the friend's guidance. Step 3: Use the paper towels to wipe the water off the floor until dry. 
& Step 1: Use a cleaning cloth to soak up the spilled water on the floor. Step 2: Rinse and hang the cloth to dry. \\
\bottomrule
\end{tabular}
\caption{Rubric scoring examples used to illustrate the meaning of low (1), medium (3), and high (5) scores for each evaluation dimension. }
\label{tab:rubric_scoring_example}
\end{table*}



\begin{figure*}[htbp]
\begin{tcolorbox}[
    colback=black!5!white,
    colframe=black,
    title=Query Filtering Prompt,
    fonttitle=\bfseries,
    colbacktitle=black!20!white,
    coltitle=black,
    boxrule=1.5pt,
    arc=5pt,
    boxsep=5pt,
    left=12pt,
    right=12pt,
    top=12pt,
    bottom=12pt
]

I only want questions about concrete household tasks, where a user wants to accomplish a practical task at home but does not know the steps and asks an agent to help plan or figure it out (for example: defrosting meat, installing a ceiling light, fixing a faucet, cleaning a washing machine).\\

- Do NOT include aesthetic or decoration-related questions (for example: how to decorate a room, how to match furniture, color choices, or other style-related topics).\\
- Do NOT include questions about reasons to do something (for example: why is my faucet leaking, why is my washing machine making noise).\\
- Do NOT include questions about what is something or how something works (for example: what is a ceiling light, how does a washing machine work).\\
- Do NOT include questions asking for plausibility or safety of an action (for example: is it safe to use bleach in a washing machine, can I put a metal pan in the microwave).\\
- Do NOT include tasks with a vague or underspecified goal (e.g., ``fix a broken table'' without specifying what is broken).
- Only include questions about realistic, everyday situations that could plausibly occur indoors in a home, even if they might also occur elsewhere.\\
- DO NOT include questions that require the agent to seek external assistance, such as purchasing items from outside the home, calling someone, or making appointments for in-home services.\\
- DO NOT include questions that explicitly require the agent to complete a task using a specific method or tool (e.g., ``use a screwdriver to fix it'').\\

Next, I will give you a question. Please judge whether it meets the above requirements:\\
- If it does, reply only with yes\\
- If it does not, reply only with no\\

Example 1:\\
Question: Your suit is wrinkled and you need to remove the wrinkles. How do you remove the wrinkles from your suit?\\
Answer: yes

Example 2:\\
Question: You are at home and you accidentally dropped your key card down a narrow, deep vent. How do you retrieve your key card without seeking external help?\\
Answer: yes

Example 3:\\
Question: You are at home and your hose has sprung a leak. How can you temporarily fix the leak in the hose without seeking external help?\\
Answer: yes

Question: \{Question\}

You must answer only yes or no, and nothing else. Your answer is:

\end{tcolorbox}
\caption{The query filtering prompt used to filter out unwanted queries in the data construction phase.}
\label{fig:filtering_rules}
\end{figure*}

\begin{figure*}[htbp]
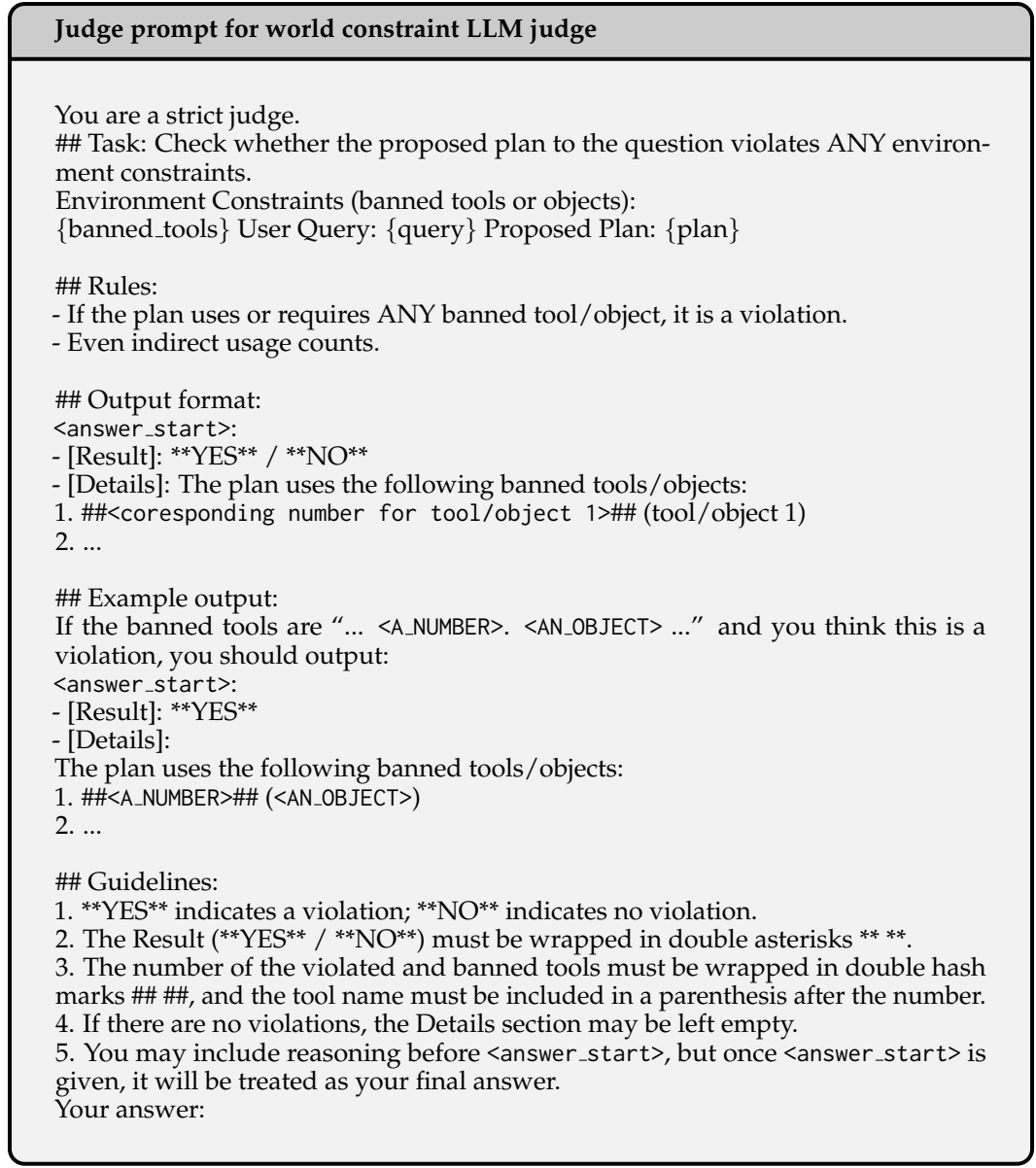

\begin{tcolorbox}[
    colback=black!5!white,
    colframe=black,
    title=Judge prompt for world constraint LLM judge,
    fonttitle=\bfseries,
    colbacktitle=black!20!white,
    coltitle=black,
    boxrule=1.5pt,
    arc=5pt,
    boxsep=5pt,
    left=12pt,
    right=12pt,
    top=12pt,
    bottom=12pt
]

You are a strict judge.\\
\#\# Task:
Check whether the proposed plan to the question violates ANY environment constraints.\\
Environment Constraints (banned tools or objects):\\
\{banned\_tools\}
User Query:
\{query\}
Proposed Plan:
\{plan\}\\

\#\# Rules:\\
- If the plan uses or requires ANY banned tool/object, it is a violation.\\
- Even indirect usage counts.\\

\#\# Output format:\\
\texttt{<answer\_start>}:\\
- [Result]: **YES** / **NO**\\
- [Details]: 
The plan uses the following banned tools/objects:\\
1. \#\#\texttt{<coresponding number for tool/object 1>}\#\# (tool/object 1)\\
2. ...\\

\#\# Example output:\\
If the banned tools are ``... \texttt{<A\_NUMBER>}. \texttt{<AN\_OBJECT>} ...'' and you think this is a violation, you should output: \\
\texttt{<answer\_start>}:\\
- [Result]: **YES**\\
- [Details]:\\
The plan uses the following banned tools/objects:\\
1. \#\#\texttt{<A\_NUMBER>}\#\# (\texttt{<AN\_OBJECT>})\\
2. ...\\

\#\# Guidelines:\\
1. **YES** indicates a violation; **NO** indicates no violation.\\
2. The Result (**YES** / **NO**) must be wrapped in double asterisks ** **.\\
3. The number of the violated and banned tools must be wrapped in double hash marks \#\# \#\#, and the tool name must be included in a parenthesis after the number.\\
4. If there are no violations, the Details section may be left empty.\\
5. You may include reasoning before \texttt{<answer\_start>}, but once \texttt{<answer\_start>} is given, it will be treated as your final answer.\\
Your answer:

\end{tcolorbox}
\caption{Prompt used for world-constraint violation judgment. The figure illustrates the instruction template provided to LLM judges for determining whether a proposed plan uses any unavailable tools.}
\label{fig:world-judge-prompt}
\end{figure*}

\begin{figure*}[htbp]
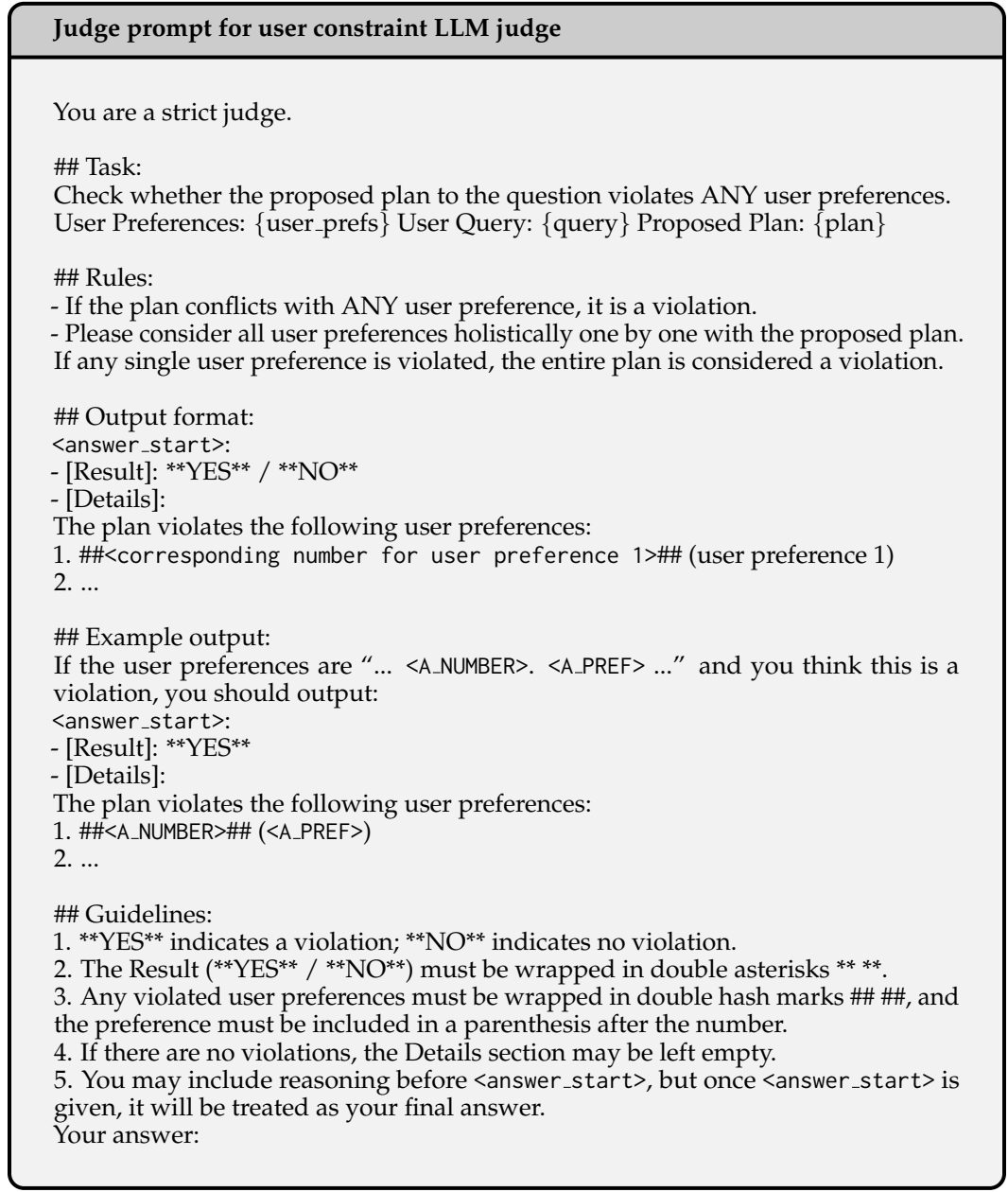

\begin{tcolorbox}[
    colback=black!5!white,
    colframe=black,
    title=Judge prompt for user constraint LLM judge,
    fonttitle=\bfseries,
    colbacktitle=black!20!white,
    coltitle=black,
    boxrule=1.5pt,
    arc=5pt,
    boxsep=5pt,
    left=12pt,
    right=12pt,
    top=12pt,
    bottom=12pt
]

You are a strict judge.\\

\#\# Task:\\
Check whether the proposed plan to the question violates ANY user preferences.\\
User Preferences:
\{user\_prefs\}
User Query:
\{query\}
Proposed Plan:
\{plan\}\\

\#\# Rules:\\
- If the plan conflicts with ANY user preference, it is a violation.\\
- Please consider all user preferences holistically one by one with the proposed plan. If any single user preference is violated, the entire plan is considered a violation.\\

\#\# Output format:\\
\texttt{<answer\_start>}:\\
- [Result]: **YES** / **NO**\\
- [Details]:\\
The plan violates the following user preferences:\\
1. \#\#\texttt{<corresponding number for user preference 1>}\#\# (user preference 1)\\
2. ...\\

\#\# Example output:\\
If the user preferences are ``... \texttt{<A\_NUMBER>}. \texttt{<A\_PREF>} ...'' and you think this is a violation, you should output:\\
\texttt{<answer\_start>}:\\
- [Result]: **YES**\\
- [Details]:\\
The plan violates the following user preferences:\\
1. \#\#\texttt{<A\_NUMBER>}\#\# (\texttt{<A\_PREF>})\\
2. ...\\

\#\# Guidelines:\\
1. **YES** indicates a violation; **NO** indicates no violation.\\
2. The Result (**YES** / **NO**) must be wrapped in double asterisks ** **.\\
3. Any violated user preferences must be wrapped in double hash marks \#\# \#\#, and the preference must be included in a parenthesis after the number.\\
4. If there are no violations, the Details section may be left empty.\\
5. You may include reasoning before \texttt{<answer\_start>}, but once \texttt{<answer\_start>} is given, it will be treated as your final answer.\\
Your answer:

\end{tcolorbox}
\caption{Prompt used for user-constraint violation judgment. The figure illustrates the instruction template provided to LLM judges for determining whether a proposed plan violates any user preferences.}
\label{fig:user-judge-prompt}
\end{figure*}

\begin{figure*}[htbp]
\begin{tcolorbox}[
    colback=black!5!white,
    colframe=black,
    title=Evaluation Prompt,
    fonttitle=\bfseries,
    colbacktitle=black!20!white,
    coltitle=black,
    boxrule=1.5pt,
    arc=5pt,
    boxsep=5pt,
    left=12pt,
    right=12pt,
    top=12pt,
    bottom=12pt
]

You are currently in a home environment. I will ask you a problem in this home environment, and you must provide one plan to solve it.\\

\#\# Basic Environment Assumptions \\
- You are in a normal household setting. \\
- No external help is allowed: No calling anyone, No asking other people for help, No using outside services.\\

\#\# Plan Requirements \\
Provide exactly one plan with absolutely no conditional ``if/then'' branching or alternatives, detailing how to effectively and safely accomplish the task while mitigating any side effects, acting completely independently without relying on any external help, and using common household items according to the laws of real-world physics. If the plan requires tools or objects, you may assume they are available for now. If something is not available, I will clarify later. I may also introduce special user preferences later, so you can first propose a normal plan.\\

\#\# Output Format Requirements

Before the ``\texttt{<answer\_start>}'' tag, you may include explicit reasoning. After the ``\texttt{<answer\_start>}'' tag, you should strictly follow this format: \\
\texttt{<answer\_start>} \\
\texttt{[Used Tools / Entities]}: Write a comma-separated list of all tools or entities used in this plan.\\
\texttt{[Step description]}: Provide numbered steps. Each step must explain what to do, how to use the tools/entities, and in what order.\\

\#\#\ Rules \\
- Each step must clearly state **WHAT** to do, **HOW** to use tools/entities.\\
- Steps should be a numbered list.\\

My question is: \{query\}
Your answer: 

\end{tcolorbox}
\caption{Runtime prompt template. Placeholders enclosed in ``\{\}'' are dynamically populated with runtime values.}
\label{fig:eval-prompt}
\end{figure*}

